\documentclass[10pt,twocolumn,letterpaper]{article}

\usepackage[pagenumbers]{wacv} 
\usepackage[accsupp]{axessibility}  

\usepackage{amsmath,amssymb,amsfonts}
\usepackage{graphicx}
\usepackage{algorithm}
\usepackage{algpseudocode}
\usepackage{multirow}
\usepackage{makecell}
\usepackage{comment}
\usepackage[table]{xcolor}

\newlength{\myheightsecond}
\newlength{\myheightthird}
\newlength{\myheightfourth}

\definecolor{wacvblue}{rgb}{0.21,0.49,0.74}
\usepackage[pagebackref,breaklinks,colorlinks,allcolors=wacvblue]{hyperref}

\def\wacvPaperID{1386} 
\def\confName{WACV}
\def\confYear{2026}

\title{CycleSL: Server-Client Cyclical Update Driven Scalable Split Learning}

\author{Mengdi Wang\textsuperscript{1,2}, Efe Bozkir\textsuperscript{1}, and Enkelejda Kasneci\textsuperscript{1,2} \\
\textsuperscript{1}Technical University of Munich\\
\textsuperscript{2}Munich Center for Machine Learning (MCML)\\
{\tt\small \{mengdi.wang, efe.bozkir, enkelejda.kasneci\}@tum.de}
}

\begin{document}
\maketitle
\begin{abstract}
Split learning emerges as a promising paradigm for collaborative distributed model training, akin to federated learning, by partitioning neural networks between clients and a server without raw data exchange. 
However, sequential split learning suffers from poor scalability, while parallel variants like parallel split learning and split federated learning often incur high server resource overhead due to model duplication and aggregation, and generally exhibit reduced model performance and convergence owing to factors like client drift and lag. To address these limitations, we introduce CycleSL, a novel aggregation-free split learning framework that enhances scalability and performance and can be seamlessly integrated with existing methods. Inspired by alternating block coordinate descent, CycleSL treats server-side training as an independent higher-level machine learning task, resampling client-extracted features (smashed data) to mitigate heterogeneity and drift. It then performs cyclical updates, namely optimizing the server model first, followed by client updates using the updated server for gradient computation. We integrate CycleSL into previous algorithms and benchmark them on five publicly available datasets with non-iid data distribution and partial client attendance. Our empirical findings highlight the effectiveness of CycleSL in enhancing model performance. Our source code is available at~\url{https://gitlab.lrz.de/hctl/CycleSL}. 
\end{abstract}    
\section{Introduction}
As a distributed collaborative machine learning paradigm, split learning (SL,~\cite{gupta2018distributed, vepakomma2018split}), which is also called split neural network (SplitNN), has recently gained strong momentum given the rapid development of distributed computing resources and the ever-growing demand for data privacy. In this paper, we focus on horizontal SL, which means the samples of clients share the same feature space but differ in the sample space. Compared to centralized learning~\cite{soykan2022survey}, where both data storage and model training occur in a centralized manner, in SL, data is distributed across a set of clients, and the training load is amortized between the server and clients. A similar concept is federated learning (FL,~\cite{mcmahan2017communication}), where each client holds a complete model copy and trains its local model using its individual data while a server periodically aggregates client models. SL is different from FL, as in SL, clients only train their models up to a cut layer and send extracted features, which are also called smashed data, to a server, while the server completes the rest of the training without requiring raw data and sends gradients back to clients for their local update. Through such a procedure, the training load can be shared among participating entities without burdening one side too much, and primary data privacy can be guaranteed, given that no data or model sharing is needed. 
\begin{figure*}[!ht]
    \centering
    \includegraphics[height=4.0cm, keepaspectratio]{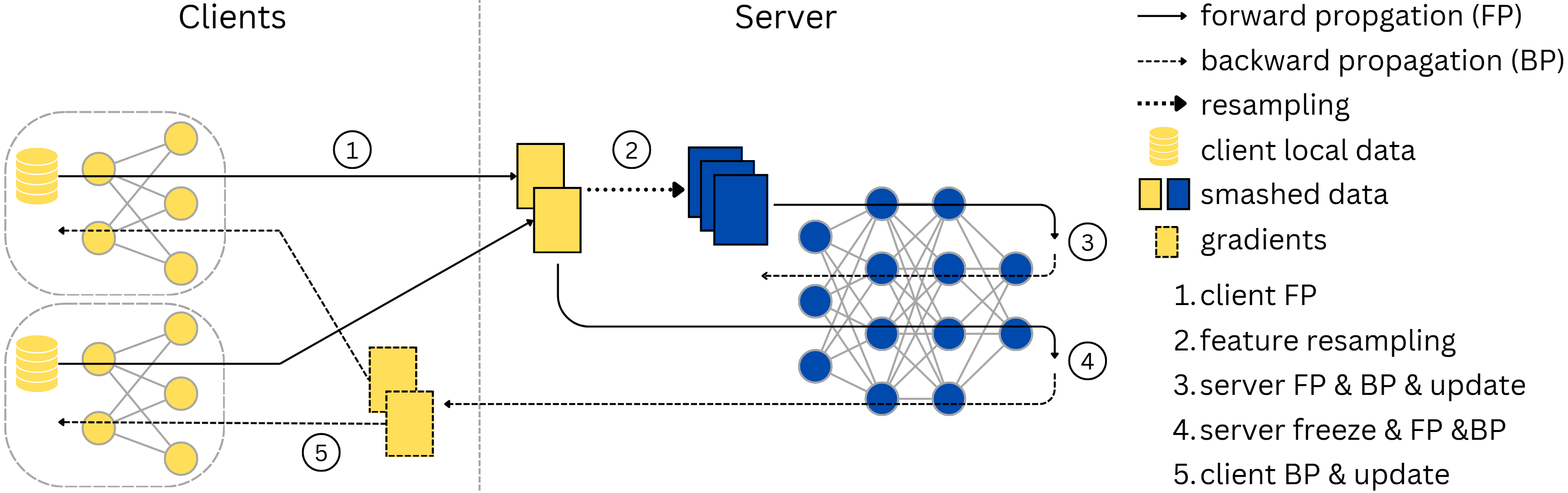}
    \caption{The CycleSL pipeline. After collecting smashed data from clients, CycleSL first forms a global feature dataset on the server side. Then CycleSL resamples features from the dataset to train the server model. Only after the server model is updated, the original feature batches are reused to compute gradients using the latest server model. Lastly, the gradients are sent back to clients for their local update.}
    \label{fig:teaser}
\end{figure*}

The canonical SL happens sequentially, meaning the server only pairs with one client each time. To be the next one being serviced without a cold start, a client needs to retrieve the latest trained client part model from the last coupled client, either through the server or a trusted third party or client-to-client peer sharing. That said, the model trained in sequential SL is not essentially different from a model trained in the centralized manner, except all hidden representations in a client-to-server mini-batch come from the same data holder. This way, sequential SL is on par with centralized learning regarding model performance but suffers from high latency due to poor scalability.
To increase the scalability of sequential SL, many parallelized variants of SL, which are often realized in combination with FL, have been introduced, such as parallel split learning (PSL,~\cite{jeon2020privacy, joshi2021splitfed, lin2024efficient}), split federated learning (SFL or SplitFed,~\cite{thapa2022splitfed, lin2024adaptsfl}), federated split learning (FSL,~\cite{turina2021federated, zhang2023privacy}), and other forms of combinations~\cite{han2021accelerating, pal2021server, abedi2023fedsl, wu2023split}. These methods generally approach scalability by duplicating the server part model or even the server itself, which could incur heavy computation and memory cost on the server side.
Further, due to the integration of FL, they often unavoidably suffer from FL problems such as client drift~\cite{karimireddy2020scaffold, charles2021convergence, wang2021field}. Consequently, they may result in degraded convergence behavior and performance compared to sequential SL. 

In this paper, we propose a novel aggregation-free SL paradigm coined~\textit{\textbf{CycleSL}}, which can be integrated into other scalable SL algorithms to improve model performance while imposing even less resource burden on the server side, especially compared to aggregation-based methods. Firstly, inspired by the model-as-sample strategy~\cite{wang2024turbosvm}, CycleSL models the training task on the server side as a standalone higher-level machine learning task with the smashed data from clients as input samples, and resamples shuffled mini-batches from the collected smashed data to counteract data heterogeneity. In the next, CycleSL adapts the inspiration from alternating (block) coordinate descent~\cite{luenberger1984linear} and performs one step further on the server side, namely updates the server first, and then uses the updated server part model to compute gradients for client local update. In effect, this is equivalent to CycleSL training the clients and server in cyclical turns, rather than in the common end-to-end manner~\cite{glasmachers2017limits} in the same gradient backward flow. A core advantage of such server-client alternating updates is that it could bring in the benefits of coordinate descent solutions and hence further counteract client drift. The main contributions of this work are four-fold:
\begin{itemize}
    \item We present a perspective to model the server-side training in split learning as a higher-level machine learning task and regard the hidden representations from clients as input samples in a feature-as-sample manner.

    \item We propose server-client cyclical update in split learning, which follows the alternating update strategy of coordinate descent and can thus benefit from it. 

    \item We introduce CycleSL, an aggregation-free split learning paradigm that realizes the aforementioned two ideas and can be combined with other scalable split learning algorithms. We integrate CycleSL into three recent methods, including PSL, SFL, and SGLR, and suggest CyclePSL, CycleSFL, and CycleSGLR accordingly. 

    \item We benchmark the aforementioned methods on five publicly available datasets with non-iid data distribution across clients and a 5\% client attendance rate. Our results show that the cycle-version methods remarkably outperform their originals on test performance.
\end{itemize}


\section{Related work}
\subsection{Scalable split learning}
To improve the scalability of vanilla sequential SL, model parallelism has been introduced into SL, which allows multiple clients to collaborate with the server simultaneously. The scalability is often realized through model duplicates on the server side, with one server model clone pairing with one client during training, whose parameters or gradients are later aggregated with the help of FL algorithms such as FedAvg~\cite{mcmahan2017communication} or FedOpt~\cite{reddi2020adaptive}. For instance, PSL~\cite{jeon2020privacy,joshi2021splitfed} enables multi-client connection by replicating the server part model and periodically averaging these model copies. SFL~\cite{thapa2022splitfed}, especially its first variant, SFLV1, does exactly the same on the server side while also engaging a trusted third party (not necessarily the server itself) to aggregate the client part models. In contrast, FSL~\cite{turina2021federated} engages multiple servers rather than multiple models on one server, making the server-client connection more flexible. CPSL~\cite{wu2023split} clusters clients into groups to reduce the latency caused by sequential processing. Unlike the methods above, which generally work for most feed-forward models, FedSL~\cite{abedi2023fedsl} is specialized for recurrent neural networks.

A direct impact of parallelism based on model duplication is the extra computation and memory consumption for server, particularly when the cut layer is shallow and most of the model parameters are kept on the server side. As a consequence, the scalability can be strongly limited. Moreover, as most scalable SL methods rely on federated aggregation, they often unavoidably suffer from FL problems. One of the most challenging aspects is client drift~\cite{karimireddy2020scaffold, charles2021convergence, wang2021field}, which describes the phenomenon that local models converge towards local optima instead of global optimum due to factors like data heterogeneity and data imbalance. This problem is especially challenging in cross-device scenarios~\cite {kairouz2021advances}, where a single user with a personal device like smartphone or tablet and his personal data forms a client. Since each individual can have a unique data distribution, their local optima could diverge to a large extent. The case could be even worse for some SL algorithms like PSL, where only the server model duplicates are aggregated while the client part models remain unsynchronized. As a result, aggregation-based SL methods often converge with lower rate and worse quality than sequential SL. The aggregation-free variants like SFLV2 (the second version of SFL~\cite{thapa2022splitfed}) address this challenge by engaging only one server model and processing all clients one by one, which is in essence not different from sequential SL on the server side and thus does not take advantage of parallelism. Client drift can be further complicated by partial participation~\cite{mcmahan2017communication, kairouz2021advances}, which means only a subset of clients can connect to the server in a round. This is especially an issue for methods without model orchestration on the client side, like PSL, since clients who lag behind due to absence in the last rounds may have a cold start with a less or even untrained local model. 

More recently, a growing body of research aimed to address the aforementioned issues. For instance, EPSL~\cite{lin2024efficient} aggregates the last layer gradients before propagation to reduce the computation burden on the server side. SGLR~\cite{pal2021server} averages gradients before sending them back to clients to mitigate client divergence without client model aggregation. It further sets different learning rates for each end to better adapt them. In~\cite{han2021accelerating}, the authors reduced communication by utilizing localized loss functions on each end. A common limitation of these methods is that they often still rely on model duplication and aggregation, and thus, their improvement can be strongly restricted in practice.

\subsection{Coordinate descent}
Coordinate descent (CD) is a classical optimization technique for multivariate functions. Given a multivariate function, CD approaches its minimum by optimizing along each dimension at a time alternatively rather than along all directions jointly. This procedure is repeated iteratively, (randomly) cycling through all the variables. According to~\cite{wright2015coordinate}, CD has multiple advantages, such as simplicity, scalability, and efficiency for high dimensions and sparsity. Though less studied, the alternating optimization mechanism of CD can result in a biased solution that resembles the effects of implicit regularization under certain conditions~\cite{nakamura2021block, zhao2022high}. The idea of CD and its application in machine learning can be traced back to the early days, covering various domains such as linear sparse problems~\cite{daubechies2004iterative}, Lasso and Ridge regularization~\cite{fu1998penalized}, support vector machines~\cite{platt1998sequential}, and matrix factorization~\cite{zhou2008large}. From the perspective of deep learning, techniques like layer-wise training~\cite{bengio2006greedy, hinton2006fast, palagi2020block} can be regarded as its applications. Though serving different purposes, regularization methods like dropout~\cite{hinton2012improving} also share some analogies with CD. More recently, block CD (BCD) has been studied as a gradient-free optimizer~\cite{carreira2014distributed, zhang2017convergent, zeng2019global} as an alternative to gradient-based and second-order methods such as stochastic gradient descent (SGD) and Newton's method. Recent advances apply BCD or alternating optimization to SL for resource optimization. For example,~\cite{wei2025pipelining} uses BCD in pipelined SL over multi-hop edges to optimize micro-batch sizes. Similarly,~\cite{zhao2025efficient} employs BCD in SFL for large language models over heterogeneous channels, focusing on bandwidth. Furthermore,~\cite{lin2024efficient} proposes BCD for parallel SL in wireless IoT. In principle, CycleSL differs from previous works as it is fully aggregation-free, targeting client drift via feature resampling and cyclical updates for general non-iid data, without further limit assumptions.
\section{Methodology}
Given the problems of former aggregation-based and -free scalable SL algorithms and the benefits of BCD, we introduce CycleSL, a novel aggregation-free SL mechanism that models the training of the server part model as a standalone higher-level task and applies feature resampling to counteract client drift, and updates the server and client part models in cyclical turns, following the alternating update strategy of BCD instead of the common end-to-end pattern. In this paper we stick to the following notations. In an SL task with $N \in \mathbb{N}^+$ clients and $T \in \mathbb{N}^+$ epochs, let $\theta_C$ be the client model and $\theta_S$ be the server model, respectively. Then SL can be regarded as a composition function $\theta_{CS} = \theta_S \circ \theta_C$ with $\theta_{CS}(x) = \theta_S(\theta_C(x))$. We use superscript $t \in [T]$ to denote epoch and subscript $i \in [N]$ to index client. We mark mini-batches using $\mathcal{B}$, with superscripts $x$ for samples, $f$ for feature, and $g$ for gradients. We represent the objective as $\mathcal{L}$. The overarching goal of SL can be given as:
\begin{equation}
    \theta_{C_i}^*, \theta_{S}^* = \arg\min_{\theta_{C_i}, \theta_{S}} \sum_{i=1}^{N} \mathcal{L}_{x_i \sim D_i}(\theta_S(\theta_{C_i}(x_i)))
\end{equation}


\begin{algorithm}[!t]
\footnotesize
\caption{CycleSL pseudo-code}
\label{alg:cyclesl}
\begin{algorithmic}
\State \textbf{Input:} clients $i \in [N]$, client datasets $D_i$, client models $\theta_{C_i}$, server model $\theta_S$, objective $\mathcal{L}$, learning rate $\eta$, SL rounds $T$, server epochs $E$.

\hfill
\State initialize $\theta_{C_1}^0, \theta_{C_2}^0, ..., \theta_{C_N}^0$ and $\theta_S^0$
\For{$t = 0, 1, ..., T-1$}
\For{client $i \in [N]$} \Comment{parallelizable}
\State client samples $\mathcal{B}^x_i \sim D_i$
\State client extracts features $\mathcal{B}^{f}_i \gets \theta_{C_i}^t(\mathcal{B}^x_i)$
\EndFor

\State server forms dataset $D^{f}_S \gets \biguplus_{i \in [N]}\mathcal{B}^{f}_i$ \Comment{Eq.~\ref{eq:server}}
\State $\theta_S^{t,0} \gets \theta_S^{t}$
\For{$e = 0, 1, ..., E - 1$}
\State server resamples $\mathcal{B}^{f}_S \sim D^{f}_S$
\State server updates $\theta_S^{t, e + 1} \gets \theta_S^{t, e} - \eta \nabla_{\theta_S^{t,e}} \mathcal{L}(\theta_S^{t,e}(\mathcal{B}^f_S))$ \Comment{Eq.~\ref{eq:server}}
\EndFor
\State $\theta_S^{t + 1} \gets \theta_S^{t, E}$

\For{client $i \in [N]$} \Comment{parallelizable}
\State server computes gradients $\mathcal{B}^g_i \gets \nabla_{\mathcal{B}^f_i} \mathcal{L}(\theta_S^{t+1}(\mathcal{B}^f_i))$
\State client computes gradients $\nabla_{\theta_i^t} \mathcal{B}^g_i$
\State client updates $\theta_{C_i}^{t+1} \gets \theta_{C_i}^t - \eta \nabla_{\theta_{C_i}^t} \mathcal{B}^g_i$ \Comment{Eq.~\ref{eq:client}}
\EndFor
\EndFor

\end{algorithmic}
\end{algorithm}

\subsection{Higher-level task with feature resampling}
In the following, we illustrate our algorithm in detail, starting with the idea of a standalone server-level task. We observe that it is, in principle, hard to refrain the aggregation-based scalable SL methods from the high resource occupation due to their nature. One exception in this regard is SFLV2 (the second version of SFL), where there is only one server with one model servicing all participating clients one by one, and it is, hence, aggregation-free. However, in substance, SFLV2 conducts sequential SL on the server side and does not benefit much from client parallelization. Moreover, both existing aggregation-based and -free methods are often inevitably influenced by client drift, as all smashed data in a client-to-server batch is extracted with the same client model from samples of the same data holder. For this reason, the server model can be stuck in local minima on the objective surface~\cite{kleinberg2018alternative, xu2022stochastic}. To alleviate this issue, we regard the smashed data collected from clients as input samples for a standalone machine learning task at a higher level. A similar idea was proposed in TurboSVM-FL~\cite{wang2024turbosvm}, where class embeddings from client models are taken as samples to fit a secondary-level support vector classifier. To prevent the client-bound batch problem, CycleSL shuffles and resamples the collected smashed data into random batches that are no longer bound to clients. This mechanism is analogous to random shuffling in conventional centralized deep learning and could confer similar advantages, such as convergence stability and generalization~\cite{haochen2019random, meng2019convergence, safran2020good, yun2021minibatch}. This procedure is mathematically expressed as follows. After receiving feature batches $\mathcal{B}^f_i$ from clients, previous SL paradigms generally update the server model using the batches directly:
\begin{gather}
    \theta_{S}^{t+1} = \arg\min_{\theta_{S}} \sum_{i=1}^{N} \mathcal{L}(\theta_S({B}^{f}_i))
\end{gather}

In contrast, CycleSL first combines these feature batches and forms a global feature dataset on the server side, then randomly resamples mini-batches $B^{f}_S$ from the feature dataset and feeds them into the server model for training:
\begin{gather}
    \theta_{S}^{t+1} = \arg\min_{\theta_{S}} \mathcal{L}_{B^{f}_S \sim \mathcal{D}^{f}_S}(\theta_S({B}^{f}_S)), D^{f}_S = \biguplus_{i \in [N]}\mathcal{B}^{f}_i \label{eq:server}
\end{gather}

A further benefit lies in the reduction of computation and memory cost, since the server part training is not aggregation-based. In addition, as the server part training is a complete standalone task by itself, flexible configurations of hyperparameters, such as learning rate decay and normalization strategies, can be independently determined to improve model performance.

\begin{table}[!t]
    \centering
    \caption{A comparison of SL methods regarding mechanisms and costs on the server side, given number of participating clients $N \in \mathbb{N}^+$. $k$ depends on CycleSL setup and generally $1 \leq k << N$.}
    \label{tab:cost}
    \footnotesize
    \begin{tabular}{l c c c c}
         & Seq. SL & Agg-based SL & Agg-free SL & CycleSL \\
        \hline
        Seq. pair & yes & no & yes & no \\
        Model agg. & no & yes & no & no \\
        Scale gain & no & yes & no & yes \\
        Res. cost & $O(1)$ & $O(N)$ & $O(1)$ & $O(1)$ \\
        Latency & $O(N)$ & $O(1)$ & $O(N)$ & $O(k)$
    \end{tabular}
\end{table}

\subsection{Server-client cyclical update}
In the conventional SL pipeline, though the forward flow of hidden representations and the backward flow of gradients are cut in the middle and transmitted between entities, the client part and server part models are still updated with gradients computed from the same backward propagation, which follows an end-to-end~\cite{glasmachers2017limits} training paradigm in general. Since the server-side training is isolated from the client part as an independent task in CycleSL, we bring in the alternating update strategy of BCD and suggest the cyclical training of server and client models. More precisely, after the server model is optimized, CycleSL freezes the server part model so that no gradients will be computed for its parameters. Then, CycleSL reuses the smashed data from clients and feeds them into the already updated server model to calculate the gradients that will be sent back to clients. The server model shall be unfrozen afterwards. In the next step, clients update their local models with the received gradients, and the next round of SL will start. That said, the server and client part models can be regarded as two blocks in BCD, which are optimized independently and alternatively in rounds. Mathematically, traditional SL algorithms update the client and server models ``simultaneously'', as they use the models at time step $t$ to compute gradients for both the client and server models:
\begin{gather}
    \theta_{C_i}^{t+1} = \arg\min_{\theta_{C_i}} \mathcal{L}(\theta_S^{t}(\theta_{C_i}({B}^{x}_i)))
\end{gather}
whereas CycleSL executes client update after server while respecting the latest update of the server model: 
\begin{gather}
    \theta_{C_i}^{t+1} = \arg\min_{\theta_{C_i}} \mathcal{L}(\theta_S^{t+1}(\theta_{C_i}({B}^{x}_i))) \label{eq:client}
\end{gather}

There are multiple benefits of such an update paradigm. First, since the gradients are only computed with respect to one set of parameters each time (either the server model or the smashed data), while the other set does not trace gradients, the memory usage is not burdening. A further benefit is that the gradient computation regarding smashed data can be easily parallelized, e.g., with one frozen clone of the server model coupling with one client. Thirdly, unlike from SFLV2, where each client is actually paired with a server in a different status due to sequential coupling, in CycleSL, all participating clients in one round couple with the identical server, thus alleviating the impact of client drift. Moreover, CycleSL can potentially benefit from the merits of BCD.

\section{Experiments}
We integrated CycleSL into recent SL algorithms, including PSL~\cite{jeon2020privacy, joshi2021splitfed}, SFL~\cite{thapa2022splitfed}, and SGLR~\cite{pal2021server}, and introduced CyclePSL, CycleSFL, and CycleSGLR accordingly. Particularly, CyclePSL is in essence identical to Algorithm~\ref{alg:cyclesl}. Counting in the two versions of SFL (SFLV1 and SFLV2), we benchmarked in total seven algorithms on five publicly available datasets. We emulated cross-device SL on a node with AMD EPYC 7763 64-Core and NVIDIA A100 80GB PCIe $\times$ 4. To ensure robustness and reliability, we replicated the experiments over five random seeds from $\{0, 1, 2, 3, 4\}$ and reported mean $\pm$ std for each metric. Our code is available at~\url{gitlab.lrz.de/hctl/CycleSL}.

\subsection{Datasets and tasks}
We utilized four classification datasets from LEAF~\cite{caldas2018leaf} and FL-bench~\cite{fl_bench}, both of which serve as standardized benchmark platforms for FL. We chose these platforms because SL is closely related to FL, and the platforms guarantee a great level of reproducibility by providing baseline model architectures, hyperparameters, and data partition strategies. The datasets were FEMNIST~\cite{lecun1998mnist,cohen2017emnist}, CelebA~\cite{liu2015deep}, Shakespeare~\cite{shakespeare2014complete,mcmahan2017communication}, and CIFAR-100~\cite{krizhevsky2009learning}. We also employed a regression dataset OpenEDS2020~\cite{palmero2020openeds2020, palmero2021openeds2020}. The task on this dataset is estimation of gaze direction given eye images captured by VR headset. An overview of the datasets is provided in Table~\ref{tab:datasets}. 
We adopted the model structures suggested by LEAF for the first three tasks, and a ResNet9~\cite{he2016deep} for CIFAR-100, and a appearance-based estimator for OpenEDS2020 as suggested in~\cite{palmero2020openeds2020}. 
For the three image classification tasks, we cut the CNN models in the middle such that both the client and server parts have a similar number of layers. For the language processing task, we kept the embeddings and the LSTM cells on the client side while having the projection head on the server side. The gaze estimator consists of a ResNet50-based feature extractor and a MLP head, which were on the client and server side respectively. We applied the Adam optimizer for all tasks as it is less sensitive to learning rate compared to SGD, and followed the suggested learning rates (if available) or conducted grid search for them. The batch sizes were determined according to data distribution among clients and we left out clients with too few samples that could not fill a full batch. 
We provide an overview of model architectures and hyperparameters in the Appendix. The choices of hyperparameters were identical for clients and server across different SL algorithms for a fair comparison.
\begin{table}[!t]
    \caption{An overview of datasets. Samples/C: samples per client.}
    \label{tab:datasets}
    \centering
    \resizebox{\columnwidth}{!}{%
    \begin{tabular}{l c c c c c}
        Dataset & FEMNIST & CelebA & Shakespeare & CIFAR100 & OpenEDS \\
        \hline
        Task & image class. & smile detec. & char pred. & image class. & gaze esti. \\
        Classes & 62 & 2 & 80 & 100 & - \\
        Clients & 3550 & 9343 & 1129 & 100 & 80 \\
        Samples/C & 226.8 & 21.4 & 3743.3 & 600 & 6880 \\
    \end{tabular}
    }
\end{table}

\begin{table*}[!ht]
    \centering
    \caption{Achieved test loss, accuracy (angular distance for OpenEDS), F1 score, and MCC (Matthews correlation coefficient).}
    \label{tab:results}
    \footnotesize
    \resizebox{\textwidth}{!}{%
    \begin{tabular}{l c c c c c c c c c}
        Method & Metric & FEMNIST & CelebA & Shakespeare & CIFAR$_{(iid)}$ & CIFAR$_{(\alpha=1.0)}$ & CIFAR$_{(\alpha=0.5)}$ & CIFAR$_{(\alpha=0.1)}$ & OpenEDS \\
        \hline
        \multirow{4}{*}{PSL} 
        & Loss & 2.050 $\pm$ 0.235 & 0.648 $\pm$ 0.076 & 3.349 $\pm$ 0.133 & 2.784 $\pm$ 0.036 & 2.638 $\pm$ 0.077 & 2.570 $\pm$ 0.083 & 1.906 $\pm$ 0.206 & 0.030 $\pm$ 0.003 \\
        & Accu & 0.525 $\pm$ 0.046 & 0.809 $\pm$ 0.013 & 0.167 $\pm$ 0.052 & 0.331 $\pm$ 0.026 & 0.351 $\pm$ 0.005 & 0.368 $\pm$ 0.016 & 0.559 $\pm$ 0.051 & 12.92°$\pm$ 0.801 \\
        & F1   & 0.200 $\pm$ 0.037 & 0.808 $\pm$ 0.013 & 0.006 $\pm$ 0.001 & 0.280 $\pm$ 0.016 & 0.280 $\pm$ 0.017 & 0.280 $\pm$ 0.029 & 0.336 $\pm$ 0.041 & NA \\
        & MCC  & 0.506 $\pm$ 0.048 & 0.617 $\pm$ 0.026 & 0.000 $\pm$ 0.001 & 0.325 $\pm$ 0.027 & 0.344 $\pm$ 0.005 & 0.359 $\pm$ 0.016 & 0.542 $\pm$ 0.049 & NA \\
        \hline
        \multirow{4}{*}{SGLR} 
        & Loss & 2.246 $\pm$ 0.248 & 0.487 $\pm$ 0.063 & 3.336 $\pm$ 0.110 & 2.786 $\pm$ 0.043 & 2.628 $\pm$ 0.081 & 2.581 $\pm$ 0.067 & 1.878 $\pm$ 0.187 & 0.033 $\pm$ 0.002 \\
        & Accu & 0.488 $\pm$ 0.051 & 0.826 $\pm$ 0.022 & 0.143 $\pm$ 0.064 & 0.319 $\pm$ 0.019 & 0.360 $\pm$ 0.034 & 0.372 $\pm$ 0.016 & 0.572 $\pm$ 0.042 & 13.60°$\pm$ 0.518 \\
        & F1   & 0.168 $\pm$ 0.036 & 0.825 $\pm$ 0.023 & 0.006 $\pm$ 0.001 & 0.262 $\pm$ 0.023 & 0.291 $\pm$ 0.028 & 0.285 $\pm$ 0.016 & 0.353 $\pm$ 0.041 & NA \\
        & MCC  & 0.467 $\pm$ 0.053 & 0.653 $\pm$ 0.046 & 0.001 $\pm$ 0.001 & 0.312 $\pm$ 0.019 & 0.354 $\pm$ 0.034 & 0.363 $\pm$ 0.016 & 0.556 $\pm$ 0.040 & NA \\
        \hline
        \multirow{4}{*}{SFLV1} 
        & Loss & 1.355 $\pm$ 0.089 & 0.217 $\pm$ 0.019 & 1.908 $\pm$ 0.101 & 2.309 $\pm$ 0.078 & 2.253 $\pm$ 0.077 & 2.350 $\pm$ 0.090 & 2.484 $\pm$ 0.177 & 0.008 $\pm$ 0.005 \\
        & Accu & 0.632 $\pm$ 0.021 & 0.905 $\pm$ 0.011 & 0.454 $\pm$ 0.027 & 0.415 $\pm$ 0.032 & 0.423 $\pm$ 0.026 & 0.416 $\pm$ 0.012 & 0.388 $\pm$ 0.062 & 6.375°$\pm$ 1.995 \\
        & F1   & 0.383 $\pm$ 0.028 & 0.905 $\pm$ 0.011 & 0.147 $\pm$ 0.015 & 0.346 $\pm$ 0.027 & 0.361 $\pm$ 0.020 & 0.339 $\pm$ 0.039 & 0.252 $\pm$ 0.039 & NA \\
        & MCC  & 0.618 $\pm$ 0.022 & 0.811 $\pm$ 0.021 & 0.404 $\pm$ 0.030 & 0.410 $\pm$ 0.033 & 0.417 $\pm$ 0.026 & 0.410 $\pm$ 0.012 & 0.380 $\pm$ 0.060 & NA \\
        \hline
        \multirow{4}{*}{SFLV2} 
        & Loss & 0.509 $\pm$ 0.039 & 0.225 $\pm$ 0.023 & \textbf{1.860} $\pm$ \textbf{0.128} & 2.062 $\pm$ 0.132 & 1.943 $\pm$ 0.125 & 1.963 $\pm$ 0.084 & 1.910 $\pm$ 0.077 & 0.008 $\pm$ 0.005 \\
        & Accu & 0.828 $\pm$ 0.010 & 0.906 $\pm$ 0.009 & \textbf{0.455} $\pm$ \textbf{0.032} & 0.475 $\pm$ 0.041 & 0.493 $\pm$ 0.030 & 0.480 $\pm$ 0.013 & 0.489 $\pm$ 0.039 & 6.487°$\pm$ 1.959 \\
        &  F1  & 0.701 $\pm$ 0.016 & 0.906 $\pm$ 0.010 & 0.156 $\pm$ 0.026 & 0.417 $\pm$ 0.041 & 0.422 $\pm$ 0.028 & 0.409 $\pm$ 0.051 & 0.331 $\pm$ 0.036 & NA \\
        & MCC  & 0.822 $\pm$ 0.010 & 0.814 $\pm$ 0.017 & \textbf{0.407} $\pm$ \textbf{0.035} & 0.471 $\pm$ 0.041 & 0.489 $\pm$ 0.030 & 0.474 $\pm$ 0.013 & 0.479 $\pm$ 0.042 & NA \\
        \hline
        \rowcolor{lightgray}
        & Loss & 0.610 $\pm$ 0.031 & 0.687 $\pm$ 0.058 & 5.044 $\pm$ 0.187 & 2.456 $\pm$ 0.103 & 2.328 $\pm$ 0.177 & 2.161 $\pm$ 0.121 & 1.405 $\pm$ 0.222 & 0.031 $\pm$ 0.004 \\
        \rowcolor{lightgray}
        & Accu & 0.833 $\pm$ 0.016 & 0.855 $\pm$ 0.012 & 0.108 $\pm$ 0.021 & 0.393 $\pm$ 0.016 & 0.422 $\pm$ 0.047 & 0.465 $\pm$ 0.024 & \textbf{0.650} $\pm$ \textbf{0.058} & 13.10°$\pm$ 0.953 \\
        \rowcolor{lightgray}
        & F1   & 0.687 $\pm$ 0.022 & 0.854 $\pm$ 0.011 & 0.011 $\pm$ 0.001 & 0.332 $\pm$ 0.028 & 0.350 $\pm$ 0.046 & 0.370 $\pm$ 0.030 & \textbf{0.436} $\pm$ \textbf{0.033} & NA \\
        \rowcolor{lightgray}
        \multirow{-4}{*}{\makecell{Cycle-\\PSL}} & MCC  & 0.827 $\pm$ 0.017 & 0.710 $\pm$ 0.024 & 0.000 $\pm$ 0.002 & 0.387 $\pm$ 0.017 & 0.416 $\pm$ 0.047 & 0.458 $\pm$ 0.025 & \textbf{0.637} $\pm$ \textbf{0.057} & NA \\
        \hline
        \rowcolor{lightgray}
        & Loss & 0.619 $\pm$ 0.060 & 0.548 $\pm$ 0.037 & 5.343 $\pm$ 0.347 & 2.402 $\pm$ 0.073 & 2.290 $\pm$ 0.159 & 2.093 $\pm$ 0.051 & \textbf{1.385} $\pm$ \textbf{0.219} & 0.032 $\pm$ 0.002 \\
        \rowcolor{lightgray}
        & Accu & 0.823 $\pm$ 0.019 & 0.860 $\pm$ 0.013 & 0.080 $\pm$ 0.014 & 0.401 $\pm$ 0.018 & 0.427 $\pm$ 0.045 & 0.472 $\pm$ 0.029 & 0.638 $\pm$ 0.047 & 13.40°$\pm$ 0.632 \\
        \rowcolor{lightgray}
        & F1   & 0.678 $\pm$ 0.031 & 0.860 $\pm$ 0.012 & 0.011 $\pm$ 0.001 & 0.342 $\pm$ 0.031 & 0.348 $\pm$ 0.049 & 0.376 $\pm$ 0.031 & 0.436 $\pm$ 0.044 & NA \\
        \rowcolor{lightgray}
        \multirow{-4}{*}{\makecell{Cycle-\\SGLR}}
        & MCC  & 0.817 $\pm$ 0.019 & 0.721 $\pm$ 0.025 & 0.000 $\pm$ 0.001 & 0.395 $\pm$ 0.019 & 0.421 $\pm$ 0.045 & 0.465 $\pm$ 0.029 & 0.624 $\pm$ 0.046 & NA \\
        \hline
        \rowcolor{lightgray}
        & Loss & \textbf{0.489} $\pm$ \textbf{0.031} & \textbf{0.205} $\pm$ \textbf{0.018} & 2.078 $\pm$ 0.091 & \textbf{1.923} $\pm$ \textbf{0.062} & \textbf{1.875} $\pm$ \textbf{0.085} & \textbf{1.850} $\pm$ \textbf{0.079} & 1.825 $\pm$ 0.131 & \textbf{0.007} $\pm$ \textbf{0.004} \\
        \rowcolor{lightgray}
        & Accu & \textbf{0.839} $\pm$ \textbf{0.008} & \textbf{0.918} $\pm$ \textbf{0.005} & 0.437 $\pm$ 0.017 & \textbf{0.491} $\pm$ \textbf{0.027} & \textbf{0.502} $\pm$ \textbf{0.026} & \textbf{0.522} $\pm$ \textbf{0.031} & 0.522 $\pm$ 0.029 & \textbf{5.938°}$\pm$ \textbf{1.553} \\
        \rowcolor{lightgray}
        & F1   & \textbf{0.710} $\pm$ \textbf{0.006} & \textbf{0.917} $\pm$ \textbf{0.006} & \textbf{0.169} $\pm$ \textbf{0.016} & \textbf{0.433} $\pm$ \textbf{0.040} & \textbf{0.426} $\pm$ \textbf{0.023} & \textbf{0.434} $\pm$ \textbf{0.042} & 0.344 $\pm$ 0.041 & NA \\
        \rowcolor{lightgray}
        \multirow{-4}{*}{\makecell{Cycle-\\SFL}}
        & MCC  & \textbf{0.834} $\pm$ \textbf{0.008} & \textbf{0.835} $\pm$ \textbf{0.011} & 0.390 $\pm$ 0.018 & \textbf{0.486} $\pm$ \textbf{0.027} & \textbf{0.497} $\pm$ \textbf{0.027} & \textbf{0.517} $\pm$ \textbf{0.031} & 0.511 $\pm$ 0.032 & NA \\
    \end{tabular}
    }
\end{table*}

We partitioned the data among clients in non-iid ways, which made the experiments more challenging and mimicked real life. For FEMNIST, CelebA, and Shakespeare, we followed the fixed non-iid distributions provided in LEAF. The CIFAR-100 dataset was partitioned using FL-bench, following the Dirichlet distribution~\cite{hsu2019measuring} with different $\alpha$ values to emulate different levels of data heterogeneity across clients. Smaller $\alpha$ implies stronger data heterogeneity.For OpenEDS2020, we regarded each user as a client. Further, we emulated partial participation with a client attendance rate of 5\%. This process means we randomly sampled 5\% of clients in each round, and only these clients could connect with the server. For all tasks, we conducted sample-wise data split~\cite{wang2021field}, which means for each client, we had a proportion of samples reserved for test, rather than a held-out set of clients who never met the server during training. The train-test split ratio was 90\%-10\% for all tasks. The main reason for applying sample-wise data split is that for some SL algorithms like PSL and SGLR, there is no model aggregation procedure on the client side, which means the held-out clients receive no updates throughout training, and a test involving these clients is irrelevant.

Our experiments are notably distinct from previous ones, as all five datasets we used contain a great amount of clients, and the data distribution among clients is strongly non-iid, whereas previous experiments often engaged only few clients and little heterogeneity. In other words, our experiments are close to a cross-device~\cite{kairouz2021advances} case where each client can be an individual user with a personal device with particular data, whereas existing results were commonly obtained in cross-silo scenarios where large-scale institutions (e.g., hospitals) act as clients. We believe the former is more challenging regarding scalability, data heterogeneity, and partial participation. Besides, experiments in cross-device scenarios are more practical in collaborative computing. 

\subsection{Results}
The test metrics are summarized in Table~\ref{tab:results} and visualized in the Appendix. We observed that for FEMNIST, the cycle-version algorithms outperformed all their originals, especially the aggregation-based ones, to a large extent, while CycleSFL performed the best. In detail, CyclePSL, CycleSGLR, and CycleSFL respectively advanced the test accuracies of PSL, SGLR, and SFLV1 from 52.5\% to 83.3\%, from 48.8\% to 82.3\%, and from 63.2\% to 83.9\%. Compared to the aggregation-free method, namely SFLV2, CycleSFL also yielded a 1.1\% accuracy improvement. Another finding is that CycleSL can also lead to an increase in model robustness and stability, as a decrease in metric standard deviations can be observed for all metrics. For CelebA, since the task is simple (binary), the non-cycle algorithms can already deliver good results. However, CyclePSL, CycleSGLR, and CycleSFL still boosted model performance on average by 4.6\%, 3.4\%, and 1.3\% regarding test accuracy. Moreover, we found that the improvement was obtained even in the case of overfitting, as can be inferred from the test loss on the CelebA task. When PSL and SGLR and their cycle-versions were overfitted (an increase in test loss, but other metrics were not negatively impacted), the cycle-version methods still yielded lower test loss at earlier rounds. Whilst not being the best-performing method for all metrics on Shakespeare, CycleSFL delivered the highest F1 score and was still on par with SFL with respect to other metrics. For CIFAR-100, we observed that the cycle-version methods generally surpassed their originals across different levels of heterogeneity. Particularly, CycleSFL, which was commonly the best performing method, improved the test accuracy from 47.5\% to 49.1\% in iid case, from 49.3\% to 50.2\% when $\alpha=1.0$, from 48.0\% to 52.2\% when $\alpha=0.5$, and from 48.9\% to 52.2\% when $\alpha=0.1$, respectively, compared to the second best method SFLV2. Another noticeable finding is that with the increase of data heterogeneity ($\alpha$ decreases), algorithms that do not require model aggregation on the client side, like PSL and SGLR, began to overtake the aggregation-based methods, which could be attributed to model personalization~\cite{tan2022towards}. Still, the cycle-version methods, namely CyclePSL and CycleSGLR, maintained their dominance over PSL and SGLR, with an increase in test accuracy from 55.9\% to 65.0\% and from 57.2\% to 63.8\% under extreme data heterogeneity (i.e., $\alpha=0.1$). The superiority of CycleSL is also maintained in regression tasks, as the cycle-version methods vastly outperform other SL algorithms with lower cosine distance loss and angular estimation error on the OpenEDS2020 dataset, with CycleSFL being the lead.

CycleSL is also promising in terms of convergence behavior. We measured the convergence speed of each algorithm by the minimum number of training rounds required to surpass specific test accuracy thresholds (45\% for FEMNIST, 75\% for CelebA, 35\% for Shakespeare, 30\% for CIFAR-100) in Table~\ref{tab:convergence} in the Appendix. We could learn from the results that the cycle-version methods commonly converged at (much) earlier stages, especially compared to their aggregation-based originals, unless the original methods failed to converge, while CycleSFL was overall the fastest algorithm in this benchmark. For instance, on the FEMNIST dataset, PSL, SGLR, and SFLV1 only started to make noticeable progress in test loss after 100 epochs, whereas their cycle-versions, CyclePSL, CycleSGLR, and CycleSFL, almost reached their convergence around the 100th round. Similar phenomena can also be observed for other tasks. Especially for the CelebA task, although PSL and SGLR started to overfit after roughly 300 and 400 rounds, CyclePSL and CycleSGLR still delivered lower test loss at earlier time points.

\begin{table}[!t]
    \caption{Impact of cut layer on test accuracy.}
    \label{tab:ablation_cut}
    \centering
    \scriptsize
    \begin{tabular}{c c c c c}
        Cut & $iid$ & $\alpha=1.0$ & $\alpha=0.5$ & $\alpha=0.1$ \\
        \hline
        1 & 0.531$\pm$0.039 & 0.558$\pm$0.026 & 0.542$\pm$0.008 & 0.547$\pm$0.037 \\
        \hline
        2 & 0.517$\pm$0.027 & 0.533$\pm$0.043 & 0.516$\pm$0.023 & 0.531$\pm$0.050 \\
        \hline
        3 & 0.487$\pm$0.035 & 0.511$\pm$0.020 & 0.500$\pm$0.027 & 0.511$\pm$0.040 \\
        \hline
        4 & 0.475$\pm$0.017 & 0.479$\pm$0.019 & 0.479$\pm$0.030 & 0.496$\pm$0.019\\
        \hline
        5 & 0.443$\pm$0.020 & 0.466$\pm$0.033 & 0.460$\pm$0.017 & 0.468$\pm$0.029 \\
        \hline
        6 & 0.427$\pm$0.030 & 0.414$\pm$0.027 & 0.412$\pm$0.013 & 0.376$\pm$0.048 \\        
    \end{tabular}
\end{table}

\begin{table}[!t]
    \caption{Impact of server epoch on test accuracy.}
    \label{tab:ablation_epoch}
    \centering
    \scriptsize
    \begin{tabular}{c c c c c}
        Epoch & $iid$ & $\alpha=1.0$ & $\alpha=0.5$ & $\alpha=0.1$ \\
        \hline
        1 & 0.487$\pm$0.013 & 0.518$\pm$0.025 & 0.502$\pm$0.021 & 0.518$\pm$0.039 \\
        \hline
        2 & 0.503$\pm$0.014 & 0.520$\pm$0.010 & 0.522$\pm$0.024 & 0.587$\pm$0.041 \\
        \hline
        4 & 0.464$\pm$0.013 & 0.497$\pm$0.030 & 0.472$\pm$0.038 & 0.587$\pm$0.033 \\
        \hline
        8 & 0.477$\pm$0.020 & 0.497$\pm$0.029 & 0.498$\pm$0.034 & 0.622$\pm$0.053 \\
    \end{tabular}
\end{table}
\subsection{Ablation study}
\paragraph{Impact of cut layer}\label{para:cut}
In SL, the choice of cut layer plays a key role in many factors such as model performance, computation overhead, transmission cost, and privacy risk. We concentrated on the impact of the cut layer on model performance for CycleSFL on the CIFAR-100 dataset with different data heterogeneity levels. For simplicity we explored the influence of block-wise cut point rather than layer-wise using the ResNet9 model. The ResNet9 model contains four convolutional blocks, two residual blocks, and one projection head. Therefore, there are six possible cut positions in total. We cut the model at each possible point and recorded the achieved test accuracy in Table~\ref{tab:ablation_cut}. As the table shows, a shallower cut point can lead to better model performance. We attributed this to the fact that since in CycleSL there is only one model on the server side, the impact of data heterogeneity and client drift mostly resides in client models. In such a case, a reduced model complexity on the client side, i.e., a shallower cut layer, would help improve convergence. However, the choice of cut layer is a complex bargain game between clients and server involving many other factors like privacy and transmission overhead. For example, a shallower cut layer may lead to data leak, as input data can be more easily reconstructed due to stronger correlation between input data and activations. We refer readers to relevant works for detailed analysis~\cite{yan2022optimal, wu2023split, kim2023bargaining}.

\paragraph{Impact of server epoch}
One simple but effective way to boost convergence in FL is to allow clients to train for multiple epochs before aggregation. In CycleSL, since the training on the server side is a standalone task, a similar strategy can be applied, namely allowing the server to update for multiple rounds before computing gradients for clients. To this end, we investigated the impact of server epoch on model performance on CIFAR-100 with different levels of data distribution heterogeneity among clients. We employed only CycleSFL for simplicity, with the number of server training epochs $E$ scaling from one to eight, and Table~\ref{tab:ablation_epoch} provides the performance. The results show that when data heterogeneity is not drastic (i.e., iid, $\alpha=1.0$, $\alpha=0.5$), the increase of server training pass from one to two generally led to better performance. However, further increments of server rounds resulted in a decrease in test accuracy. In contrast, under extreme distribution heterogeneity ($\alpha=0.1$), the increase of server epoch up to eight can consistently improve model performance. In general, server epoch increment is associated with better model performance when both the number of epochs and data heterogeneity are not high. We speculated that the choice of server training round in CycleSL is a trade-off between model personalization~\cite{tan2022towards} and regularization, as a stronger trained server may lead to stabler gradient steps for clients, which could motivate client models to converge to similar local optima but reduces local model personalization. To further explore the existence and effect of gradient stability, we introduced another study in the following.
\begin{table}[!t]
    \centering
    \caption{Grad. norms over clients and epochs (scaled by $10^{-3}$).}
    \label{tab:grad_norm}
    \resizebox{\columnwidth}{!}{%
    \begin{tabular}{l c c c c}
        Method & $iid$ & $\alpha=1.0$ & $\alpha=0.5$ & $\alpha=0.1$ \\
        \hline
        PSL & 2.281$\pm$1.508 & 2.547$\pm$6.715 & 4.569$\pm$47.20 & 4.369$\pm$38.97 \\
        \hline
        SGLR & 1.938$\pm$1.249 & 2.150$\pm$5.478 & 3.284$\pm$35.53 & 2.754$\pm$14.04 \\
        \hline
        SFLV1 & 1.963$\pm$1.217 & 2.192$\pm$5.309 & 3.860$\pm$39.08 & 3.751$\pm$30.54\\
        \hline
        SFLV2 & 1.389$\pm$0.908 & 1.537$\pm$3.750 & 2.648$\pm$26.53 & 2.404$\pm$18.97 \\
        \hline
        CyclePSL & 1.307$\pm$0.913 & 1.521$\pm$4.157 & 2.805$\pm$29.96 & 2.773$\pm$23.98 \\
        \hline
        CycleSGLR & 1.147$\pm$0.770 & 1.290$\pm$3.393 & 1.979$\pm$21.15 & 1.764$\pm$8.942 \\
        \hline
        CycleSFL & 1.311$\pm$0.852 & 1.468$\pm$ 3.674 & 2.575$\pm$26.04 & 2.578$\pm$19.56 
    \end{tabular}
    }
\end{table}

\paragraph{Gradient stability}
\label{para:grad}
We speculated that one of the potential factors for why CycleSL performs well lies in that it may produce gradients of both low magnitude and low variance, which could benefit the training with stable gradient steps, especially on the client side. To investigate this effect, we recorded the norms of the gradients (averaged inside mini-batch) that the server sends back to clients on the CIFAR-100 dataset across various heterogeneity degrees, and computed the mean and std of the norms over SL epochs and clients. The results are reported in Table~\ref{tab:grad_norm}. As the level of data heterogeneity escalates, both the magnitude and variance of gradient norms generally increase. Compared to other SL methods, the cycle-version methods commonly yield gradients of both lower norm and deviation. This could possibly benefit the training procedure with stabler optimization steps and verify our hypothesis to a certain degree. However, it should be noticed that smaller gradients do not inevitably result in better model performance, since CycleSGLR delivers gradients of lower norms compared to CyclePSL and CycleSFl but performs worse than them. Besides, vanishing gradients can result in slow learning.
\section{Discussion}
\subsection{Application and future work}
As CycleSL can be easily combined with other scalable SL methods, its applications can be valuable in practice. Particularly, the choice of foundation algorithm could be determined according to the individual case. For instance, when client-side aggregation is risky due to privacy concerns or model personalization is wished, CyclePSL and CycleSGLR can be employed. In contrast, if a performant model and faster convergence are desired, CycleSFL is a good option. Given its low burden and high scalability, CycleSL can be particularly beneficial in cross-device cases where the number of participating clients is large and the computing resources per client are generally limited, such as in edge computing. Moreover, since the server part training is an isolated task in CycleSL, an independent and flexible setting for hyperparameters such as model split, regularization method, and aggregation frequency~\cite{lin2024adaptsfl} could be applied to further advance model performance and convergence rate. Due to its robustness, CycleSL is potentially more tolerant of hyperparameters such as learning rates and batch sizes, which can be costly to tune, especially in privacy-sensitive scenarios. The feature resampling strategy of CycleSL lays the foundation for future works like ensemble learning and knowledge distillation in SL, which have already become popular in FL~\cite{lin2020ensemble, attota2021ensemble, wu2024fedel}. 

Convergence analysis in SL, which can be regarded as function composition, is considerably more challenging than in FL. Existing studies for its convergence property like~\cite{nesterov2013gradient, beck2013convergence, richtarik2014iteration, zeng2019global} are in an early stage. Therefore we demonstrate the effectiveness of our method with comprehensive empirical validation and ablation studies and regarded theoretical analysis as a valuable direction for future work. 
In addition, since we deem the smashed data from clients as input samples for a standalone task in CycleSL, client selection or reinforcement based on client features could be a further research direction, and techniques like sample- or client-wise attention mechanisms~\cite{cheng2021ba} for smashed data await further investigation. In this work, we restrict our discussion to SL with label sharing~\cite{gupta2018distributed, vepakomma2018split} for simplicity, meaning that there is only one split point and clients send both extracted features and labels to the server. One can easily extend the methods to SL without label sharing with two or even more model cut points on demand.

\subsection{Limitations}
The integration of CycleSL can notably reduce the memory and computation burden on the server side over the aggregation-based algorithms. One drawback of CycleSL is that its latency and computation burden are larger than SFLV2, as SFLV2 only conducts forward and backward propagation once on the server side, whereas the smashed data is fed into the server part model twice in CycleSFL. However, CycleSFL's memory cost is lower than SFLV2, since CycleSFL only computes gradients with respect to one set of parameters each time while the other set is frozen. The latency and computation can be lightened with a reasonable choice of hyperparameters, such as batch size for the second-stage task. Another limitation of CycleSL is that the combination of CycleSL and other scalable algorithms may inherit the problems of these methods themselves. For instance, we learned from the test loss on the CelebA dataset that PSL and SGLR began to overfit after roughly 300 and 400 iterations, respectively. Although the integration of CycleSL reached lower test loss at earlier rounds, CyclePSL and CycleSGLR did not refrain from overfitting. Similarly, if the original algorithms fail to converge, the incorporation of CycleSL may not converge as well, as can be inferred from the results of CyclePSL and CycleSGLR on the Shakespeare dataset. Further, an imbalanced model split could lead to suboptimal performance of CycleSL, as CycleSFL, with only a single linear layer on the server side, cannot outperform SFL in the Shakespeare task.

\subsection{Privacy concern}
Integrating CycleSL into other scalable SL methods requires no additional data or model transfer. Therefore, the privacy claims of the original algorithms still apply, and the former privacy-enhancing mechanisms, like differential privacy and k-anonymity, would still work. Furthermore, thanks to its aggregation-free property, CycleSL is presumablely more robust against client-level noise injection than the aggregation-based methods and could improve the privacy-utility trade-off, as handling noisy models is, in general, a harder task than handling noisy samples. Moreover, since the smashed data is resampled on the server side and not client-bound anymore, the noise in the hidden representations could be compensated over features from different clients. Random resampling can also contribute to model robustness and stability. The impact of malicious participants could also be reduced with such a procedure.

\section{Conclusion}
In this work, we presented a novel aggregation-free scalable SL algorithm called CycleSL. Inspired by BCD, CycleSL models the server part training as a standalone higher-level task, and applies resampling to smashed data to counteract client drift. Further it updates the server and clients in cyclical turns and benefits from BCD merits. CycleSL can be seamlessly combined with other SL methods to optimize model performance and resource consumption. By integrating CycleSL into existing methods, including PSL, SGLR, and SFL, we introduced CyclePSL, CycleSGLR, and CycleSFL accordingly. Our empirical results demonstrate that CycleSL can significantly improve model performance.

\section*{Acknowledgements}
We acknowledge the funding by the Deutsche Forschungsgemeinschaft (DFG, project number KA 4539/5-1), and by the Munich Center for Machine Learning (MCML).

{
    \small
    \bibliographystyle{ieeenat_fullname}
    \bibliography{main}
}

\newpage
\appendix
\onecolumn
\section*{Appendix}
\section{Implementation}
We implemented in eight ten SL/FL algorithms, including PSL, SFLV1, SFV2, SGLR, FedAvg (excluded from paper due to page limit), CyclePSL, CycleSFL, and CycleSGLR. All methods were implemented with PyTorch. An overview of our experiment environment is given in Table~\ref{tab:environment}. Additionally we provided the code for sequential SL (SSL) and its cycle-version (CycleSSL). Since they are not scalable methods, we excluded them from benchmark. Our implementation can be found on~\url{https://gitlab.lrz.de/hctl/CycleSL}. By following the instructions we provided, our experiment results should be completely reproducible up to numerical error. 

\section{Randomness}
To ensure the robustness and reliability of results, we replicated the experiments for five times, with each time being initialized with a different and unique random seed from $\{0, 1, 2, 3, 4\}$. The random seed was fed to all libraries which can potentially be influenced by randomness in the beginning, including $numpy.random.seed$, $torch.manual\_seed$, and $random.seed$. We reported all our metrics in form of mean $\pm$ std over the five seeds.

\section{Environment}
The benchmarks were conducted on a computer with AMD EPYC 7763 and NVIDIA A100 80GB PCIe. All SL algorithms were implemented in PyTorch. An overview of the hardware and software of our environment is given in Table~\ref{tab:environment}. It should be noticed that although we conducted the experiments on a powerful machine, we have tried to optimize our code so that it can be run on a normal PC as well, even without dedicated GPU.
\begin{table}[!ht]
    \caption{Experiment environment by July 12th, 2025.}
    \label{tab:environment}
    \centering
    \begin{tabular}{l c}
    Server & Specification  \\
    \hline
    CPU & AMD EPYC 7763 64-Core \\
    GPU & NVIDIA A100 80GB PCIe$\times$4 \\
    RAM & 1TB \\
    OS & Ubuntu 22.04.5 LTS \\
    \end{tabular}
    \qquad
    \begin{tabular}{l c}
    Library & Version \\
    \hline
    Python & 3.13.5 by Anaconda \\
    PyTorch & 2.7.1 for CUDA 12.8 \\
    Scikit-Learn & 1.7.0\\
    WandB & 0.21.0 \\
    \end{tabular}
\end{table}

\section{Server-side latency} 
As CycleSL performs feature resampling and feeds the smashed data into the server model twice, we were interested in quantifying the time cost incurred. On the CIFAR-100 dataset across different data heterogeneity levels, we measured the elapsed time on the server side between the arrival of smashed data and the transmission of gradients while ignoring the cost caused by model transfer between CPU and GPU. The recorded time cost was accumulated over SL rounds. We mainly compared three algorithms, namely SFLV1, SFLV2, and CycleSFL, as they are representative of aggregation-based, aggregation-free, and CycleSL respectively, and equivalent in regard of client behavior. The results are provided in Table~\ref{tab:latency}. As the results show, the server processing time is rather consistent across various degrees of data heterogeneity. Among three methods, SFLV2 is the most time-efficient. SFV1 requires more computation time due to model aggregation. As CycleSL conducts feature resampling and a second time model feedforward, it incurs the highest server processing time. However, this time cost can be optimized with sophisticated choices for hyperparameters on the server side, especially batch size.
\begin{table}[!ht]
    \centering
    \caption{Server-side processing time (in seconds).}
    \label{tab:latency}
    \begin{tabular}{l c c c c}
         Method & $iid$ & $\alpha=1.0$ & $\alpha=0.5$ & $\alpha=0.1$ \\
         \hline
         SFLV1 & 19.3 & 19.7 & 19.1 & 19.9 \\
         \hline
         SFLV2 & 10.4 & 10.4 & 10.7 & 10.3 \\
         \hline
         CycleSFL & 39.1 & 39.5 & 39.7 & 39.9
    \end{tabular}
\end{table}

\section{Toy example for gradient stability} \label{sec:toy}
Along side the empirical validation about gradient stability in subsection~\ref{para:grad}, we also provided a toy example in this regard in 1D. Consider a simplified regression task in SL where both the client and server part models are a single linear layer with one neuron. We further ignore activation functions and bias terms. Then, for a sample point $(x, y)$, its predicted value can be given as $\hat y =  w_s w_c x$, where $w_c$ and $w_s$ are the parameters of client and server part models respectively. Consider mean squared error as a loss function, i.e., $\ell = (y - \hat y)^2$. Traditionally, SL follows an end-to-end~\cite{glasmachers2017limits} pattern, which means both $w_c$ and $w_s$ are updated in the same gradient backward flow: $w_s' \gets w_s - 2 \eta w_c x (w_s w_c x - y), w_c' \gets w_c - 2 \eta w_s x (w_s w_c x - y)$ where $w_c'$ and $w_s'$ are the updated parameters and $\eta > 0$ is learning rate. In contrast, CycleSL updates $w_s$ and $w_c$ one after another: $w_s' \gets w_s - 2 \eta w_c x (w_s w_c x - y), w_c' \gets w_c - 2 \eta w_s' x (w_s' w_c x - y)$.

Comparing the two update strategies, it is clear that both are identical with respect to server side training, while they respectively use the old and the new server models to update the client part model. We now compare the two gradient steps for the client model, namely $2 \eta w_s x (w_s w_c x - y)$ and $2 \eta w_s' x (w_s' w_c x - y)$, when approaching convergence. For simplicity, we limit our discussion to the case where all $w_c, w_s, x, y > 0$ and $w_s w_c x > y$. All other cases can be analyzed similarly. Since $w_s w_c x > y$, we expect that with a proper choice of $\eta$, $w_s'$ is reduced for a decently small step during server training, i.e., $\frac{y}{w_c x} < w_s' < w_s$ such that $w_s' w_c x$ shrinks towards $y$. Observe the function $f(w_s) = w_s x (w_s w_c x - y)$ and its derivative $f'(w_s) = 2 w_s w_c x^2 - xy = x(w_s w_c x - y + w_s w_c x)$. When approaching convergence, i.e. $w_s w_c x - y \rightarrow 0^+$, we have $f'(w_s) > 0$, which means $f(w_s)$ is increasing in its neighborhood. Thus for $\frac{y}{w_c x} < w_s' < w_s$ the following applies:
$f(w_s') < f(w_s) \Leftrightarrow 2 \eta w_s' x (w_s' w_c x - y) < 2 \eta w_s x (w_s w_c x - y)$. That said, the alternating update strategy of CycleSL could possibly result in a smaller gradient step on the client side compared to the conventional end-to-end paradigm when approaching convergence, which may lead to stabler gradient steps.

\section{Communication cost} 
As FL is more communication-costly in contrast to SL, some variants of FL have been proposed to reduce the model transmission overhead, such as knowledge distillation-based FL (KDFL,~\cite{li2019fedmd}) and partial training FL (PTFL~\cite{caldas2018expanding, diao2020heterofl}, also known as sub-model extraction FL). We gave a conceptual comparison of communication cost among standard FL, KDFL, PTFL, and SL (not necessarily CycleSL). Let $N$ be the number of clients, $M$ be the number of parameters in a full model, $T$ be training rounds, $B$ be batch size, $D$ be the size of public/global dataset, $L$ be the activation dimension at cut layer for SL, $0<k<1$ be the proportion of parameters shared in PTFL. Generally $L << M$ and $kM << M$. The commnuication cost is given in Table~\ref{tab:communication}. It should be noted that these training paradigms commonly have many advanced variants~\cite{alam2022fedrolex, wang2024feddse, wu2024fiarse, wang2023dafkd} and mechanisms can substantially change, and hence this comparison does not always apply.
\begin{table}[!ht]
    \centering
    \caption{Communication cost comparison.}
    \label{tab:communication}
    \begin{tabular}{l c c c c}
        & FL & KDFL & PTFL & SL \\
         \hline
        Require public/server data & no & yes & no & no \\
        Communication cost & $O(MT)$ & $O(DT)$ & $O(kMT)$ & $O(BLT)$
    \end{tabular}
\end{table}

\section{Hyperparameters} 
For FEMNIST, CelebA, and Shakespeare, we followed the hyperparameters suggested by LEAF, especially the learning rates, unless they performed to be too small or too large in the experiment. For CIFAR-100 and OpenEDS2020, we kept batch size of 64 and conducted grid search for learning rate in range of $\{1e-5, 1e-4, 1e-3, 1e-2, 1e-1\}$ for each algorithm. The overall best performing learning rate was $1e-4$ and $1e-3.5$. An overview of hyperparameters is given in Table~\ref{tab:hyperparameters}.
\begin{table}[!ht]
    \caption{Details of hyperparameters (consistent for clients and server across SL methods unless specifically mentioned).}
    \label{tab:hyperparameters}
    \centering
    \begin{tabular}{l c c c c c}
        Dataset & FEMNIST & CelebA & Shakespeare & CIFAR-100 & OpenEDS2020 \\
        \hline
        Batch Size & 32 & 16 & 32 & 64 & 64 \\
        Optimizer & Adam & Adam & Adam & Adam & Adam \\
        Learning Rate & $3e-4$ & $1e-2$ & $3e-2$ & $1e-4$ & $1e-3.5$ \\
    \end{tabular}
\end{table}

\section{Model architecture} 
For FEMNIST, CelebA, and Shakespeare, we followed the model structures as suggested by LEAF (\url{https://github.com/TalwalkarLab/leaf/tree/master/models}), which are CNN, CNN, and LSTM, respectively. For the two CNN models, we cut them in the middle such that client and server parts have similar numbers of layers. For the LSTM model, we kept the embeddings and recurrent cells on the client while the projection head on the server. The model architectures, sources, and cut points are summarized in Tables~\ref{tab:femnist_cnn}--\ref{tab:shakespeare_lstm} respectively.
\begin{table}[!ht]
    \caption{CNN architecture for the FEMNIST task.}
    \label{tab:femnist_cnn}
    \centering
    \begin{tabular}{l c}
        \textbf{Layer} & \textbf{Specification} \\ 
        \hline
        Input & shape $1 \times 28 \times 28$ \\
        Conv2d & kernel size 5, in/out channel 1/32, same padding \\ 
        ReLU & - \\
        MaxPooling & kernel size 2, stride 2 \\
        Conv2d & kernel size 5, in/out channel 32/64, same padding \\ 
        ReLU & - \\
        MaxPooling & kernel size 2, stride 2 \\
        \hline
        \textbf{Cut Layer} & \textbf{client/server cut point} \\
        \hline
        Flatten & - \\
        Linear & in/out dimension 3136/2048 \\
        ReLU & - \\
        Linear & in/out dimension 2048/62 \\
    \end{tabular}
\end{table}

\begin{table}[!ht]
    \caption{LSTM architecture for the Shakespeare task.}
    \label{tab:shakespeare_lstm}
    \centering
    \begin{tabular}{l c}
        \textbf{Layer} & \textbf{Specification} \\
        \hline
        Embedding & number of embeddings 80, dimension 8 \\
        LSTM & 3n/hidden dimension 8/256, hidden layers 2 \\ 
        \hline
        \textbf{Cut Layer} & \textbf{client/server cut point} \\
        \hline
        Linear & in/out dimension 256/80 \\
    \end{tabular}
\end{table}

\begin{table}[!ht]
    \caption{CNN architecture for the CelebA task.}
    \label{tab:celeba_cnn}
    \centering
    \begin{tabular}{l c}
        \textbf{Layer} & \textbf{Specification} \\ 
        \hline
        Input & shape $3 \times 84 \times 84$ \\
        Conv2d & kernel size 3, in/out channel 3/32, same padding \\ 
        BatchNorm2d & - \\
        MaxPooling & kernel size 2, stride 2 \\
        ReLU & - \\
        Conv2d & kernel size 3, in/out channel 32/32, same padding \\ 
        BatchNorm2d & - \\
        MaxPolling & kernel size 2, stride 2 \\
        ReLU & - \\
        \hline
        \textbf{Cut Layer} & \textbf{client/server cut point} \\
        \hline
        Conv2d & kernel size 3, in/out channel 32/32, same padding \\ 
        BatchNorm2d & - \\
        MaxPolling & kernel size 2, stride 2 \\
        ReLU & - \\
        Conv2d & kernel size 3, in/out channel 32/32, same padding \\ 
        BatchNorm2d & - \\
        MaxPolling & kernel size 2, stride 2 \\
        ReLU & - \\
        Flatten & - \\
        Linear & in/out dimension 800/2 \\ 
    \end{tabular}
\end{table}

For the CIFAR-100 task, we adopted an ResNet9 network~\cite{he2016deep}. Our implementation followed~\url{https://www.kaggle.com/code/kmldas/cifar10-resnet-90-accuracy-less-than-5-min?scriptVersionId=38462746&cellId=28}). ResNet9 contains four convolutional blocks, two residual blocks, and a projection head. To balance the number of layers, we kept two convolutional blocks and one residual block on the client side, while the rest and the projection head on the server side. Implementation details can be found in~\url{https://github.com/AnonymWriter/CycleSL/blob/main/models.py}. In the ablation study, we further investigated the influence of cut point on CycleSL. The corresponding results were discussed in subsection~\ref{para:cut}.

For the OpenEDS2020 task, we employed an appearance-based estimation model as suggested in~\cite{palmero2020openeds2020}. The model consists of a pre-trained ResNet50-based feature extractor and a MLP head. We kept the feature extractor on the client side and the projection head on the server side.

\section{Data distribution}
The histograms of number of samples per client of the FEMNIST, CelebA, and Shakespeare datasets are given in Figure~\ref{fig:appendix_hist}. Particularly, the CIFAR-100 dataset was partitioned with Dirichlet distribution using different $\alpha$ values to emulate different levels of data heterogeneity across clients (smaller $\alpha$ implies stronger data heterogeneity). The partition was done via FL-bench~\cite{fl_bench}. The impact of $\alpha$ on label distribution can be observed in Figure~\ref{fig:alpha}. OpenEDS2020 is considered as self-partitioned, as the data comes user-wise and each user was treated as a client in our experiments.
\setlength{\myheightthird}{3.0cm}
\begin{figure}[!ht]
    \centering
    \subfloat[FEMNIST]{\includegraphics[height=\myheightthird, keepaspectratio]{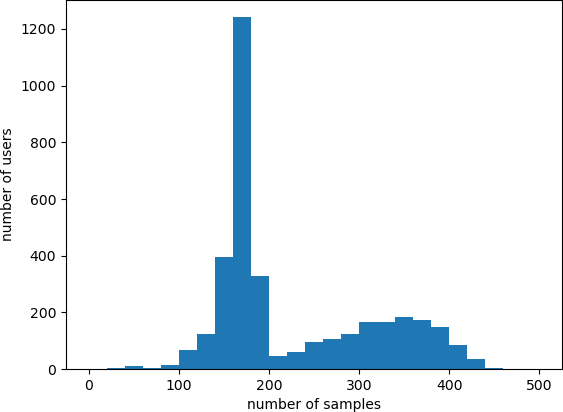}
    \label{fig:femnist_hist}}
    \hfill 
    \subfloat[CelebA]{\includegraphics[height=\myheightthird, keepaspectratio]{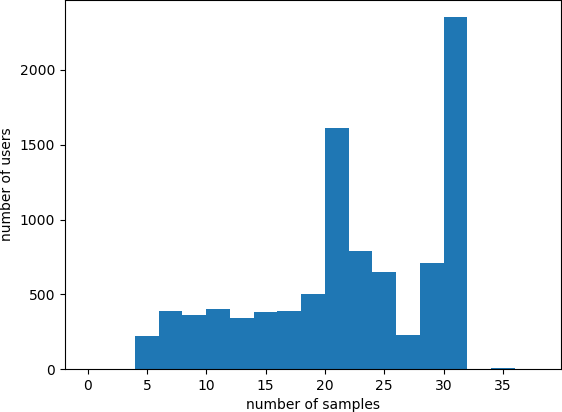}
    \label{fig:celeba_hist}}
    \hfill
    \subfloat[Shakespeare]{\includegraphics[height=\myheightthird, keepaspectratio]{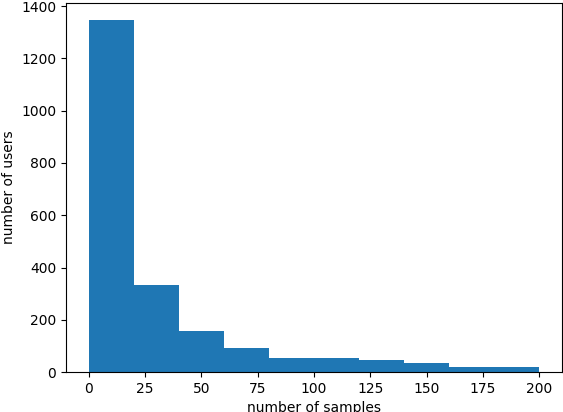}
    \label{fig:shakespeare_histo}}
    \hfill
    \subfloat[OpenEDS2020]{\includegraphics[height=\myheightthird, keepaspectratio]{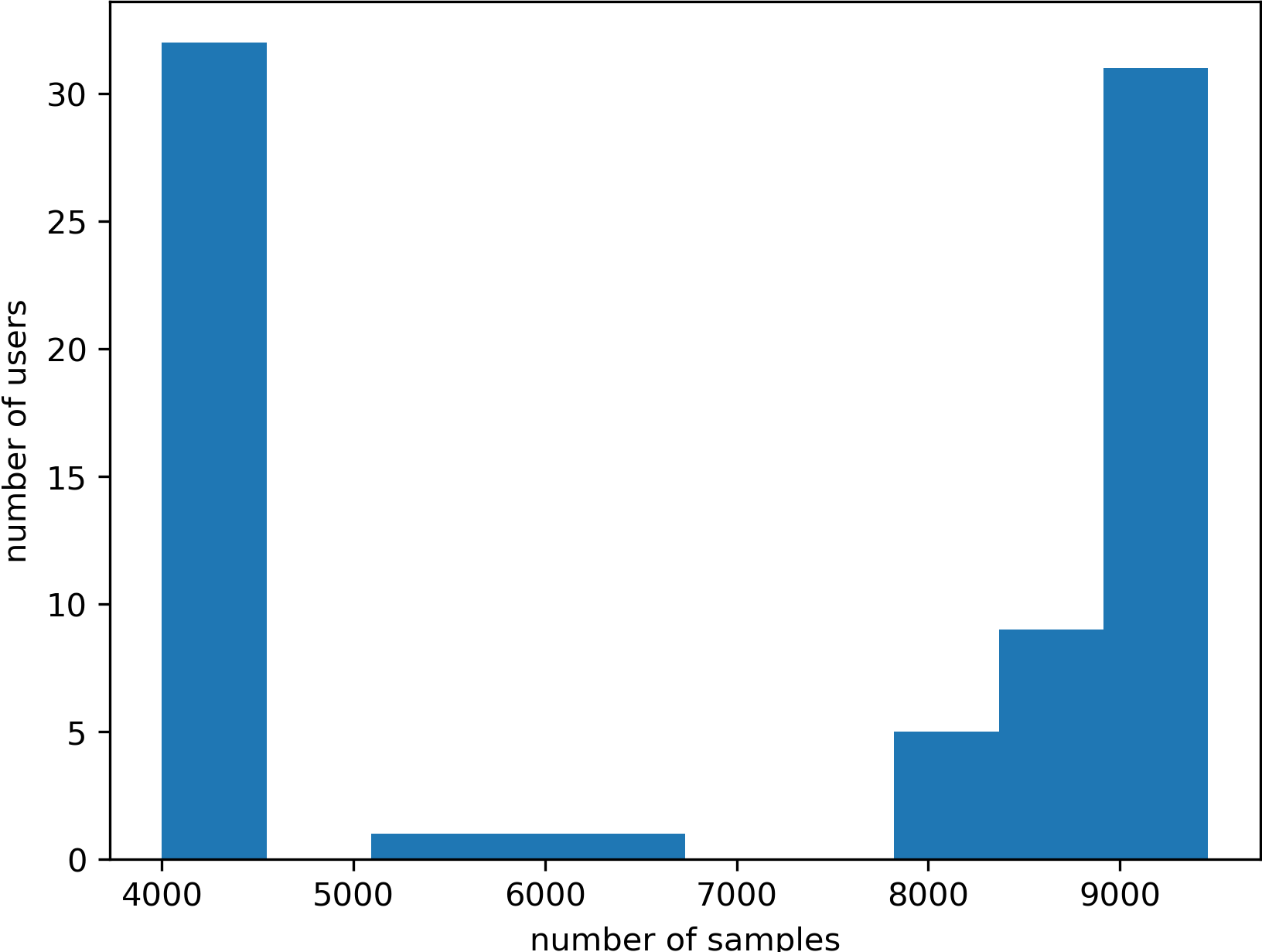}
    \label{fig:openeds_histo}}
    \caption{Histograms of samples per user for the FEMNIST, CelebA, Shakespeare, and OpenEDS2020 datasets.}
    \label{fig:appendix_hist}
\end{figure}

\begin{figure}[!ht]
    \centering
    \subfloat[iid]{\includegraphics[height=\myheightthird, keepaspectratio]{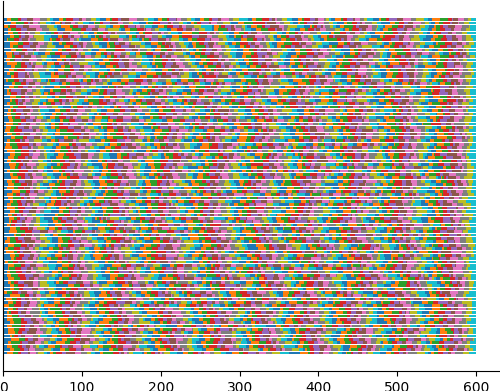} \label{fig:appendix_iid}}
    \hspace{0.5cm}
    \subfloat[$\alpha=1.0$]{\includegraphics[height=\myheightthird, keepaspectratio]{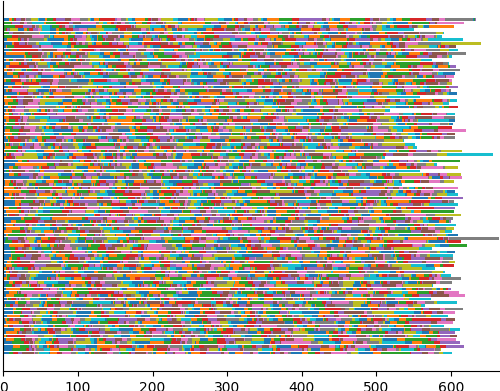}
    \label{fig:appendix_alpha_1.0}}
    \hspace{0.5cm}
    \subfloat[$\alpha=0.5$]{\includegraphics[height=\myheightthird, keepaspectratio]{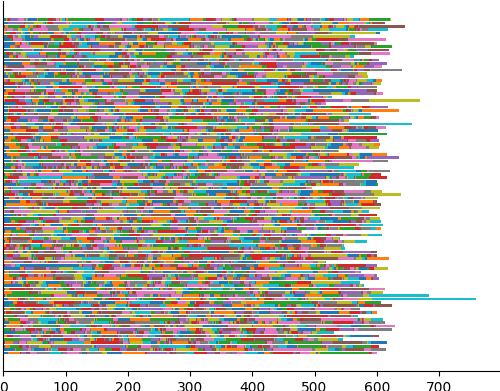}
    \label{fig:appendix_alpha_0.5}}
    \hspace{0.5cm}
    \subfloat[$\alpha=0.1$]{\includegraphics[height=\myheightthird, keepaspectratio]{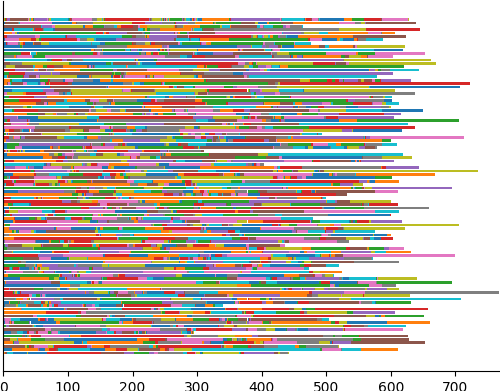}
    \label{fig:appendix_alpha_0.1}}
    \caption{Label distributions among clients in CIFAR-100 (smaller $\alpha$ implies stronger data heterogeneity).}
    \label{fig:alpha}
\end{figure}

\section{Additional results}

\subsection{Convergence rate}
We measured the convergence speed of each algorithm by recording the first time that their test accuracies surpassed certain thresholds (45\% for FEMNIST, 75\% for CelebA, 35\% for Shakespeare, 30\% for CIFAR-100) in Table~\ref{tab:convergence}. 
\begin{table}[!ht]
    \caption{Minimal epochs required to reach certain test accuracy for each task (45\% for FEMNIST, 75\% for CelebA, 35\% for Shakespeare, 30\% for CIFAR-100). Smaller is better.}
    \label{tab:convergence}
    \centering
    \resizebox{\textwidth}{!}{%
    \begin{tabular}{l c c c c c c c}
        Method & FEMNIST & CelebA & Shakespr & CIFAR$_{(iid)}$ & CIFAR$_{(\alpha=1.0)}$ & CIFAR$_{(\alpha=0.5)}$ & CIFAR$_{(\alpha=0.1)}$ \\
        \hline
        PSL & 459 & 255 & $>600$ & $>1000$ & $>1000$ & $>1000$ & 476 \\
        SGLR & 517 & 244 & $>600$ & $>1000$ & $>1000$ & $>1000$ & 476\\
        SFLV1 & 184 & 39 & 154 & 293 & 253 & 289 & 342 \\
        SFLV2 & 8 & 31 & 105 & 131 & 118 & 149 & \textbf{162} \\
        FedAvg & 287 & 65 & 386 & 335 & 337 & 488 & $>1000$ \\
        \hline
        CyclePSL & 8 & 94 & $>600$ & 584 & 552 & 526 & 422 \\
        CycleSGLR & 10 & 94 & $>600$ & 581 & 534 & 532 & 395 \\
        CycleSFL & \textbf{6} & \textbf{29} & \textbf{98} & \textbf{116} & \textbf{107} & \textbf{132} & \textbf{162} \\
    \end{tabular}
    }
\end{table}

\subsection{Metric plots}
The test metrics, including loss (cross entropy), accuracy, F1 score, and MCC (Matthews correlation coefficient), are plotted in Figures~\ref{fig:appendix_femnist_result}--\ref{fig:appendix_cifar100_0.1_result}, respectively. It should be noticed that although some methods like PSL and SGLR overfitted for CelebA (increase in test loss), metrics like accuracy and F1 score were not negatively impacted. Hence we still reported metrics around 600th epoch.
\setlength{\myheightsecond}{3.0cm}
\begin{figure}[!ht]
    \centering
    \subfloat{\includegraphics[height=\myheightsecond, keepaspectratio]{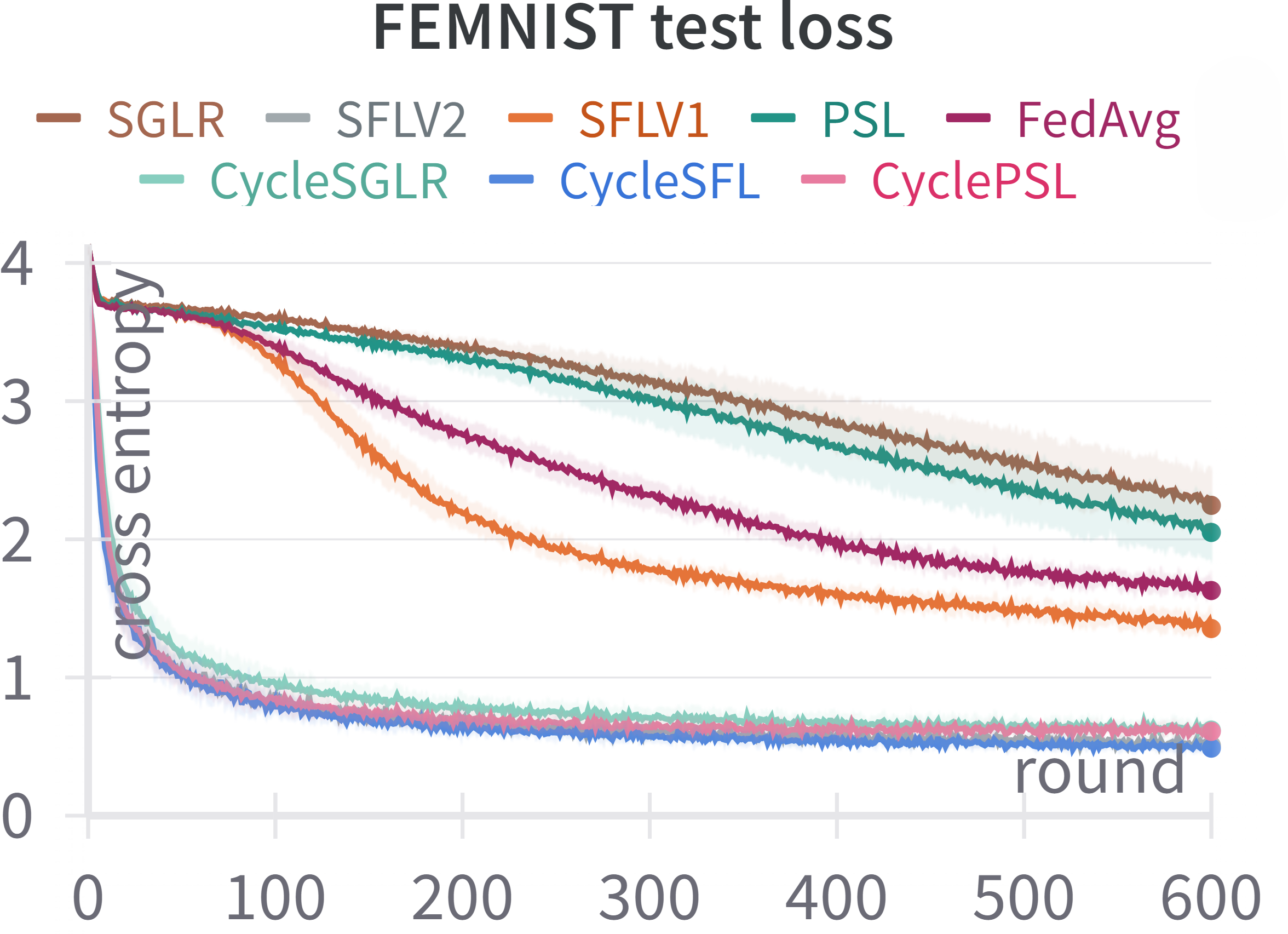}
    \label{fig:appendix_femnist_loss}}
    \hfill
    \subfloat{\includegraphics[height=\myheightsecond, keepaspectratio]{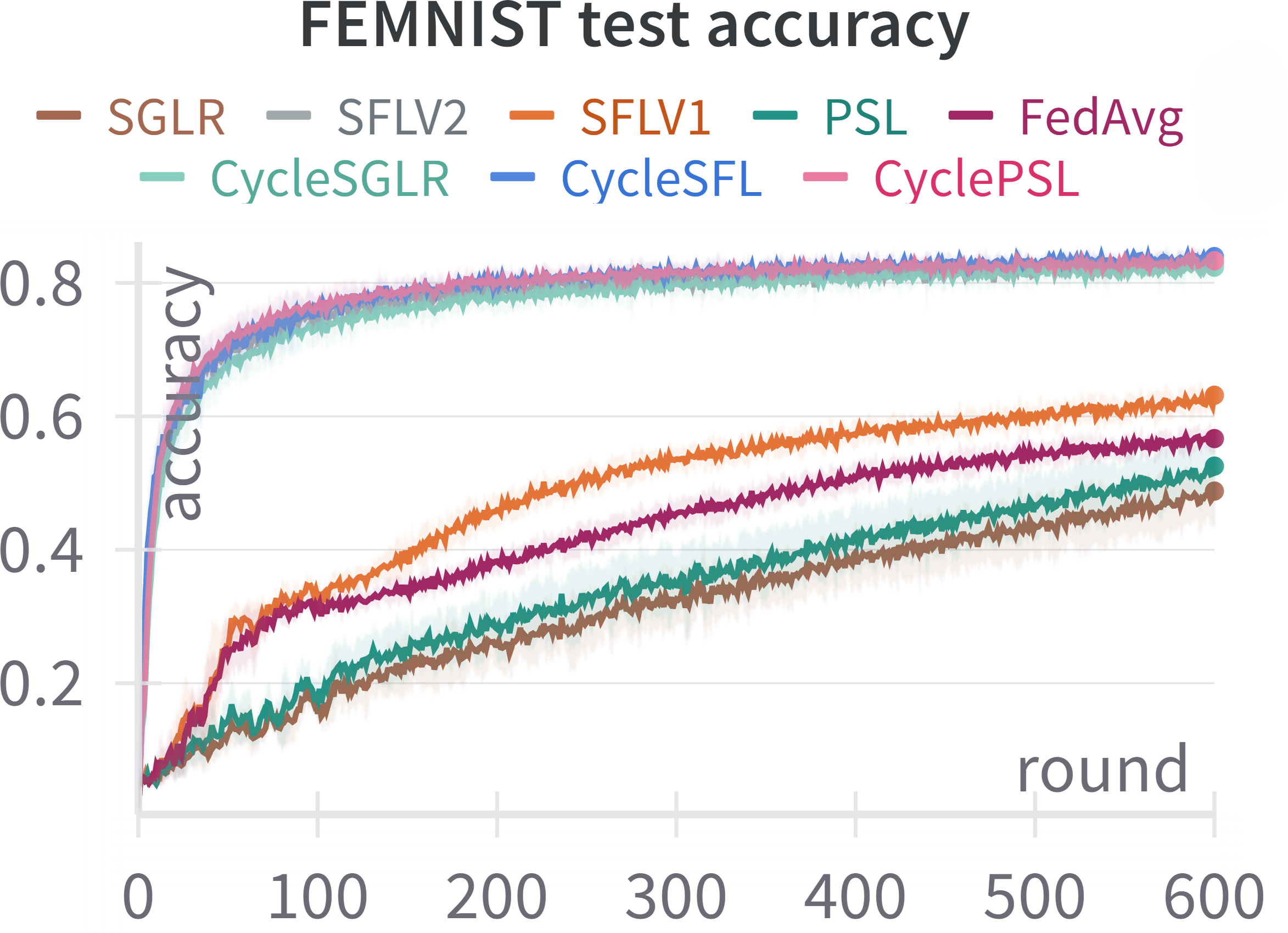}
    \label{fig:appendix_femnist_accu}}
    \hfill
    \subfloat{\includegraphics[height=\myheightsecond, keepaspectratio]{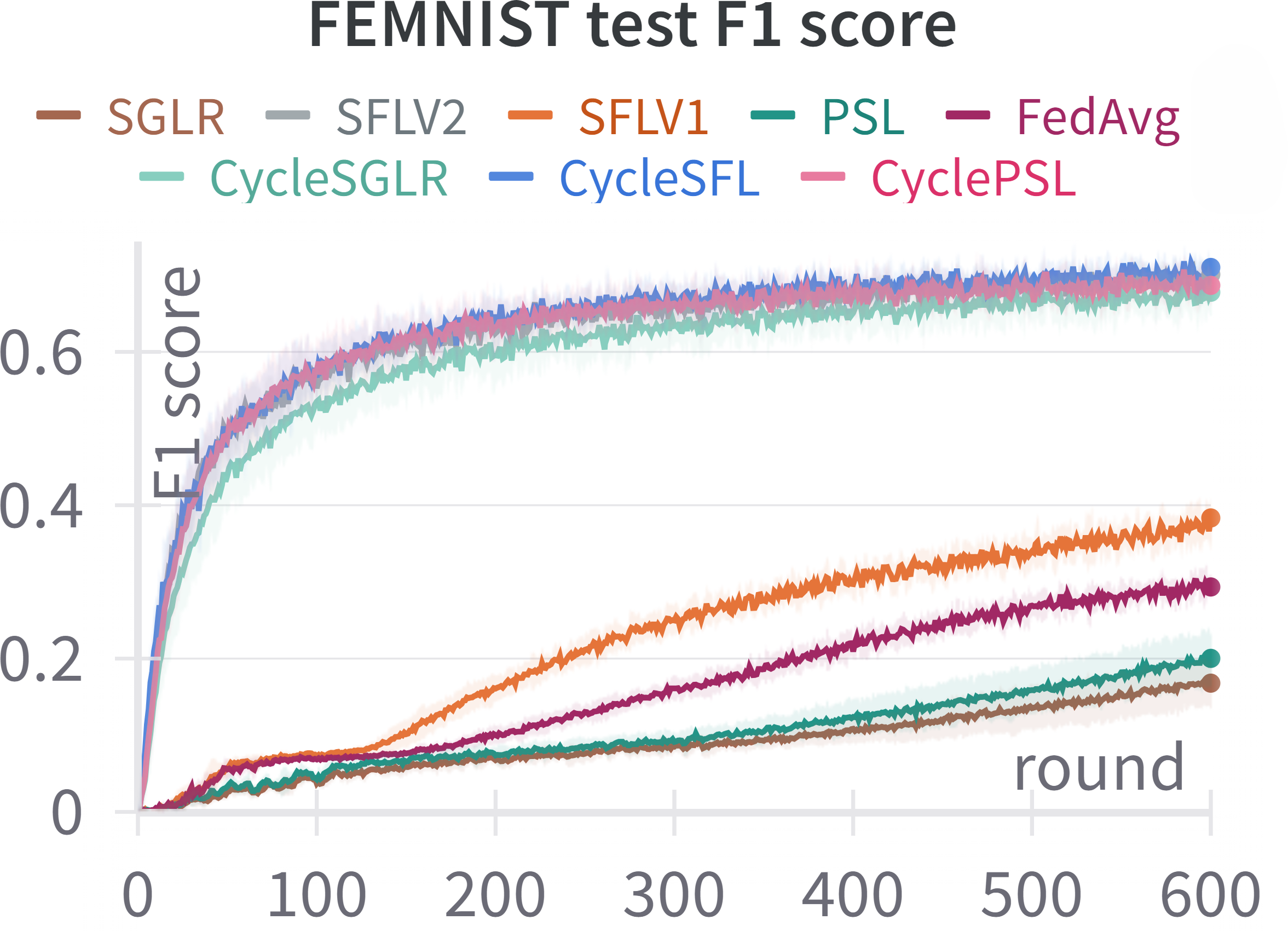}
    \label{fig:appendix_femnist_f1}}
    \hfill
    \subfloat{\includegraphics[height=\myheightsecond, keepaspectratio]{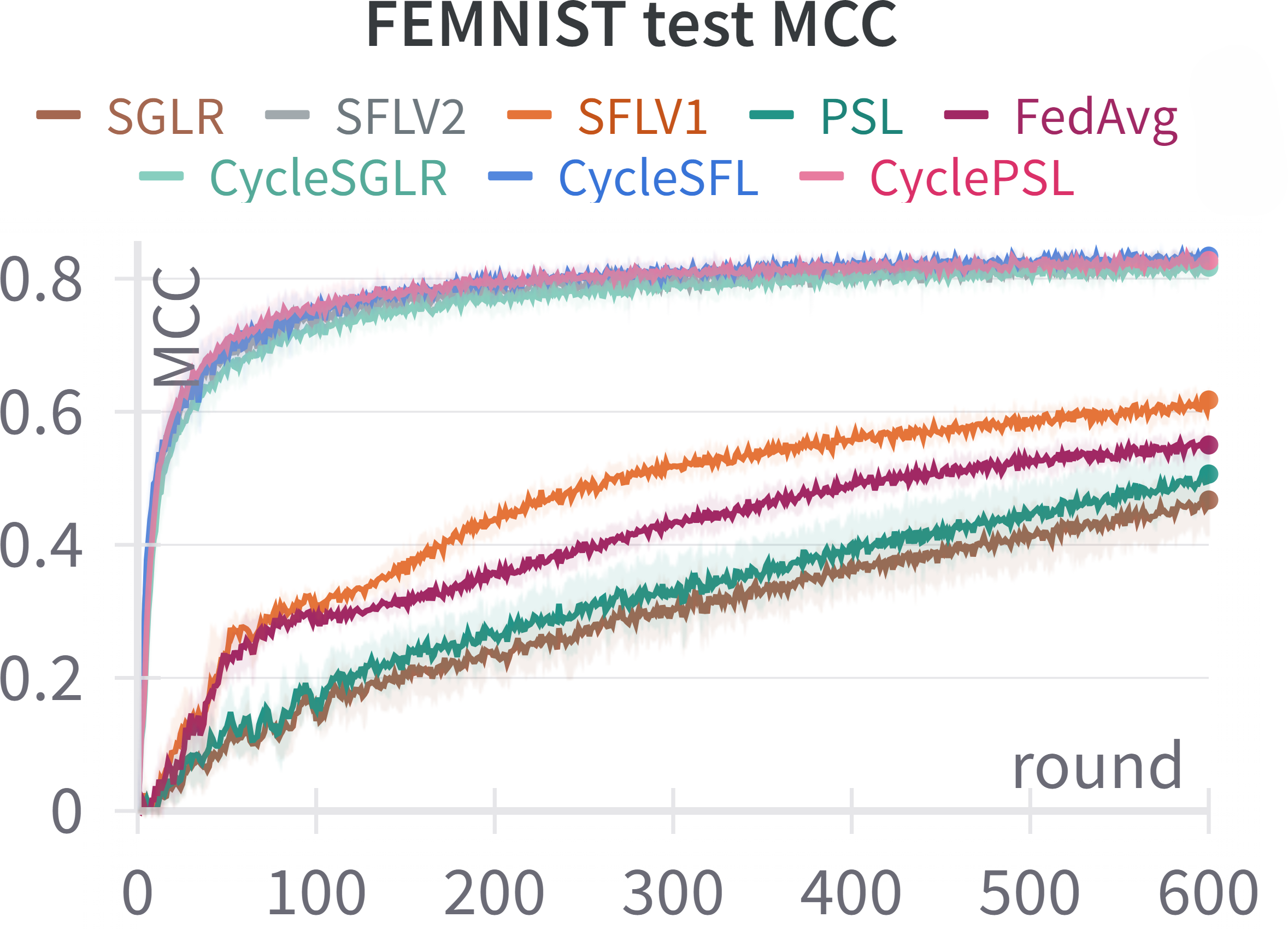}
    \label{fig:appendix_femnist_mcc}}
    \caption{Test metrics for the FEMNIST task.}
    \label{fig:appendix_femnist_result}
\end{figure}

\begin{figure}[!ht]
    \centering
    \subfloat{\includegraphics[height=\myheightsecond, keepaspectratio]{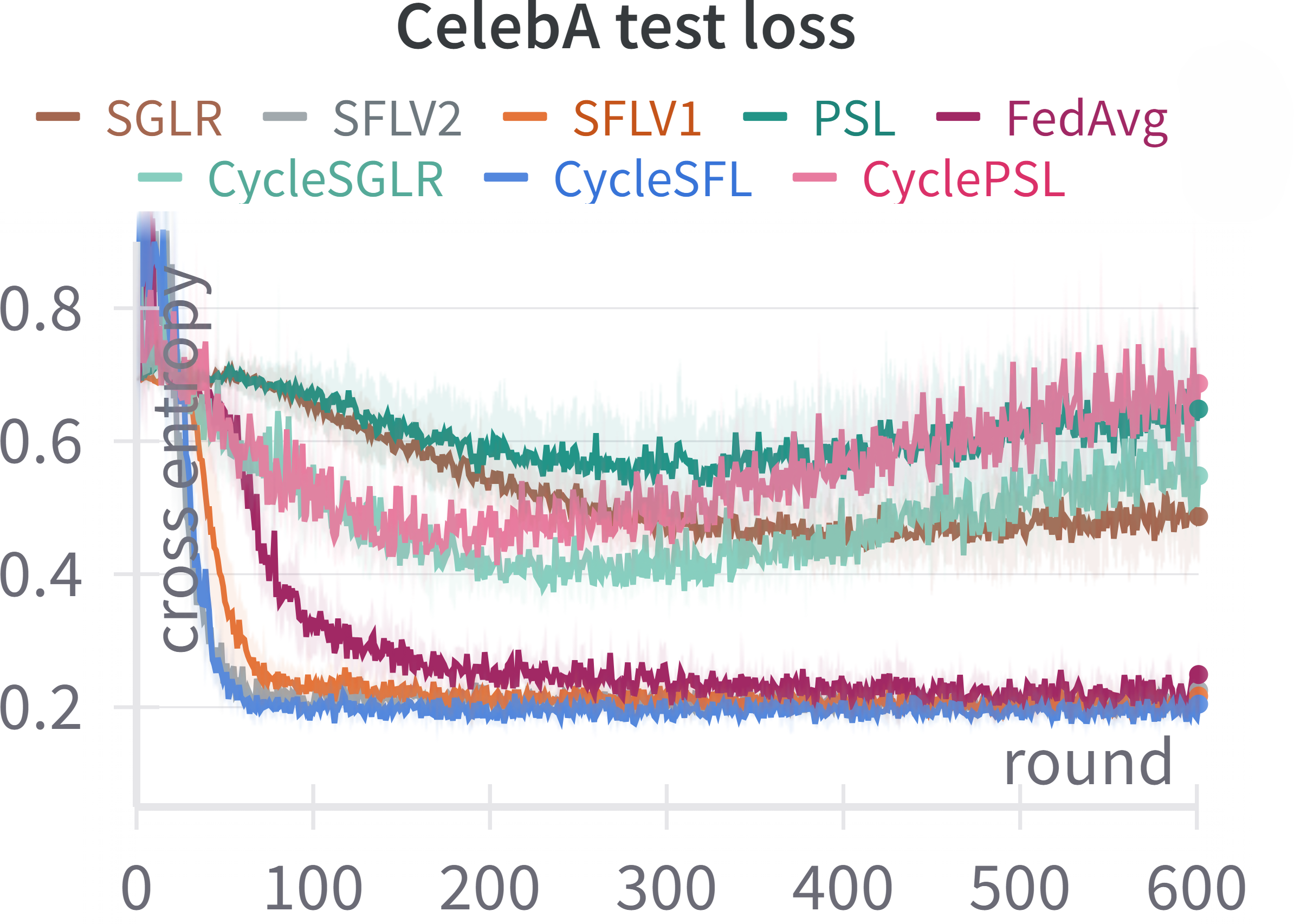}
    \label{fig:appendix_celeba_loss}}
    \hfill
    \subfloat{\includegraphics[height=\myheightsecond, keepaspectratio]{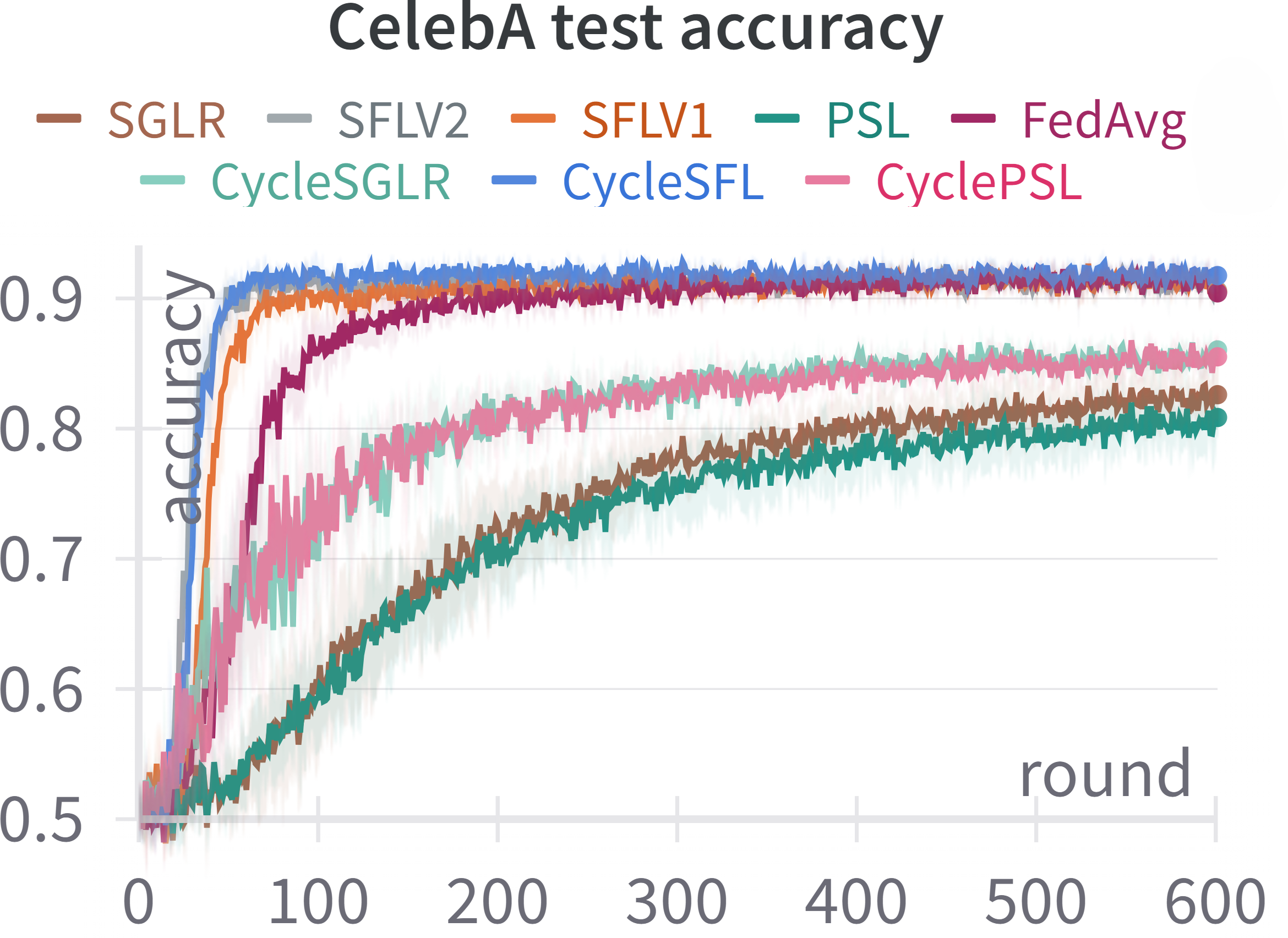}
    \label{fig:appendix_celeba_accu}}
    \hfill
    \subfloat{\includegraphics[height=\myheightsecond, keepaspectratio]{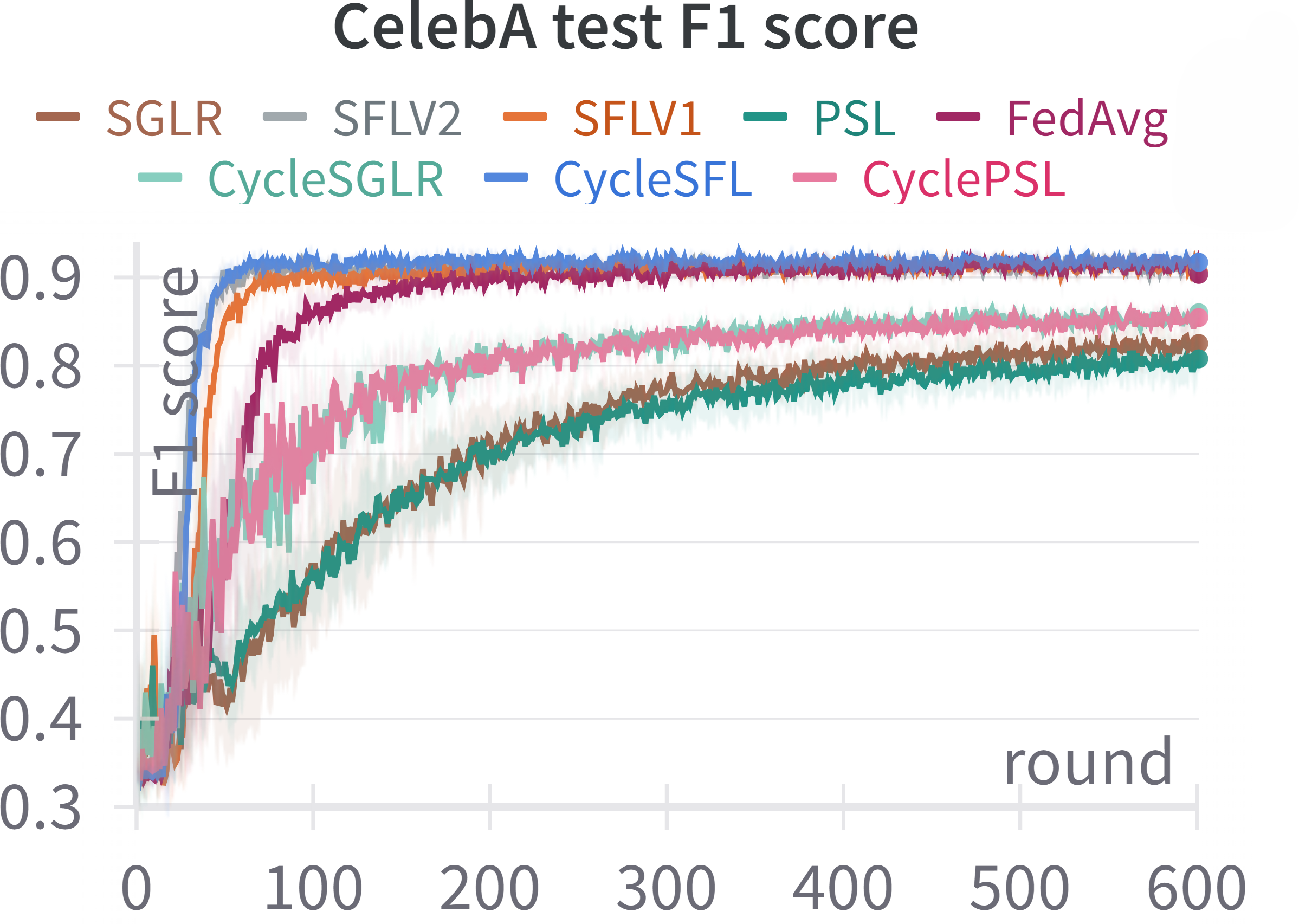}
    \label{fig:appendix_celeba_f1}}
    \hfill
    \subfloat{\includegraphics[height=\myheightsecond, keepaspectratio]{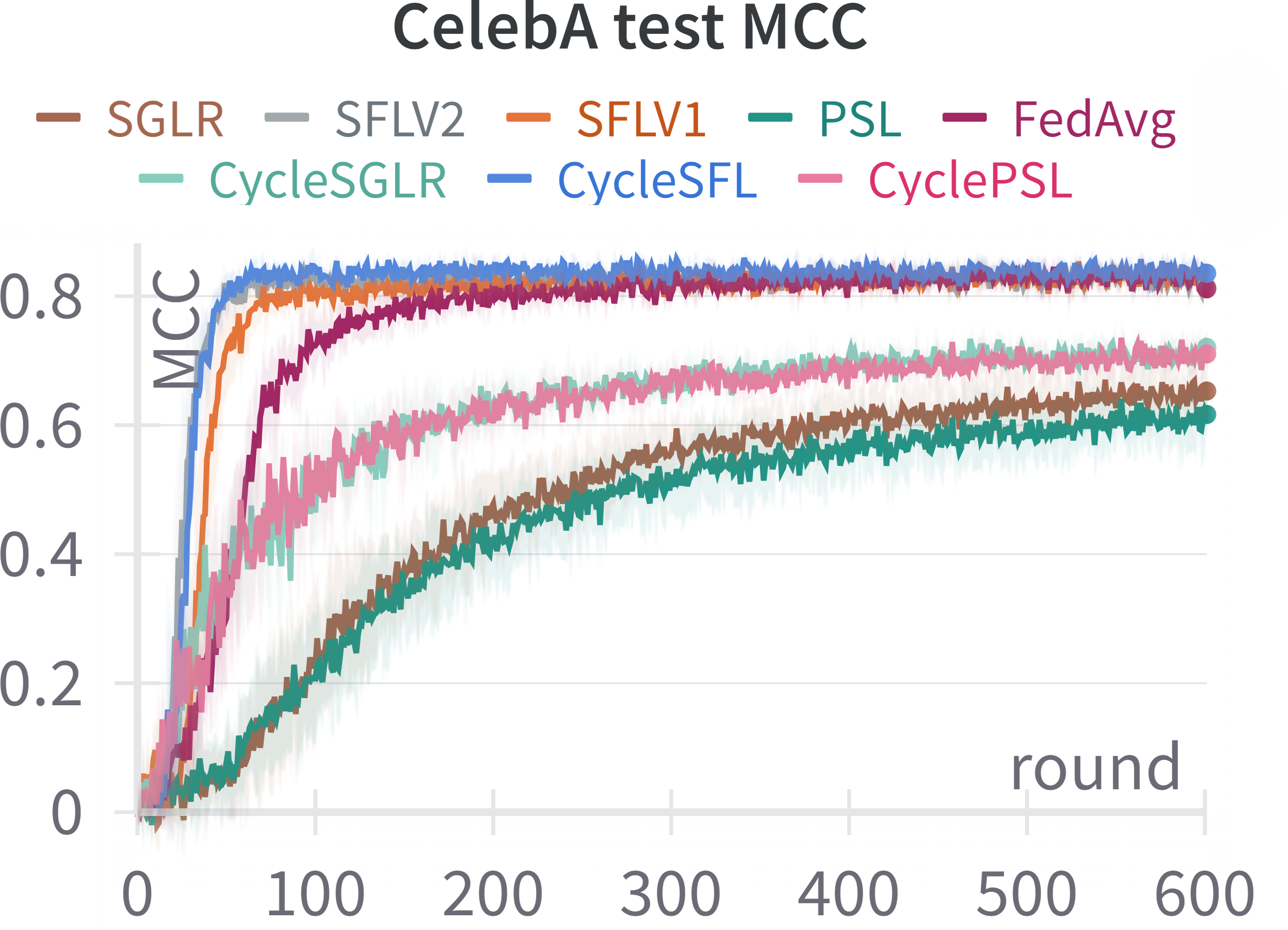}
    \label{fig:appendix_celeba_mcc}}
    \caption{Test metrics for the CelebA task.}
    \label{fig:appendix_celeba_result}
\end{figure}

\begin{figure}[!ht]
    \centering
    \subfloat{\includegraphics[height=\myheightsecond, keepaspectratio]{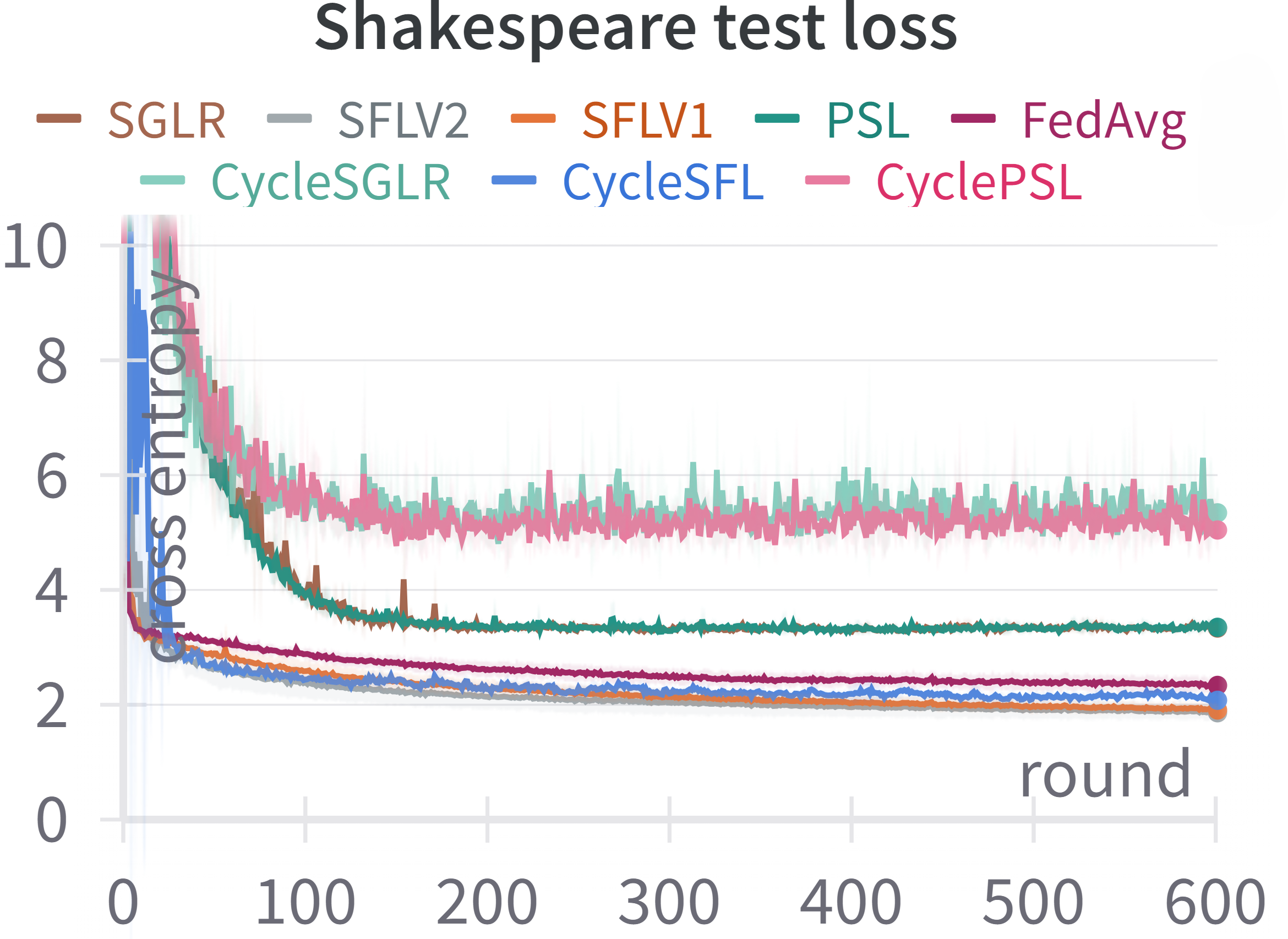}
    \label{fig:appendix_shakespeare_loss}}
    \hfill
    \subfloat{\includegraphics[height=\myheightsecond, keepaspectratio]{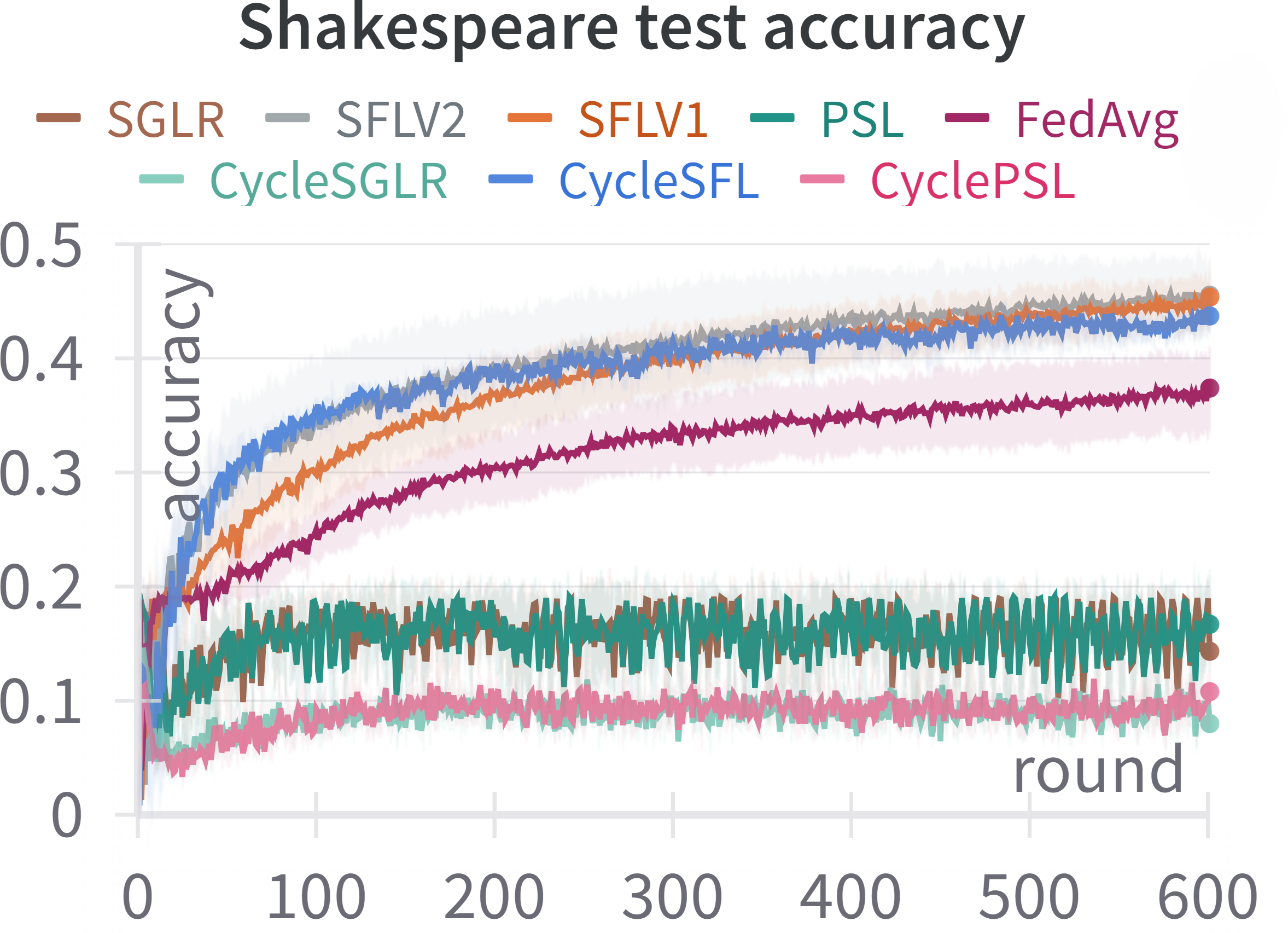}
    \label{fig:appendix_shakespeare_accu}}
    \hfill
    \subfloat{\includegraphics[height=\myheightsecond, keepaspectratio]{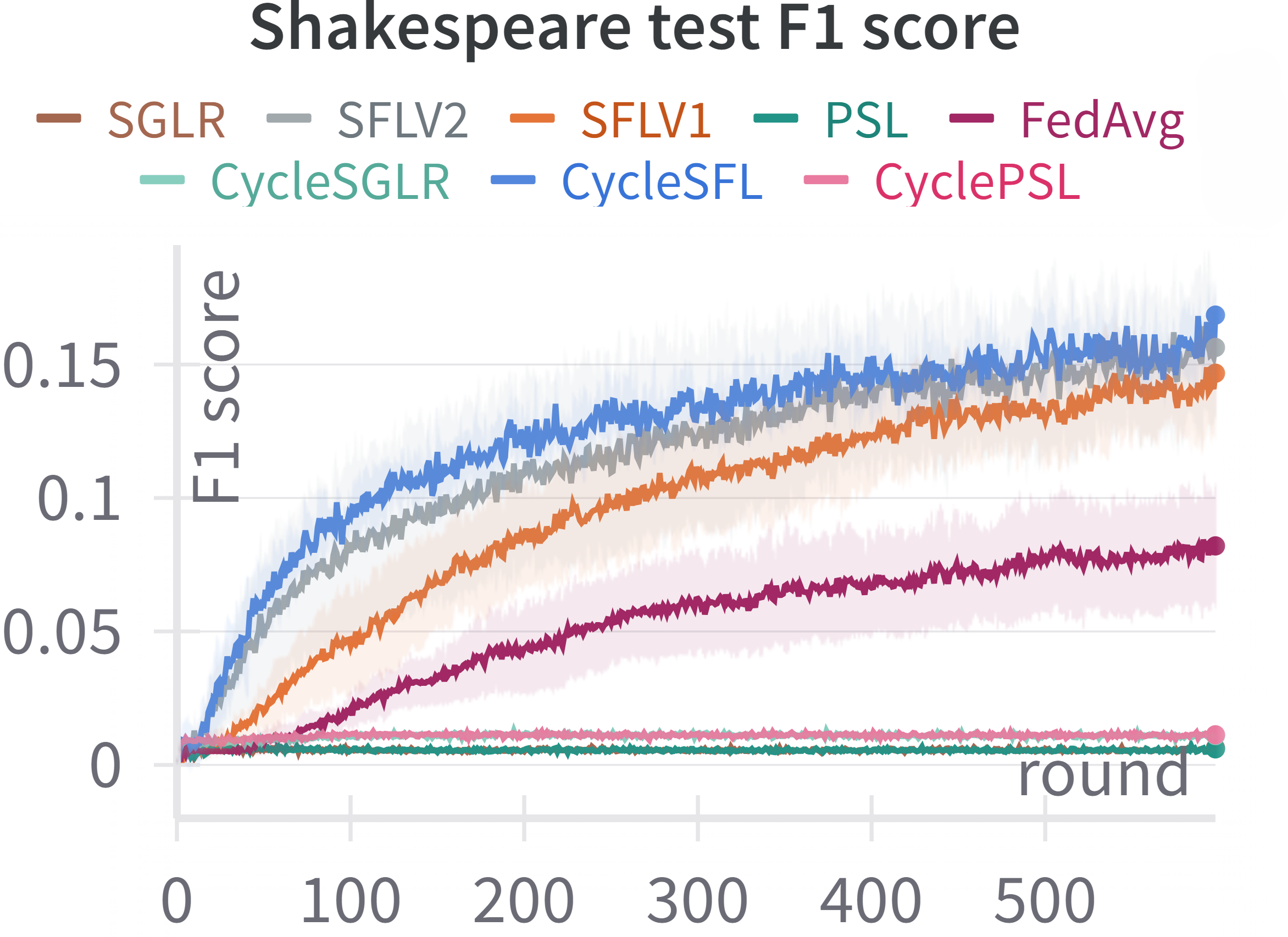}
    \label{fig:appendix_shakespeare_f1}}
    \hfill
    \subfloat{\includegraphics[height=\myheightsecond, keepaspectratio]{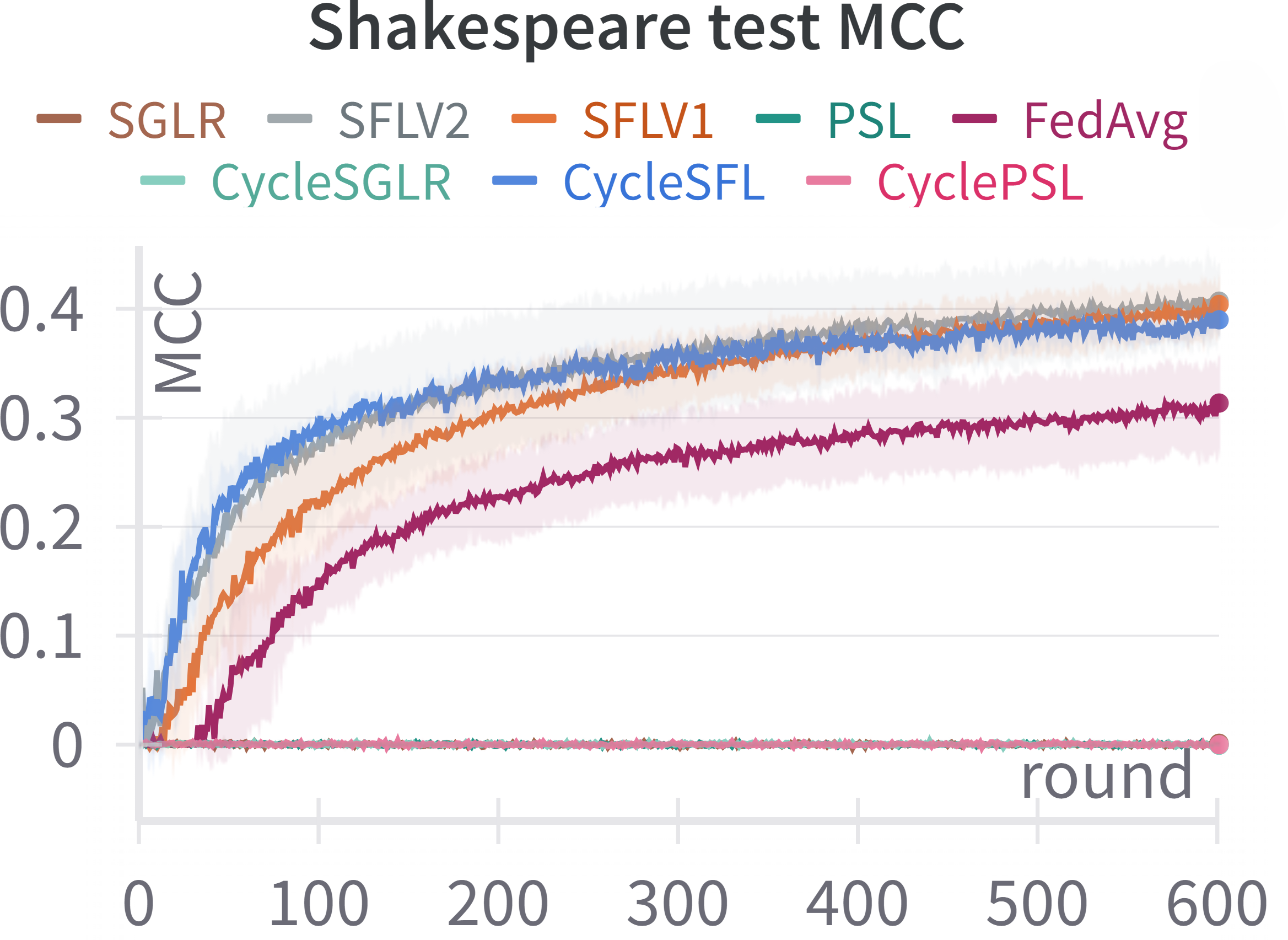}
    \label{fig:appendix_shakespeare_mcc}}
    \caption{Test metrics for the Shakespeare task.}
    \label{fig:appendix_shakespeare_result}
\end{figure}

\begin{figure}[!ht]
    \centering
    \subfloat{\includegraphics[height=\myheightsecond, keepaspectratio]{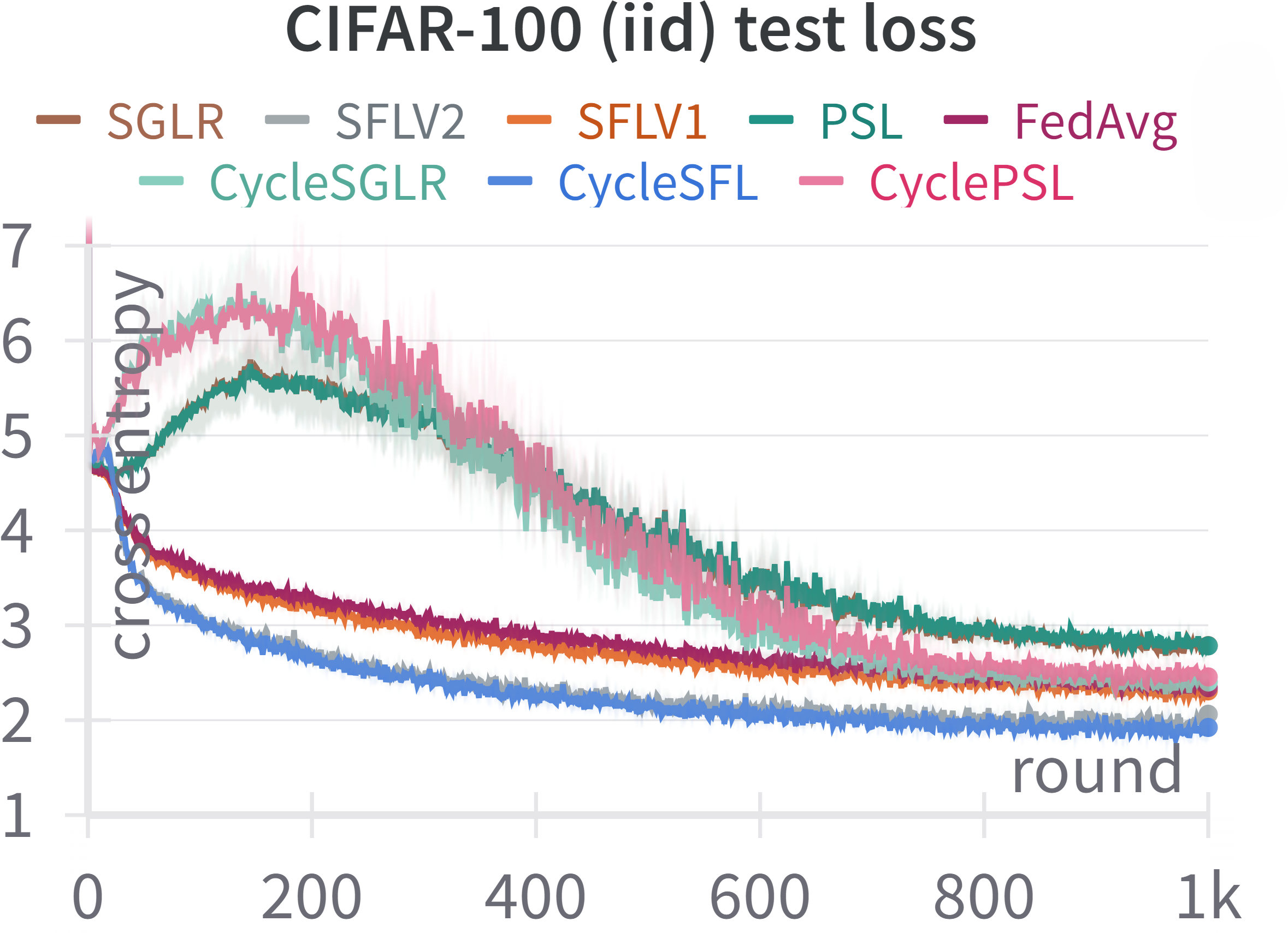}
    \label{fig:appendix_cifar100_iid_loss}}
    \hfill
    \subfloat{\includegraphics[height=\myheightsecond, keepaspectratio]{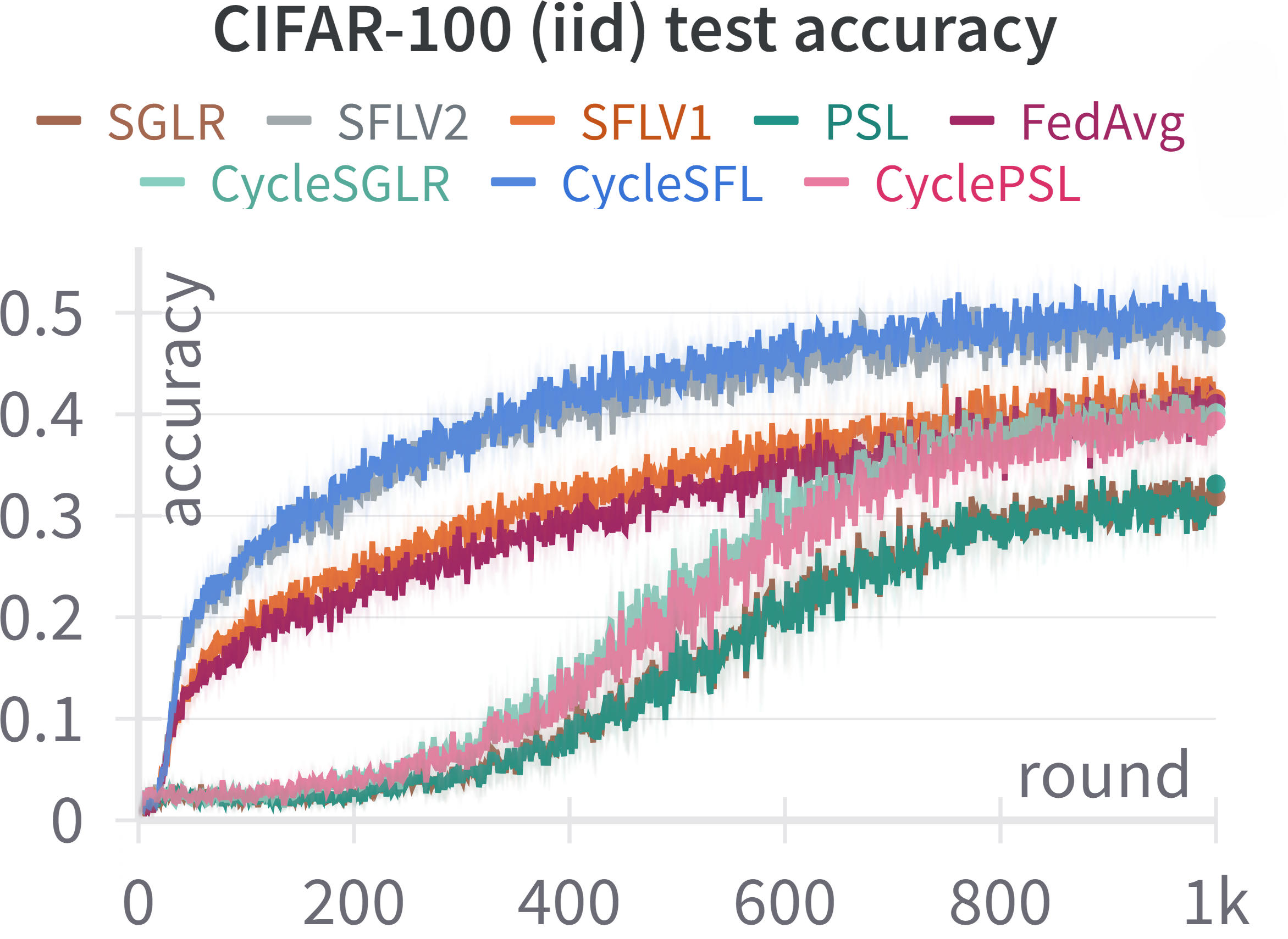}
    \label{fig:appendix_cifar100_iid_accu}}
    \hfill
    \subfloat{\includegraphics[height=\myheightsecond, keepaspectratio]{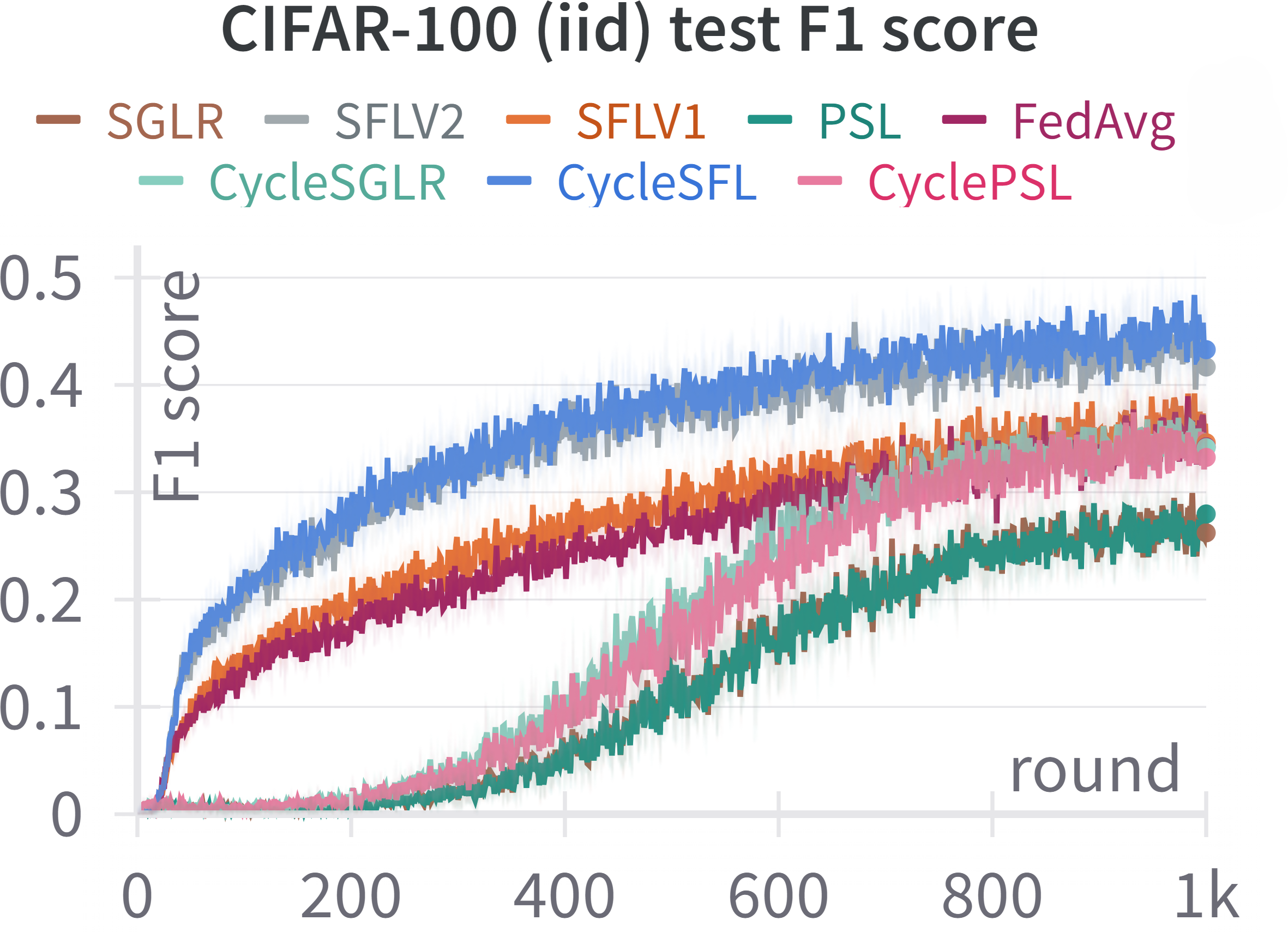}
    \label{fig:appendix_cifar100_iid_f1}}
    \hfill
    \subfloat{\includegraphics[height=\myheightsecond, keepaspectratio]{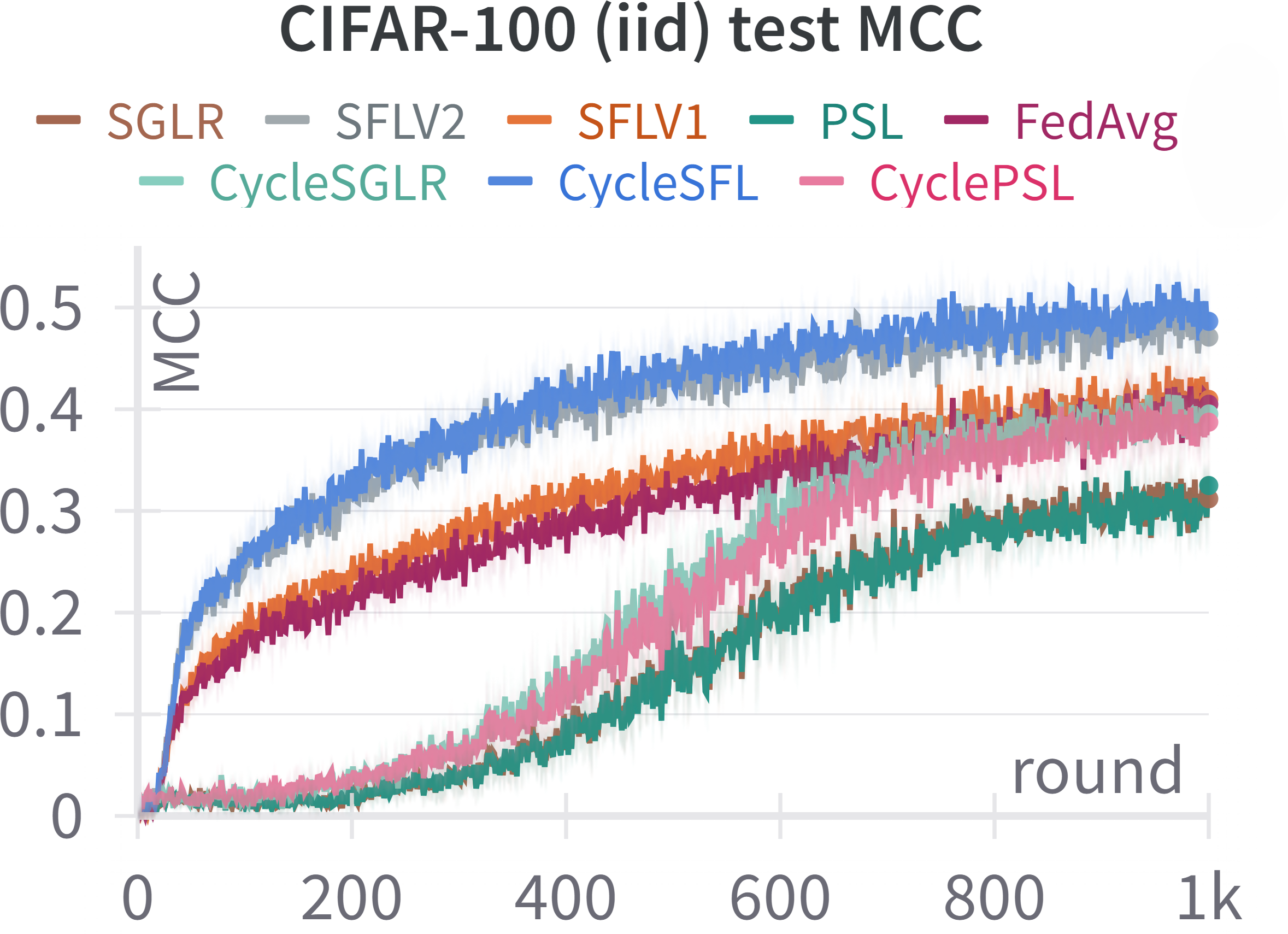}
    \label{fig:appendix_cifar100_iid_mcc}}
    \caption{Test metrics for the CIFAR-100 task (iid).}
    \label{fig:appendix_cifar100_iid_result}
\end{figure}

\begin{figure}[!ht]
    \centering
    \subfloat{\includegraphics[height=\myheightsecond, keepaspectratio]{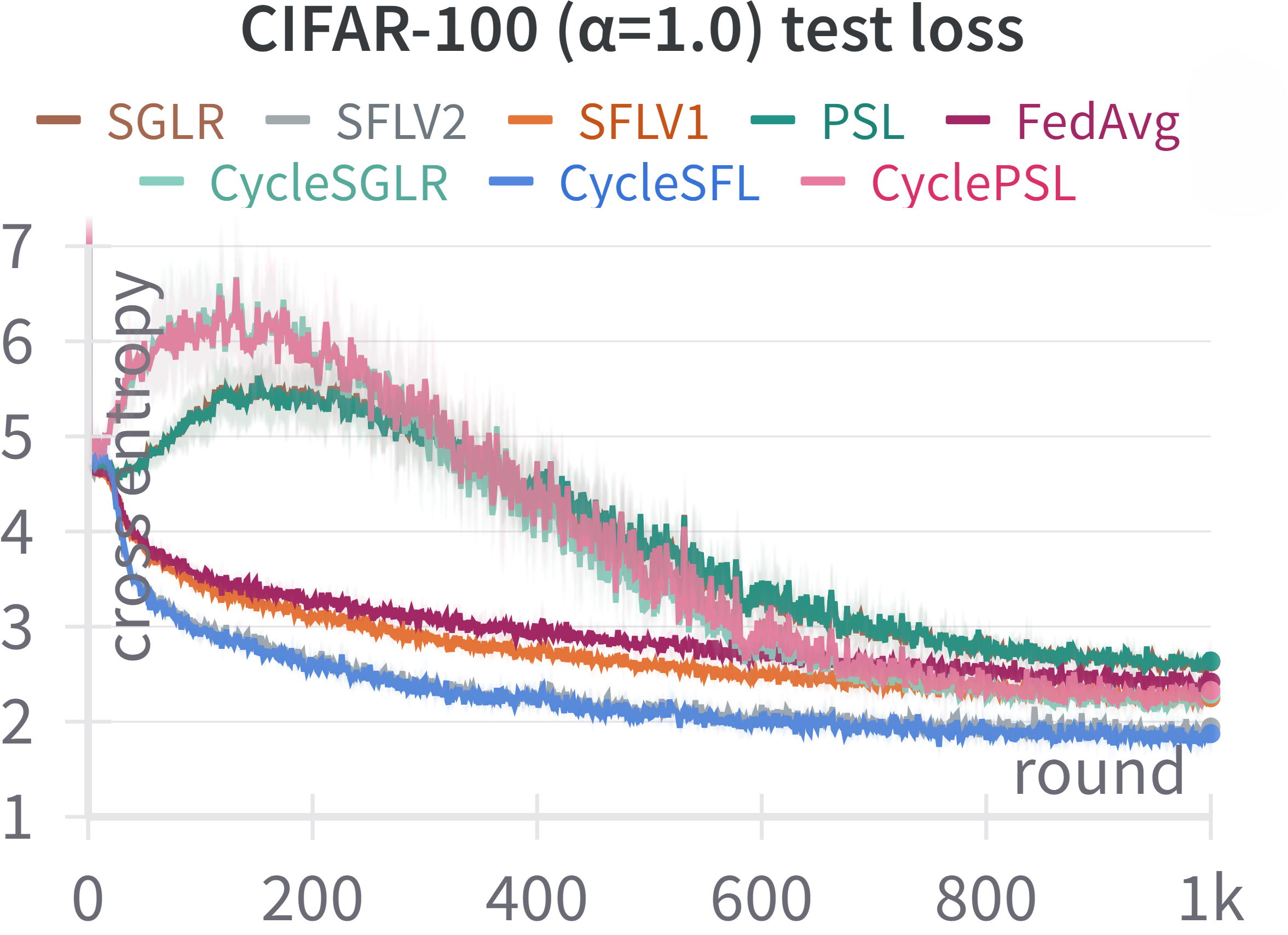}
    \label{fig:appendix_cifar100_1.0_loss}}
    \hfill
    \subfloat{\includegraphics[height=\myheightsecond, keepaspectratio]{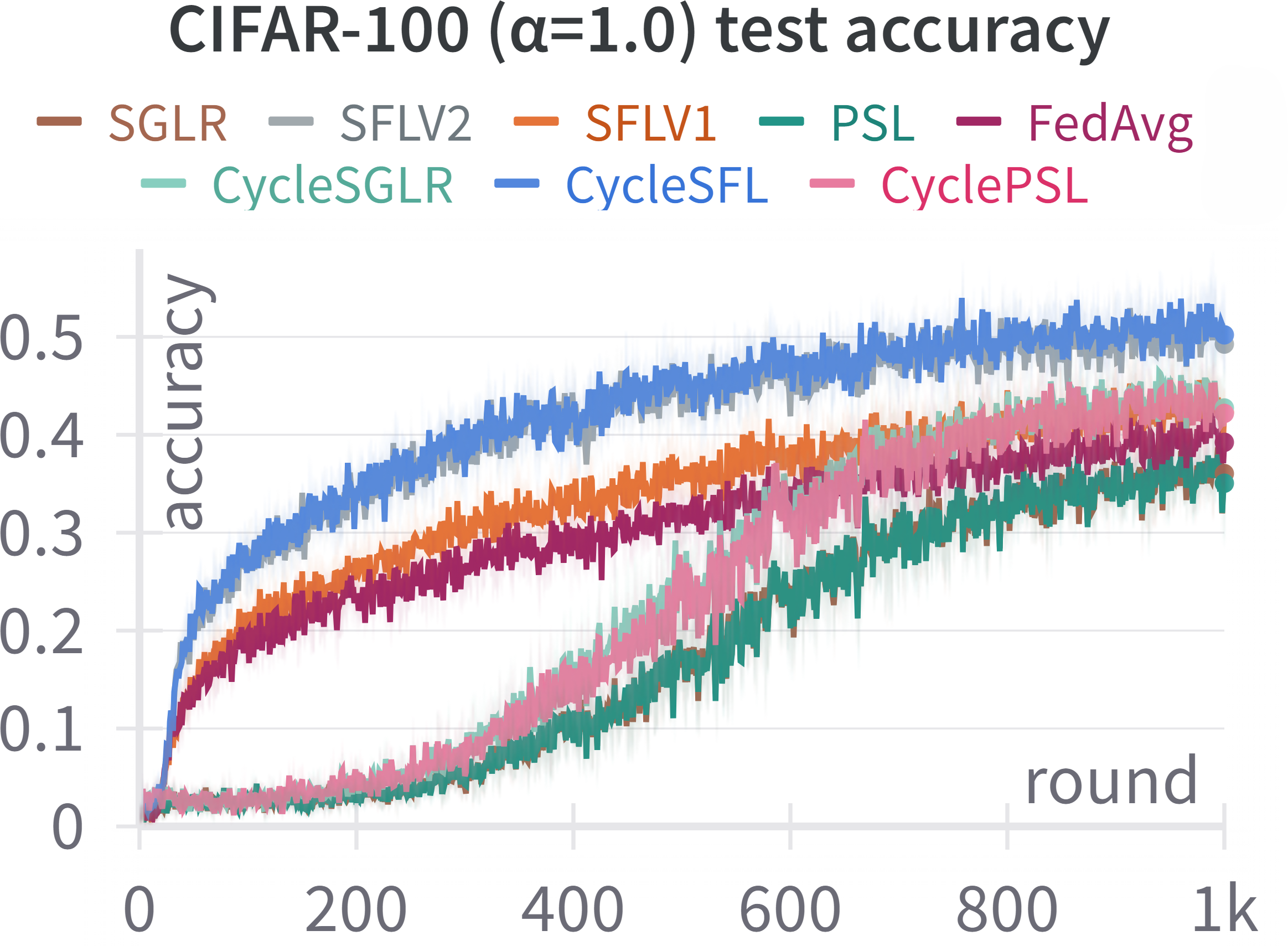}
    \label{fig:appendix_cifar100_1.0_accu}}
    \hfill
    \subfloat{\includegraphics[height=\myheightsecond, keepaspectratio]{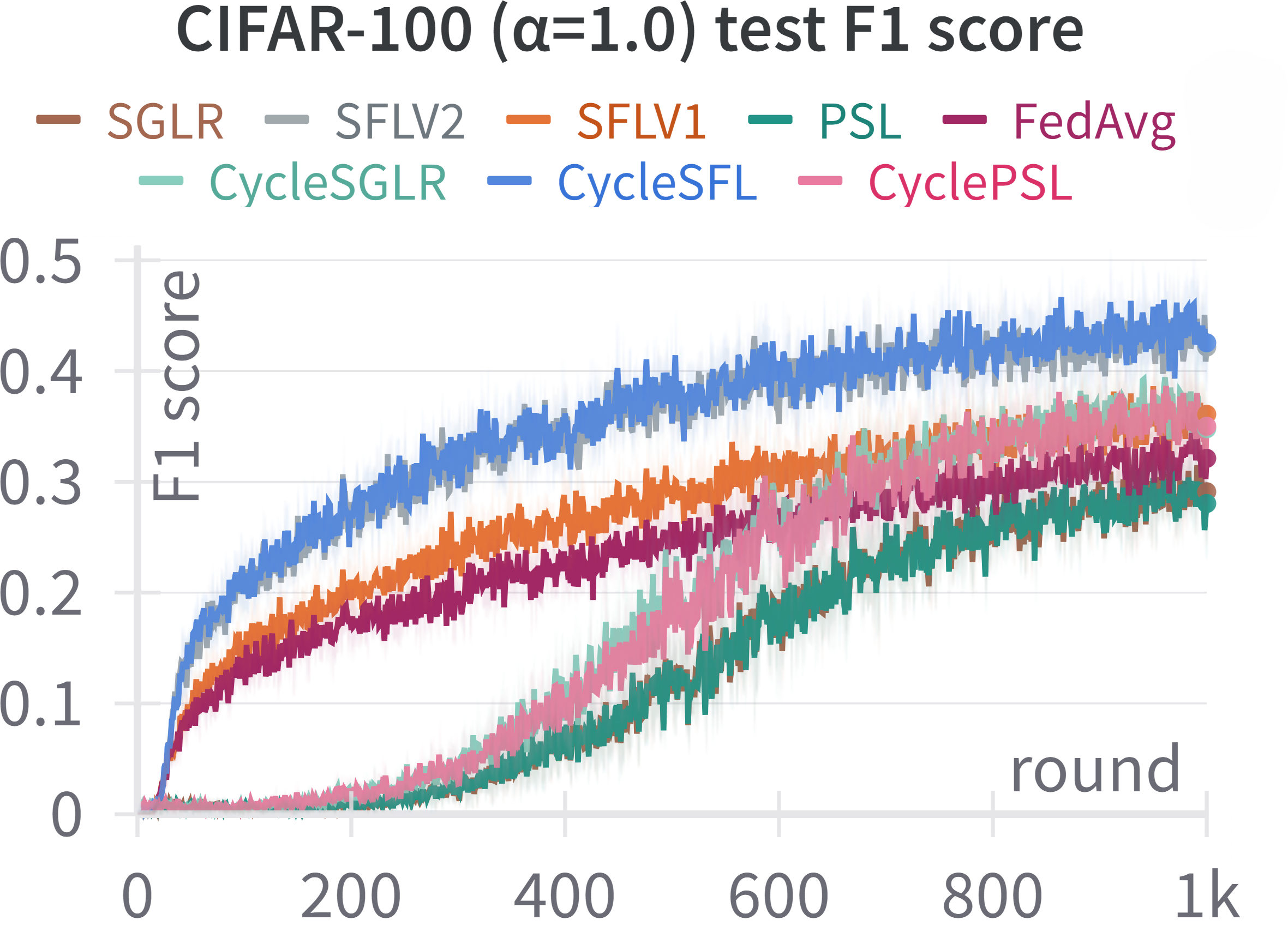}
    \label{fig:appendix_cifar100_1.0_f1}}
    \hfill
    \subfloat{\includegraphics[height=\myheightsecond, keepaspectratio]{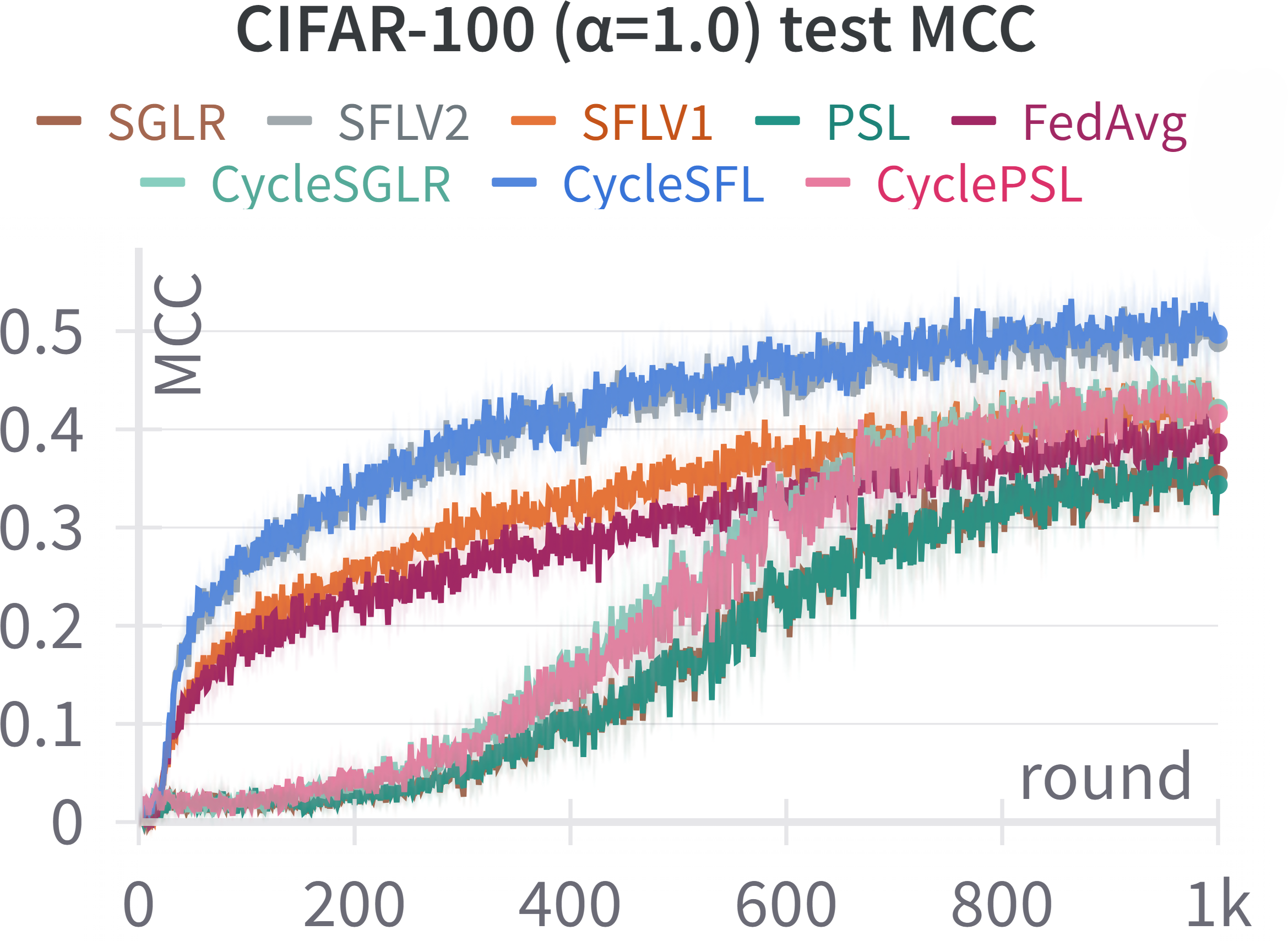}
    \label{fig:appendix_cifar100_1.0_mcc}}
    \caption{Test metrics for the CIFAR-100 task $(\alpha = 1.0)$.}
    \label{fig:appendix_cifar100_1.0_result}
\end{figure}

\begin{figure}[!ht]
    \centering
    \subfloat{\includegraphics[height=\myheightsecond, keepaspectratio]{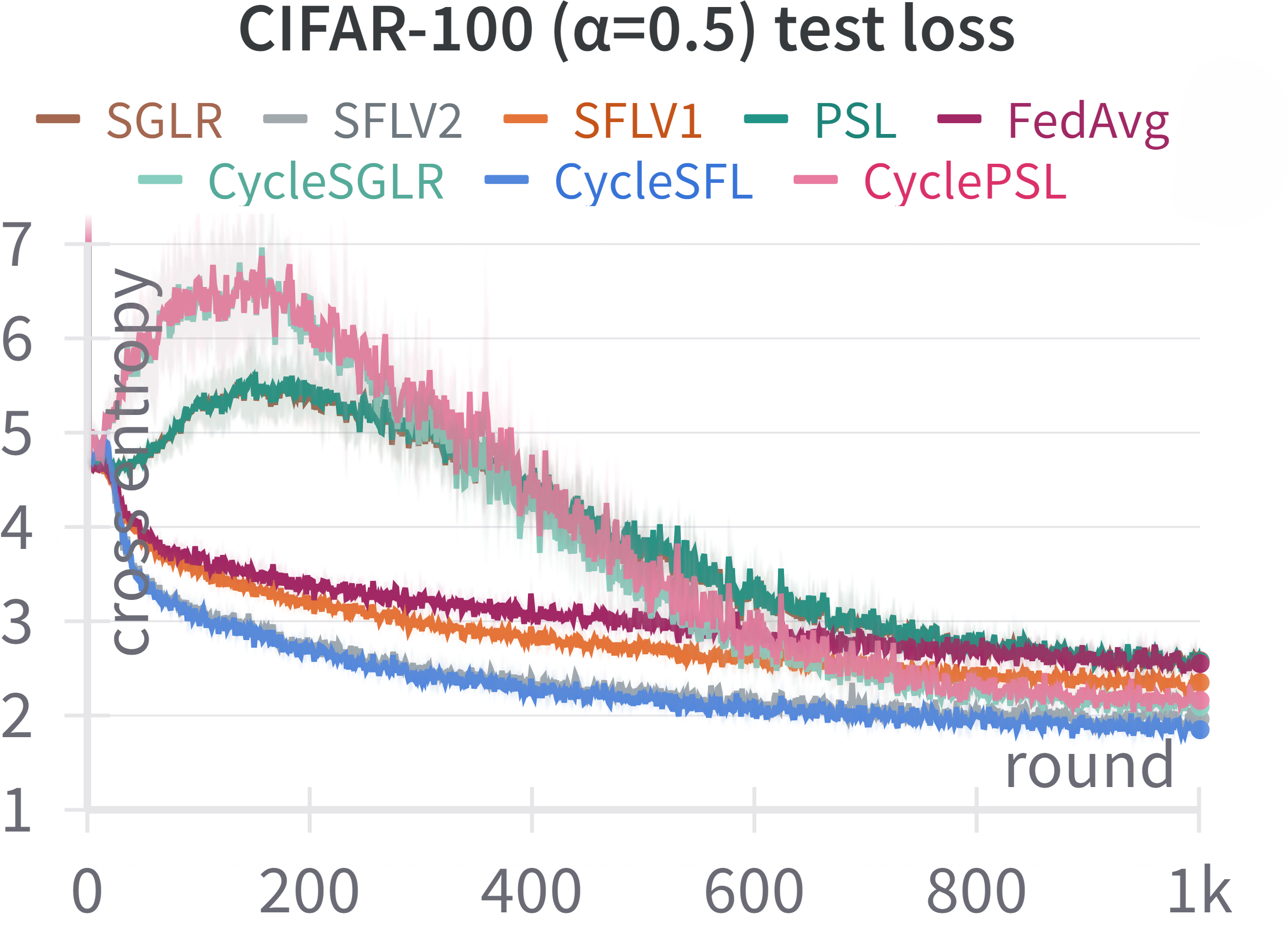}
    \label{fig:appendix_cifar100_0.5_loss}}
    \hfill
    \subfloat{\includegraphics[height=\myheightsecond, keepaspectratio]{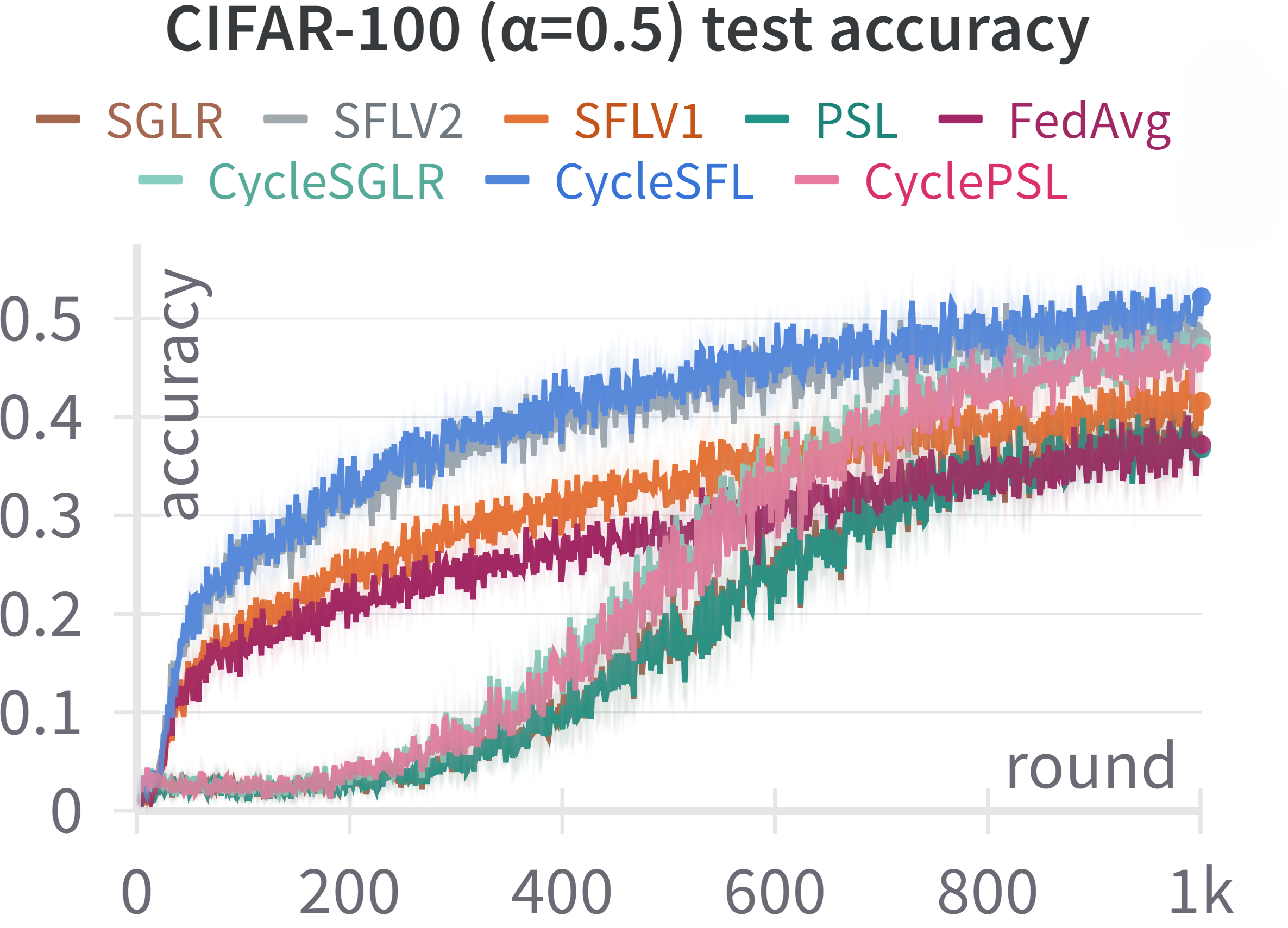}
    \label{fig:appendix_cifar100_0.5_accu}}
    \hfill
    \subfloat{\includegraphics[height=\myheightsecond, keepaspectratio]{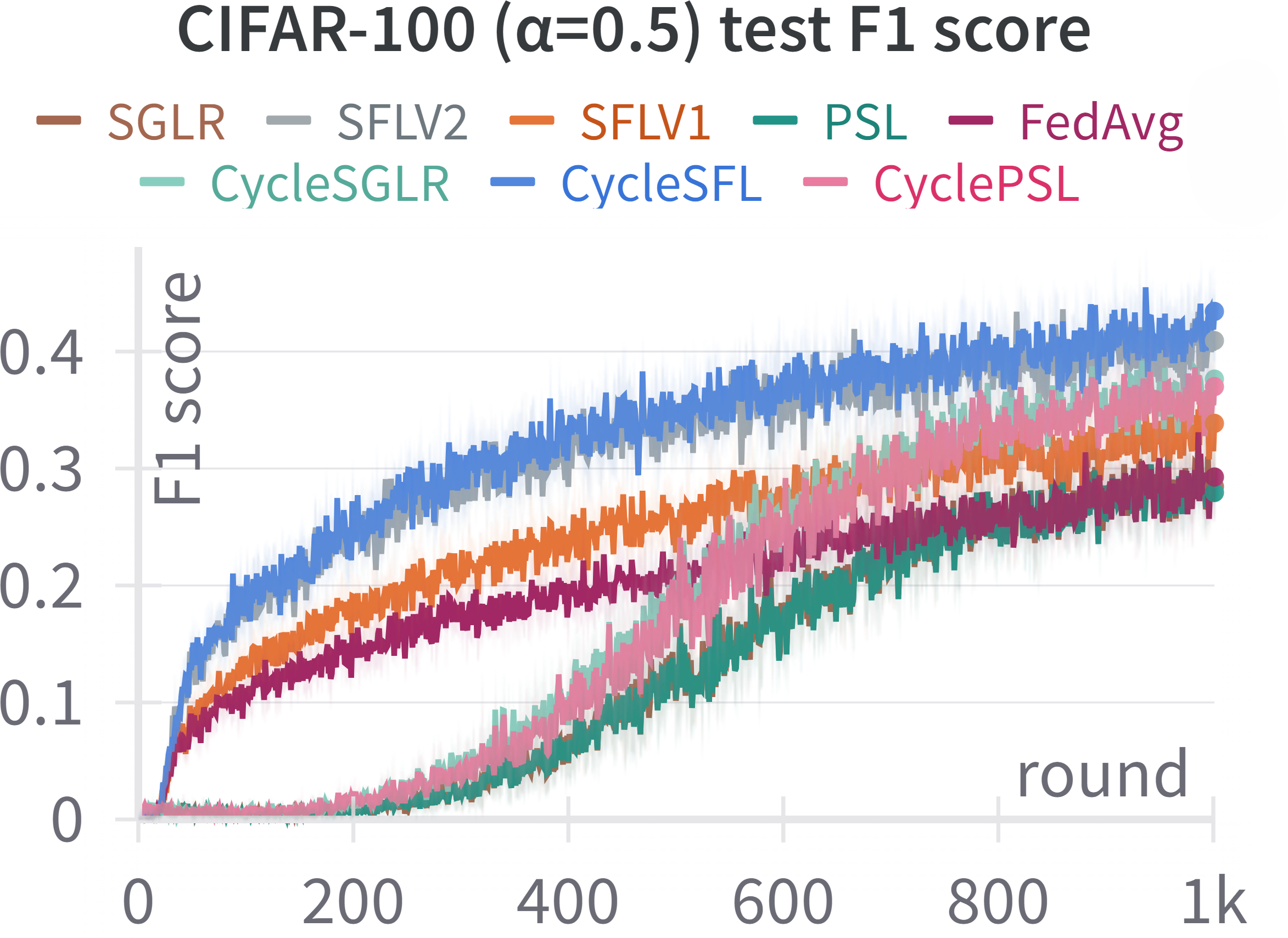}
    \label{fig:appendix_cifar100_0.5_f1}}
    \hfill
    \subfloat{\includegraphics[height=\myheightsecond, keepaspectratio]{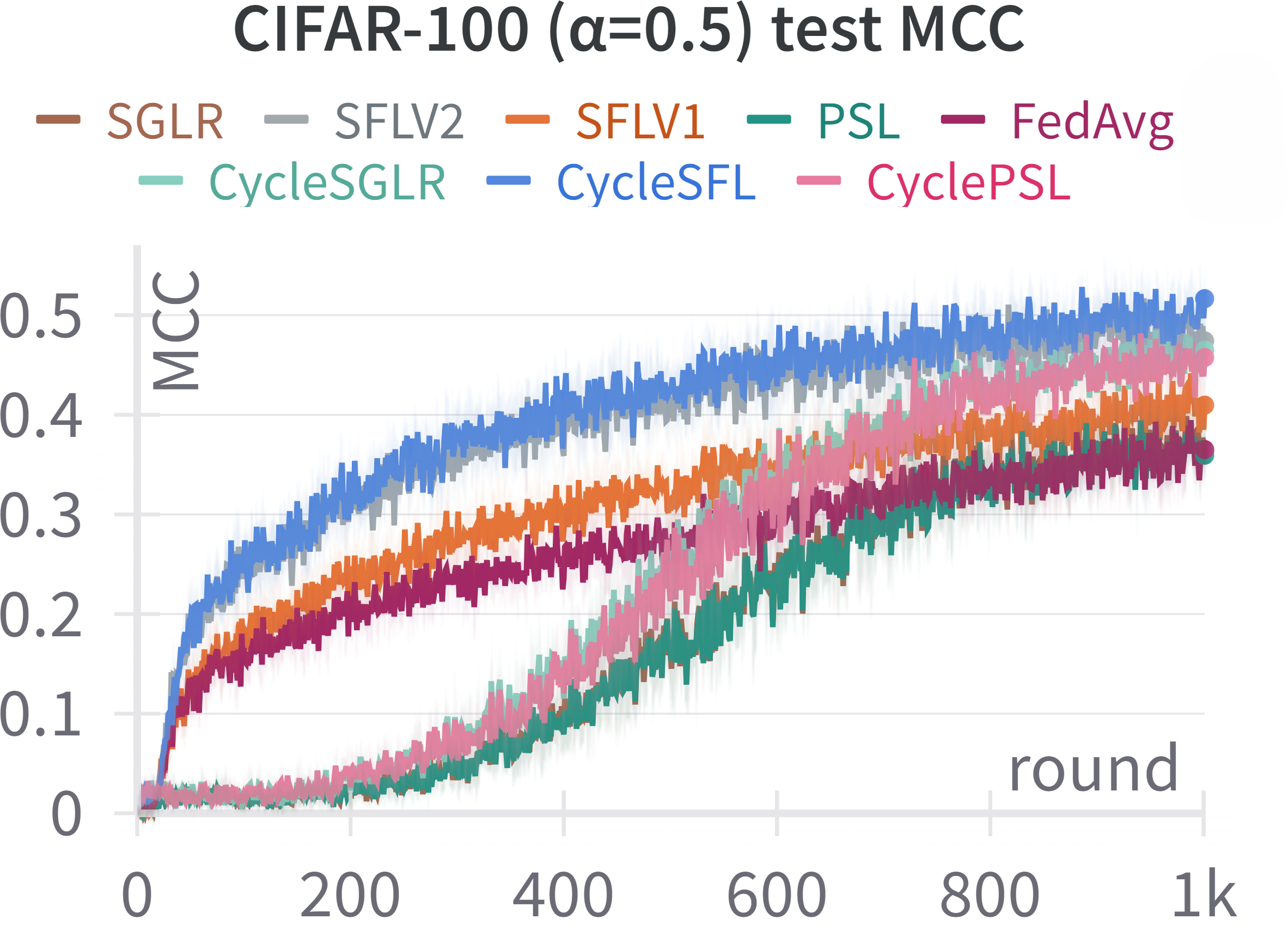}
    \label{fig:appendix_cifar100_0.5_mcc}}
    \caption{Test metrics for the CIFAR-100 task $(\alpha = 0.5)$.}
    \label{fig:appendix_cifar100_0.5_result}
\end{figure}

\begin{figure}[!ht]
    \centering
    \subfloat{\includegraphics[height=\myheightsecond, keepaspectratio]{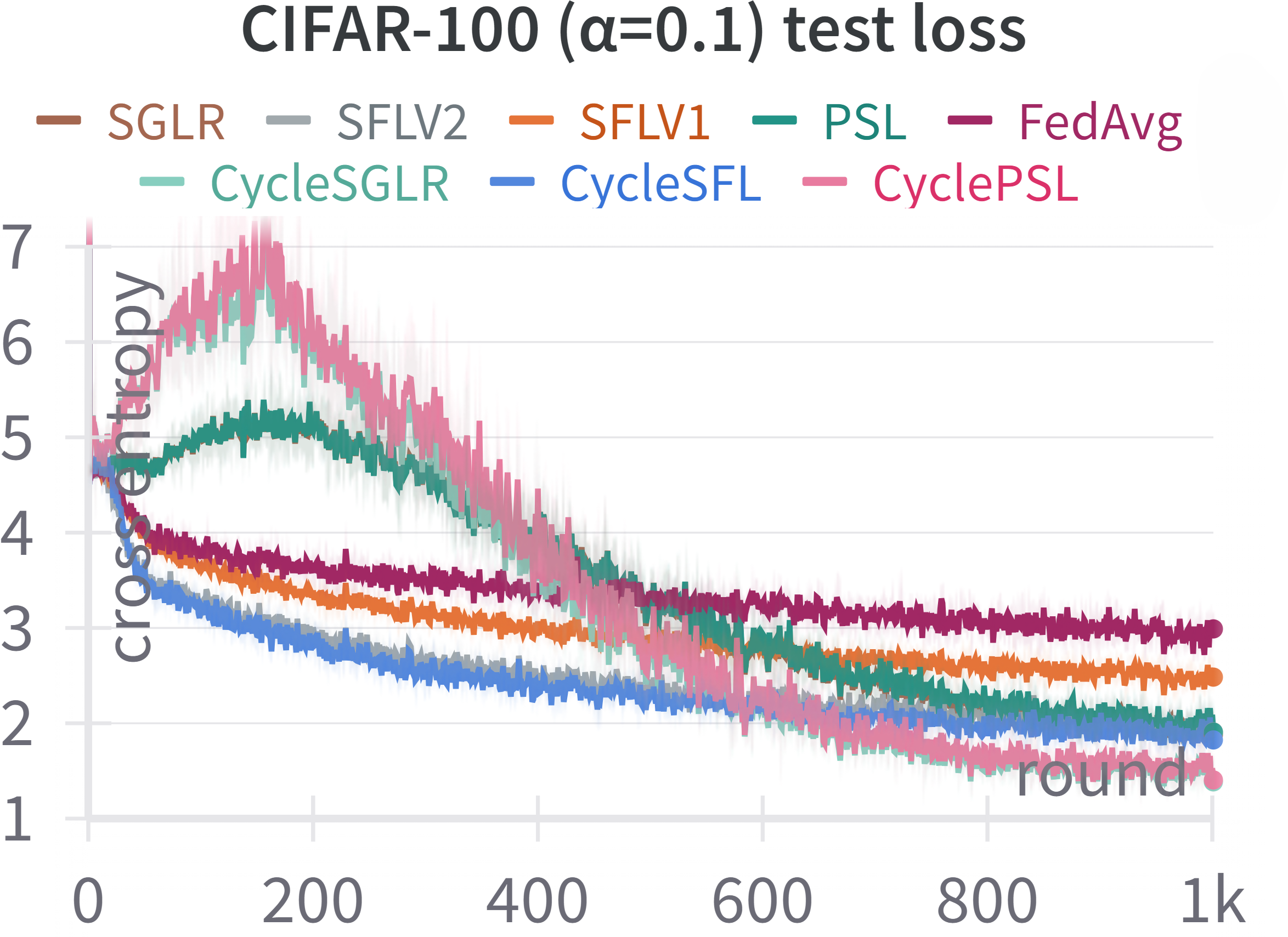}
    \label{fig:appendix_cifar100_0.1_loss}}
    \hfill
    \subfloat{\includegraphics[height=\myheightsecond, keepaspectratio]{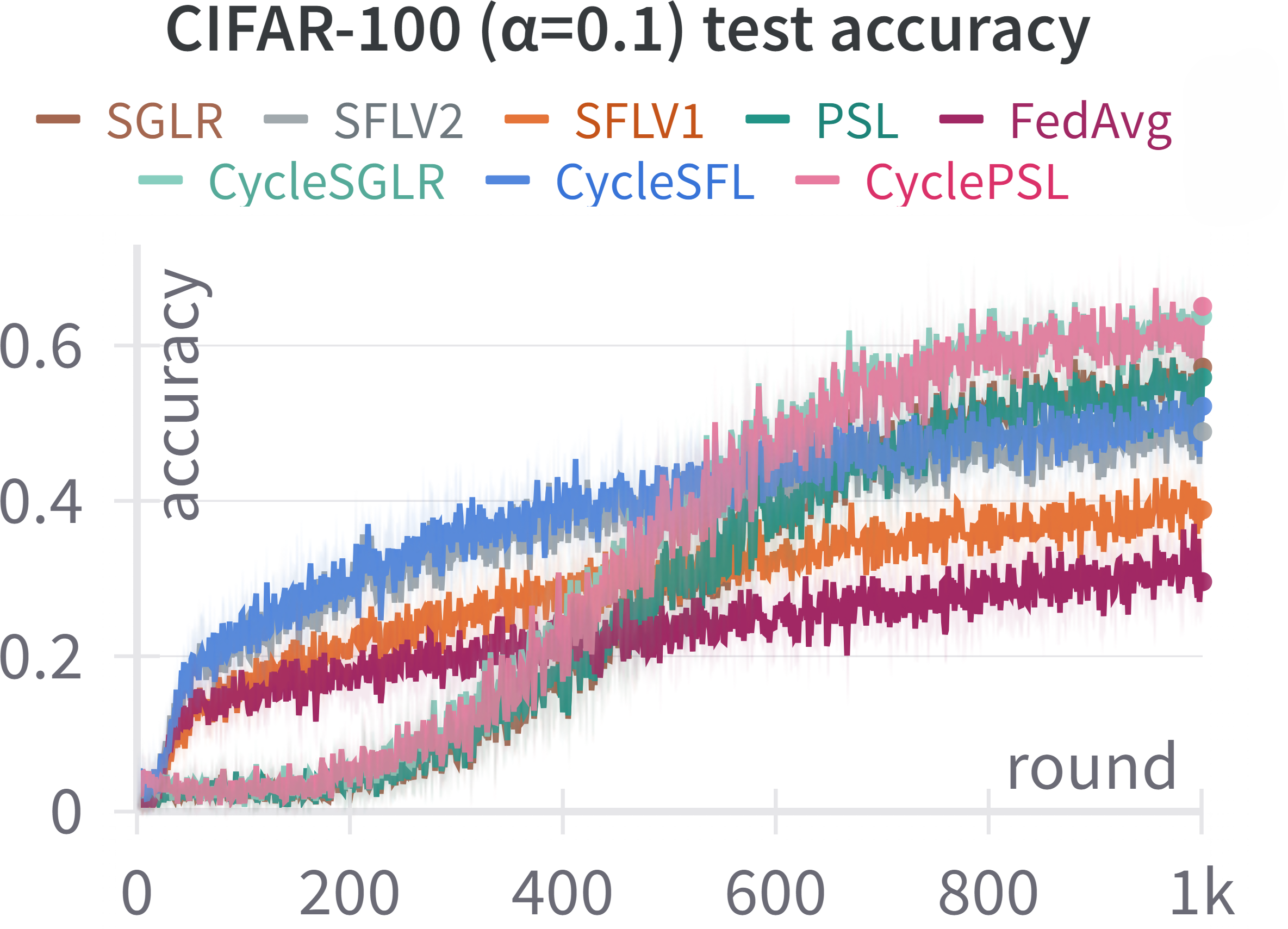}
    \label{fig:appendix_cifar100_0.1_accu}}
    \hfill
    \subfloat{\includegraphics[height=\myheightsecond, keepaspectratio]{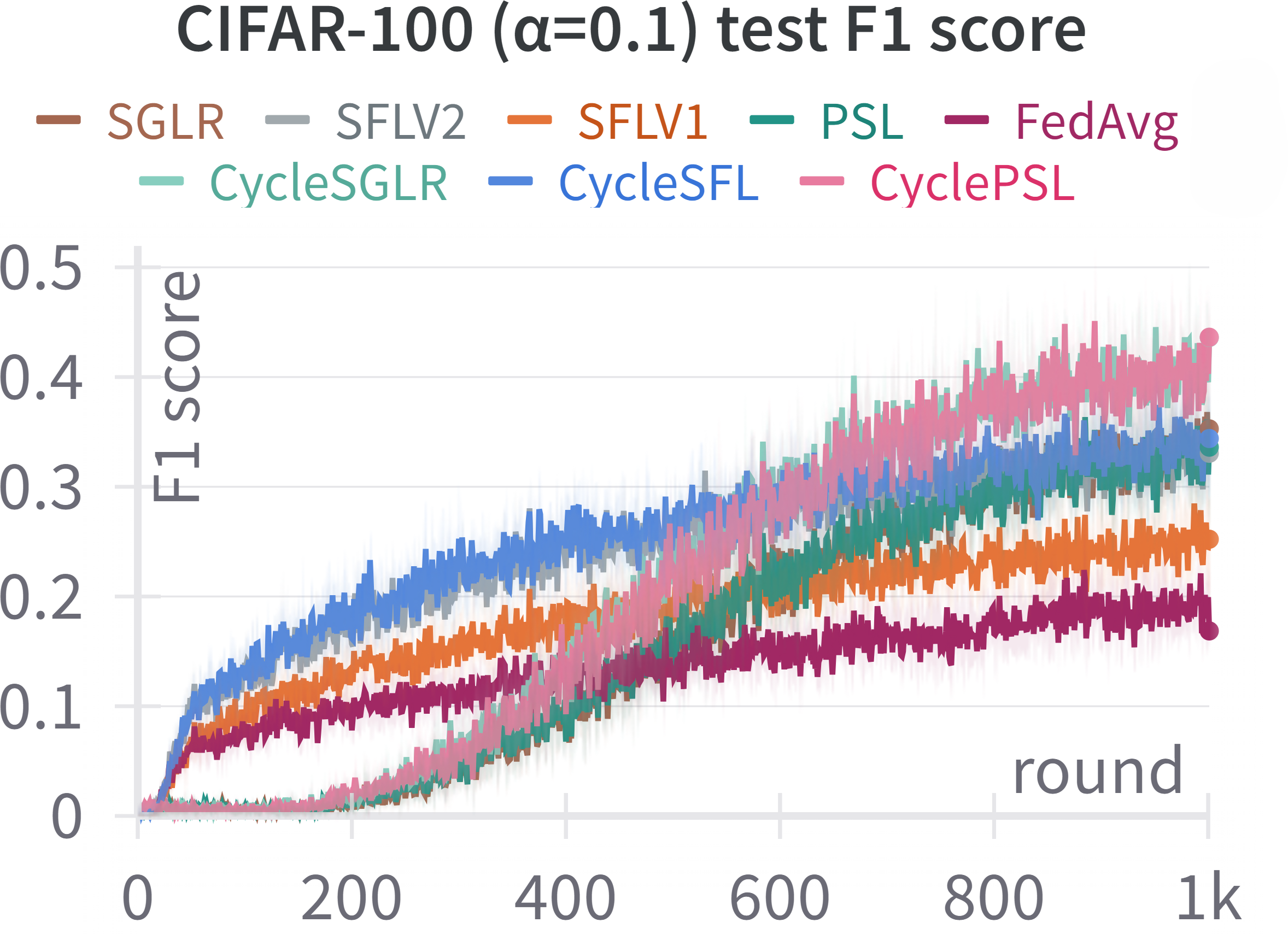}
    \label{fig:appendix_cifar100_0.1_f1}}
    \hfill
    \subfloat{\includegraphics[height=\myheightsecond, keepaspectratio]{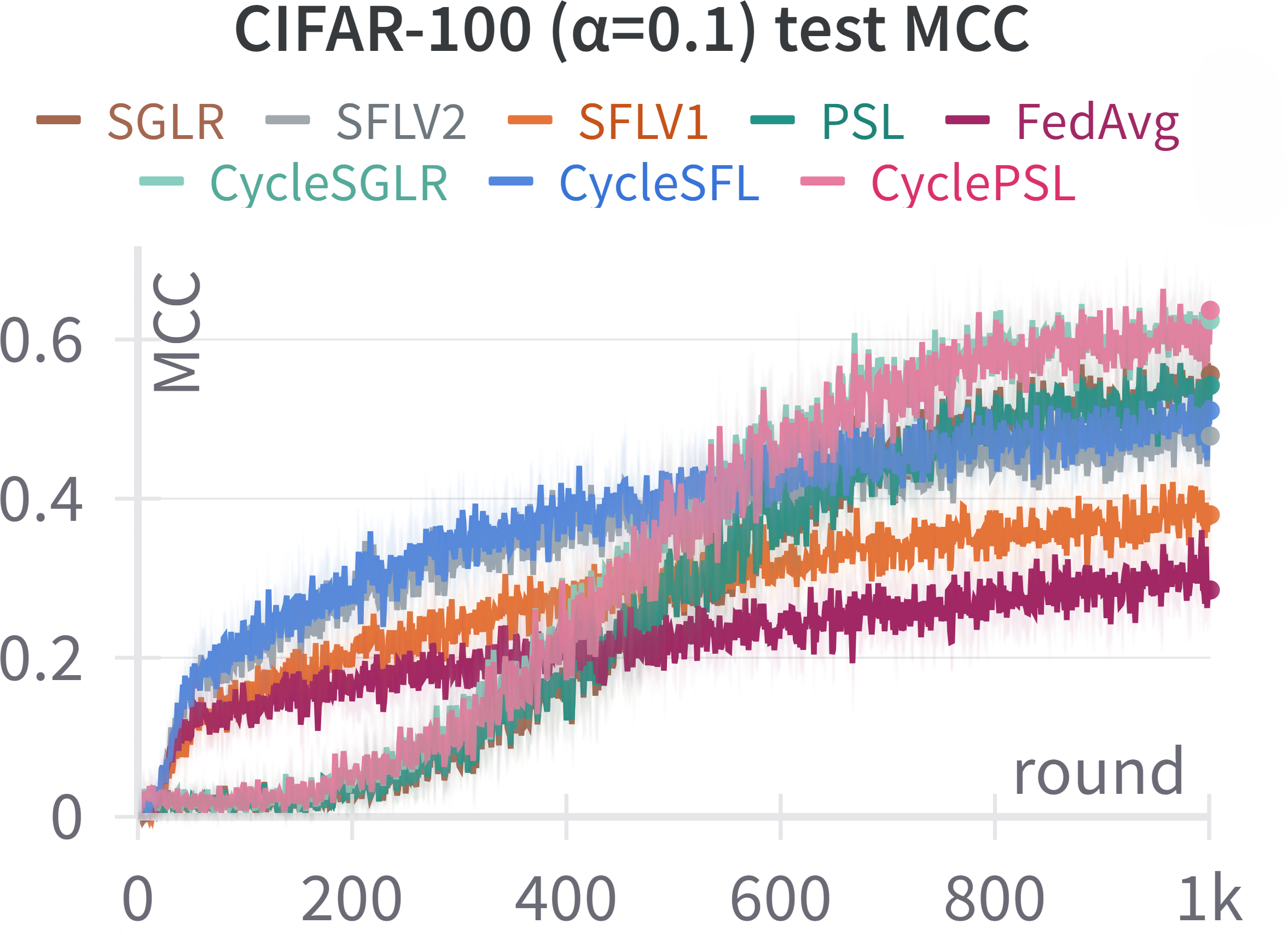}
    \label{fig:appendix_cifar100_0.1_mcc}}
    \caption{Test metrics for the CIFAR-100 task $(\alpha = 0.1)$.}
    \label{fig:appendix_cifar100_0.1_result}
\end{figure}

\begin{figure}[!ht]
    \centering
    \subfloat{\includegraphics[height=\myheightsecond, keepaspectratio]{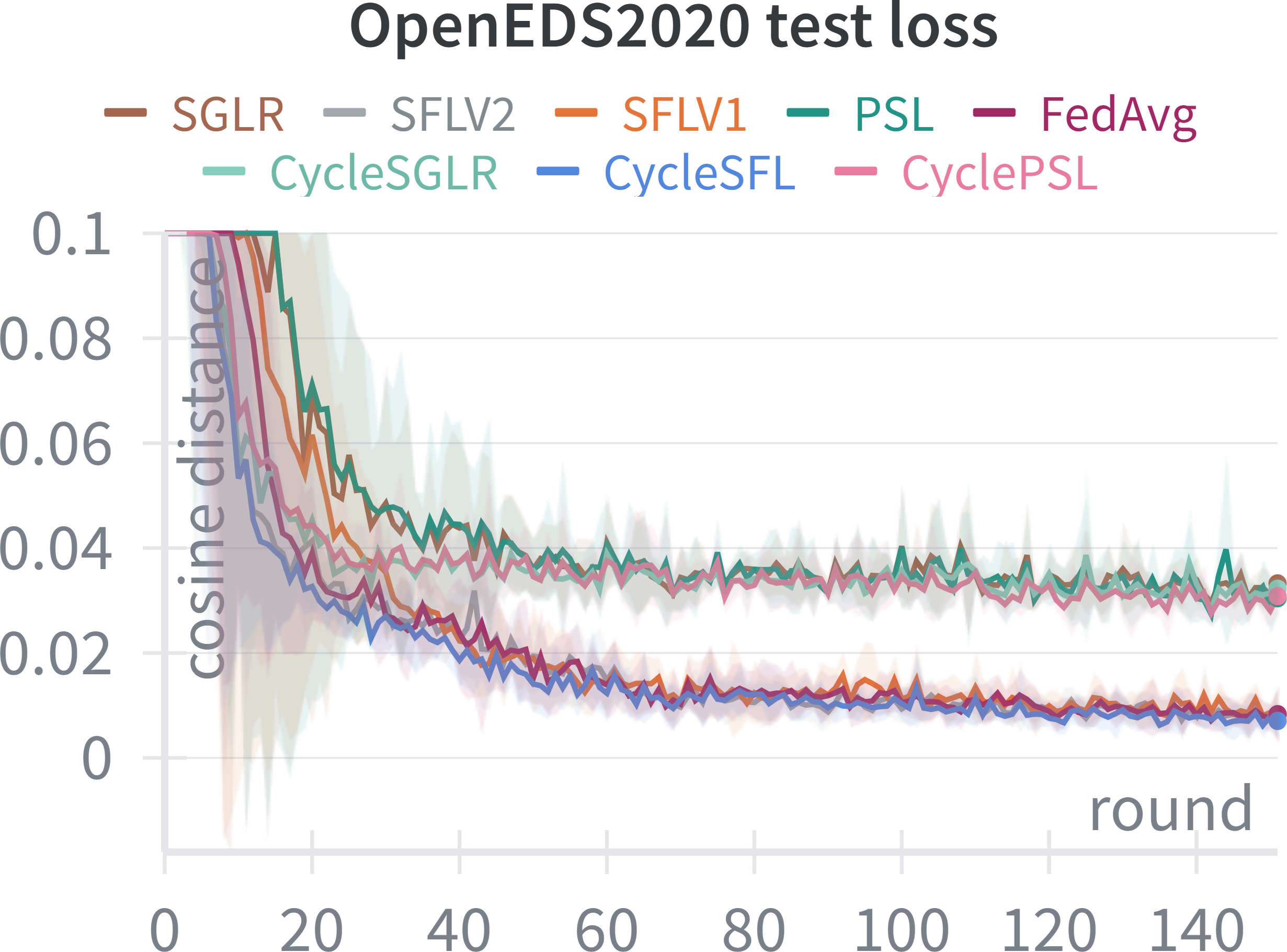}
    \label{fig:appendix_openeds_loss}}
    \hspace{1.0cm}
    \subfloat{\includegraphics[height=\myheightsecond, keepaspectratio]{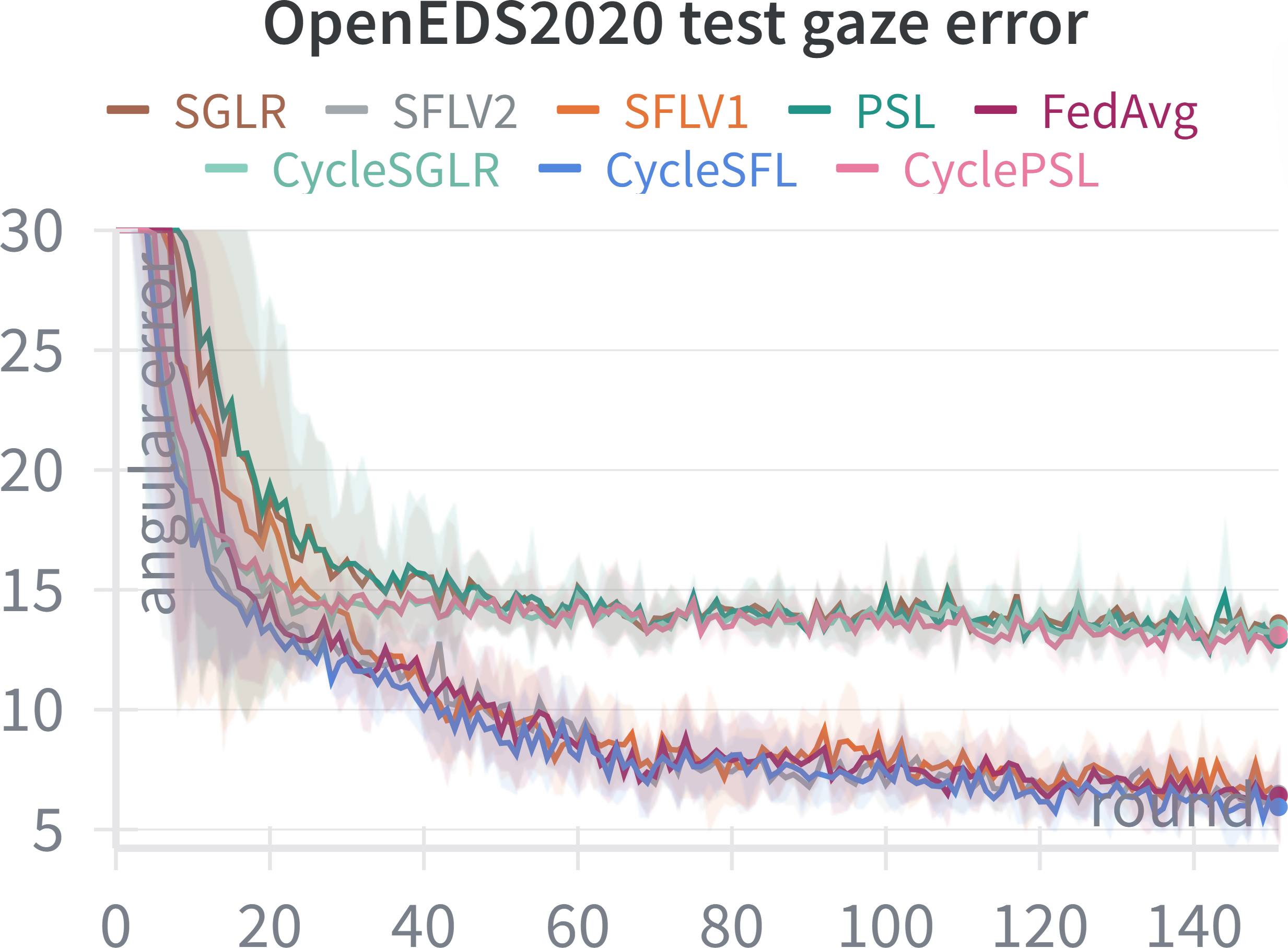}
    \label{fig:appendix_openeds_error}}
    \caption{Test metrics for the OpenEDS2020 task.}
    \label{fig:appendix_openeds_result}
\end{figure}

\clearpage
\subsection{Ablation study - impact of cut layer and server round}
The impact of cut layer (block-wise) on CycleSFL test loss on the CIFAR-100 dataset is plotted in Figure~\ref{fig:appendix_ablation_cut_cifar100_loss}. And the influence of server epoch on CycleSFL test loss on the CIFAR-100 dataset is visualized in Figure~\ref{fig:appendix_ablation_epoch_cifar100_loss}.
\setlength{\myheightfourth}{3.0cm}
\begin{figure}[!ht]
    \centering
    \subfloat[iid]{\includegraphics[height=\myheightfourth, keepaspectratio]{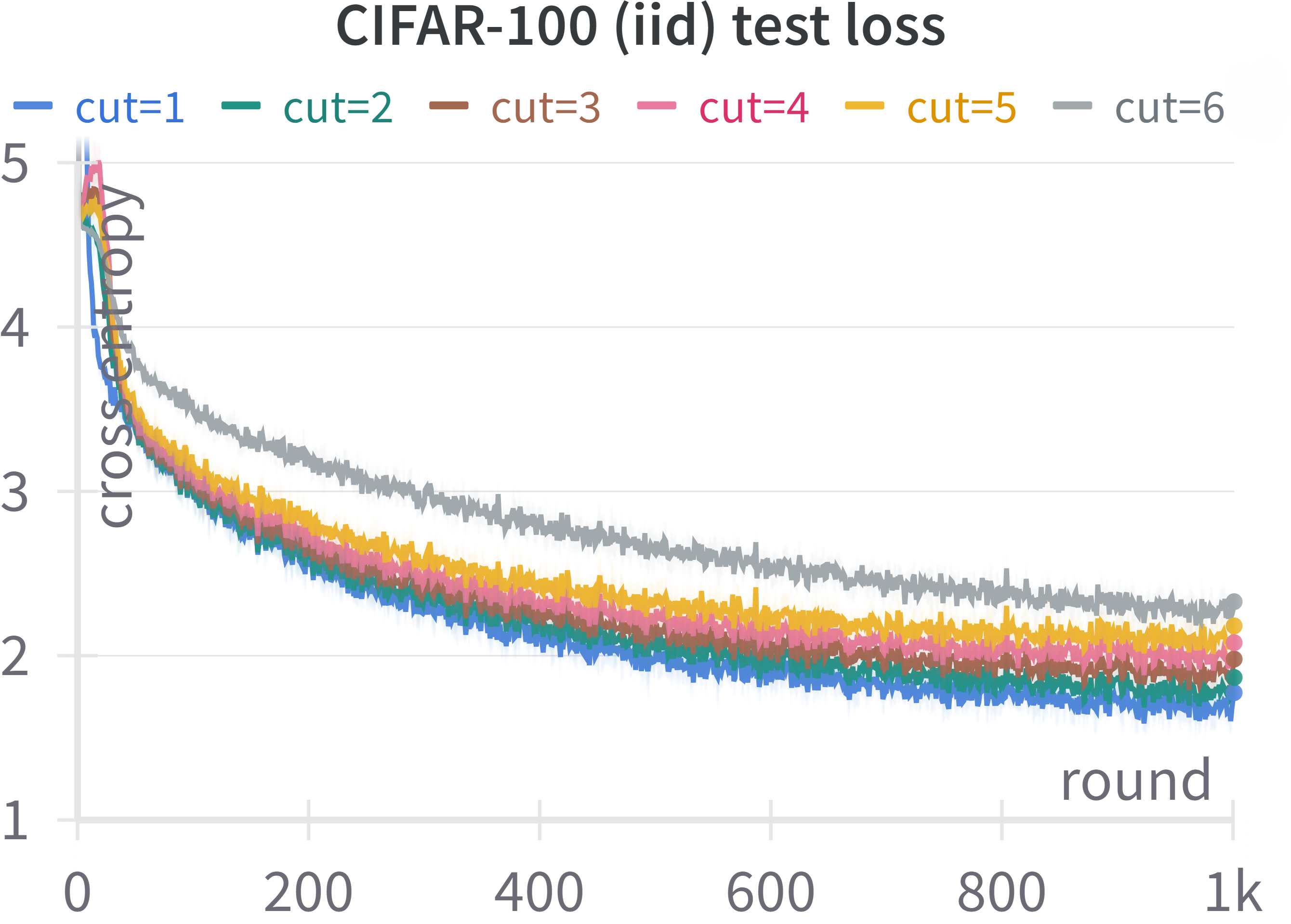}
    \label{fig:appendix_ablation_cut_cifar100_iid_loss}}
    \hfill
    \subfloat[$\alpha=1.0$]{\includegraphics[height=\myheightfourth, keepaspectratio]{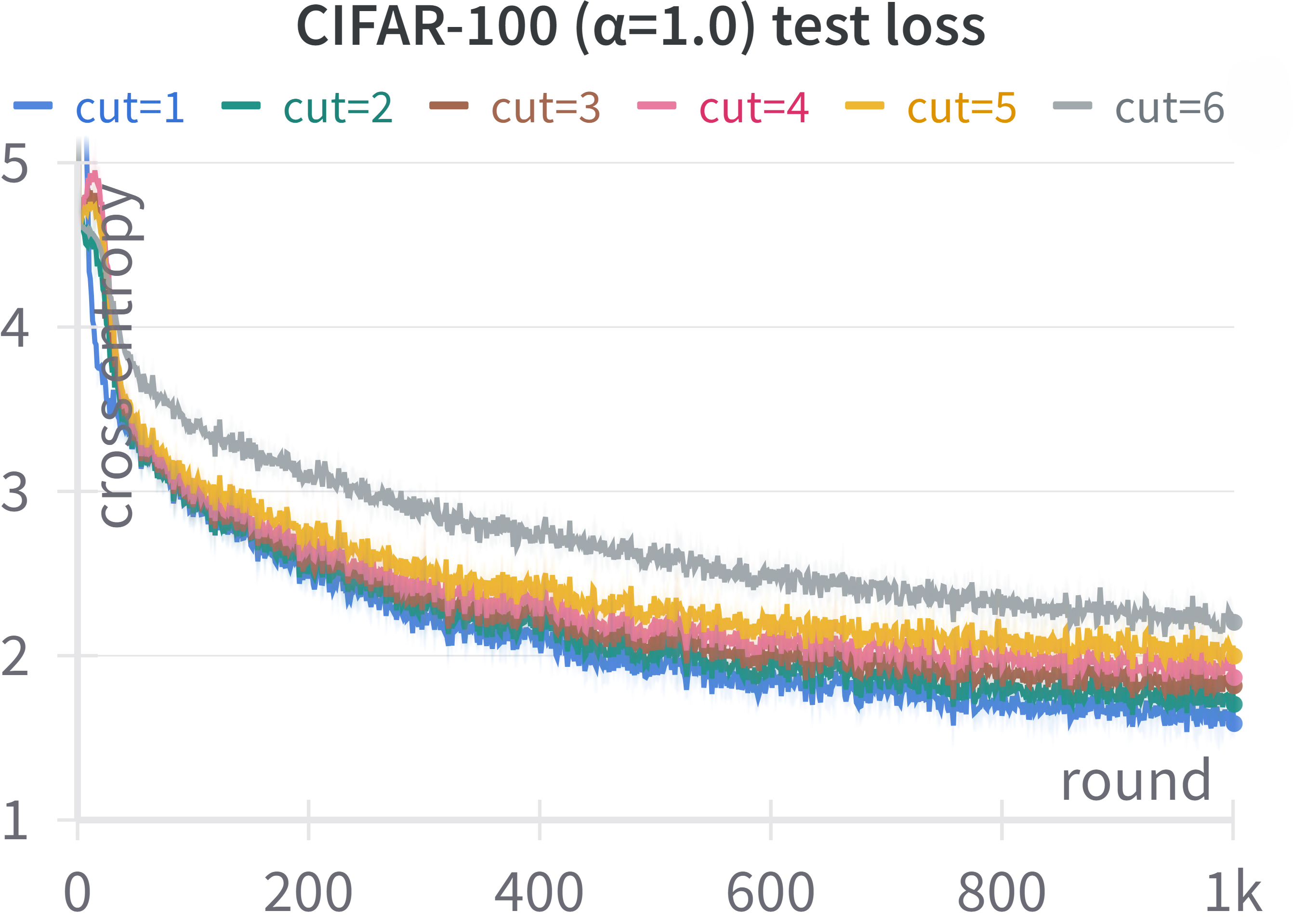}
    \label{fig:appendix_ablation_cut_cifar100_1.0_loss}}
    \hfill
    \subfloat[$\alpha=0.5$]{\includegraphics[height=\myheightfourth, keepaspectratio]{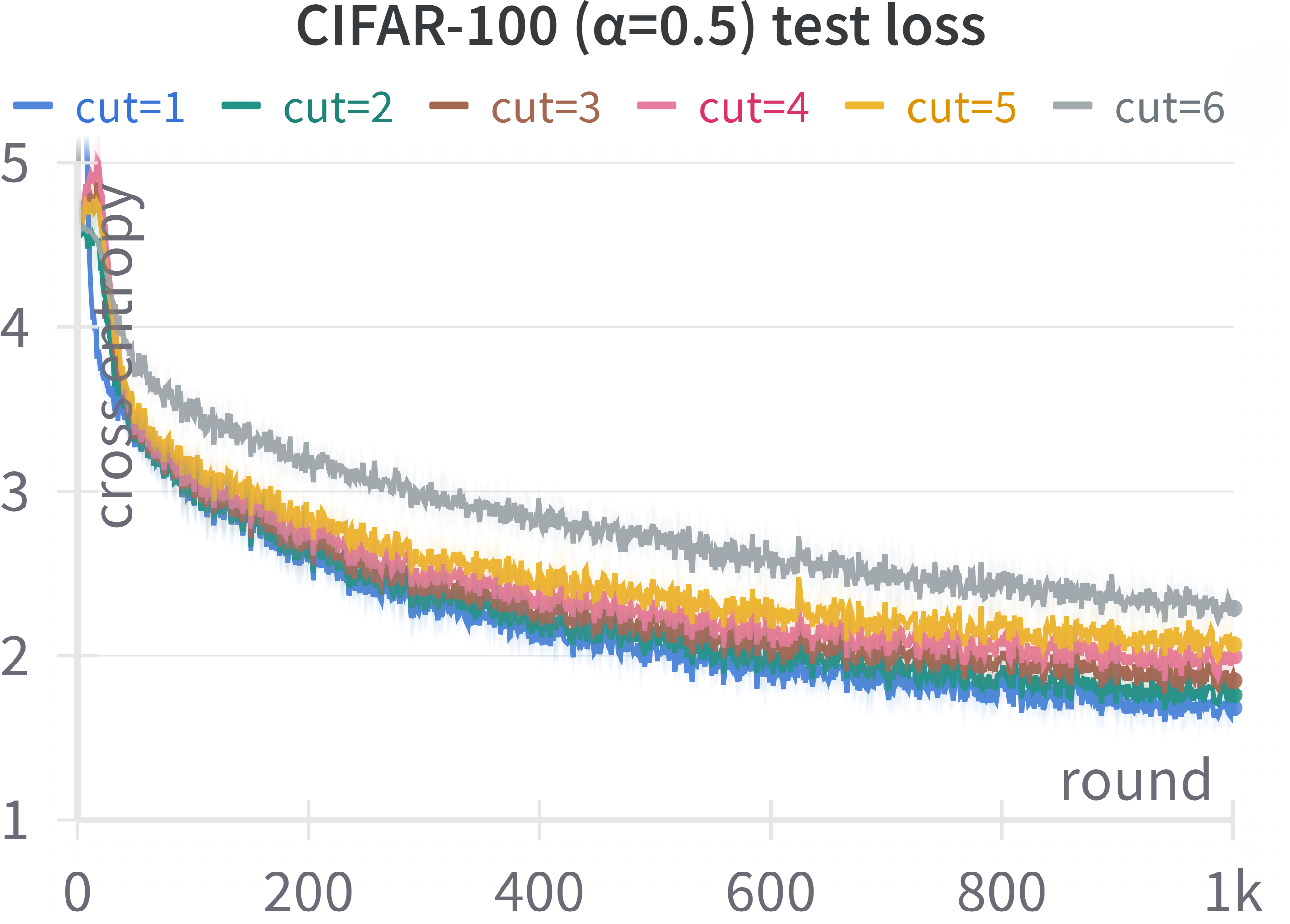}
    \label{fig:appendix_ablation_cut_cifar100_0.5_loss}}
    \hfill
    \subfloat[$\alpha=0.1$]{\includegraphics[height=\myheightfourth, keepaspectratio]{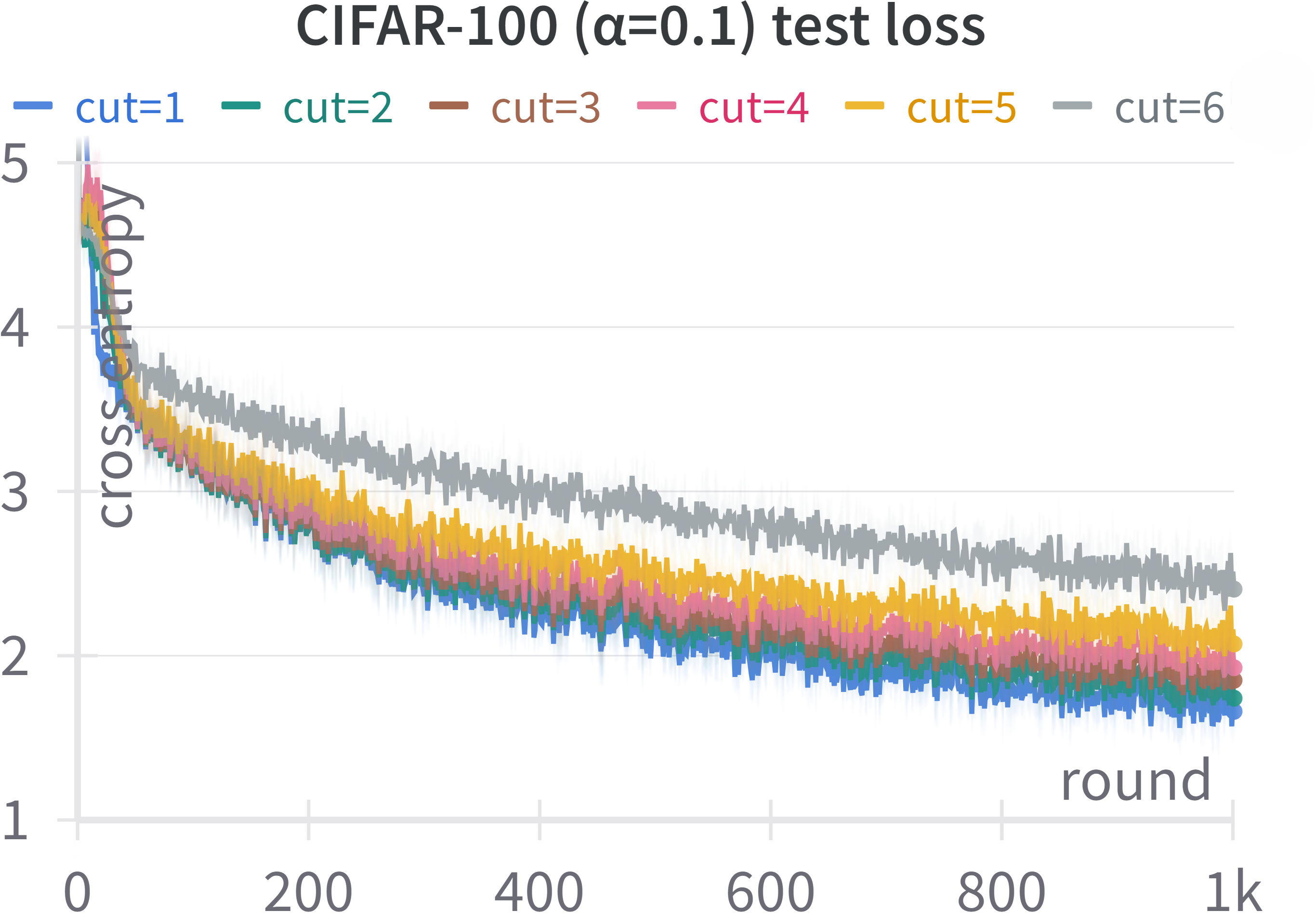}
    \label{fig:appendix_ablation_cut_cifar100_0.1_loss}}
    \caption{Impact of cut layer on CycleSFL test loss on CIFAR-100.}
    \label{fig:appendix_ablation_cut_cifar100_loss}
\end{figure}

\begin{figure}[!ht]
    \centering
    \subfloat[iid]{\includegraphics[height=\myheightfourth, keepaspectratio]{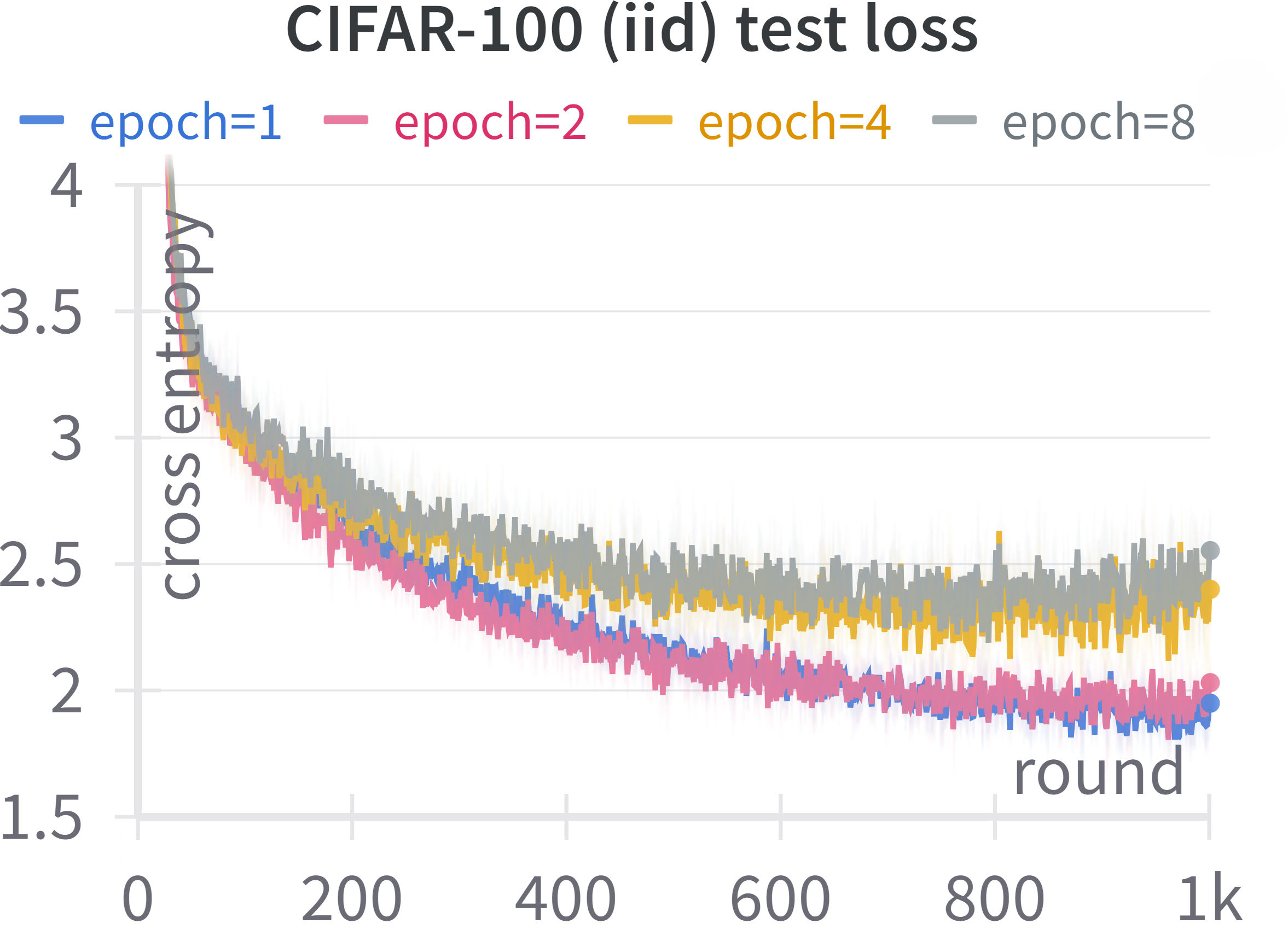}
    \label{fig:appendix_ablation_epoch_cifar100_iid_loss}}
    \hfill
    \subfloat[$\alpha=1.0$]{\includegraphics[height=\myheightfourth, keepaspectratio]{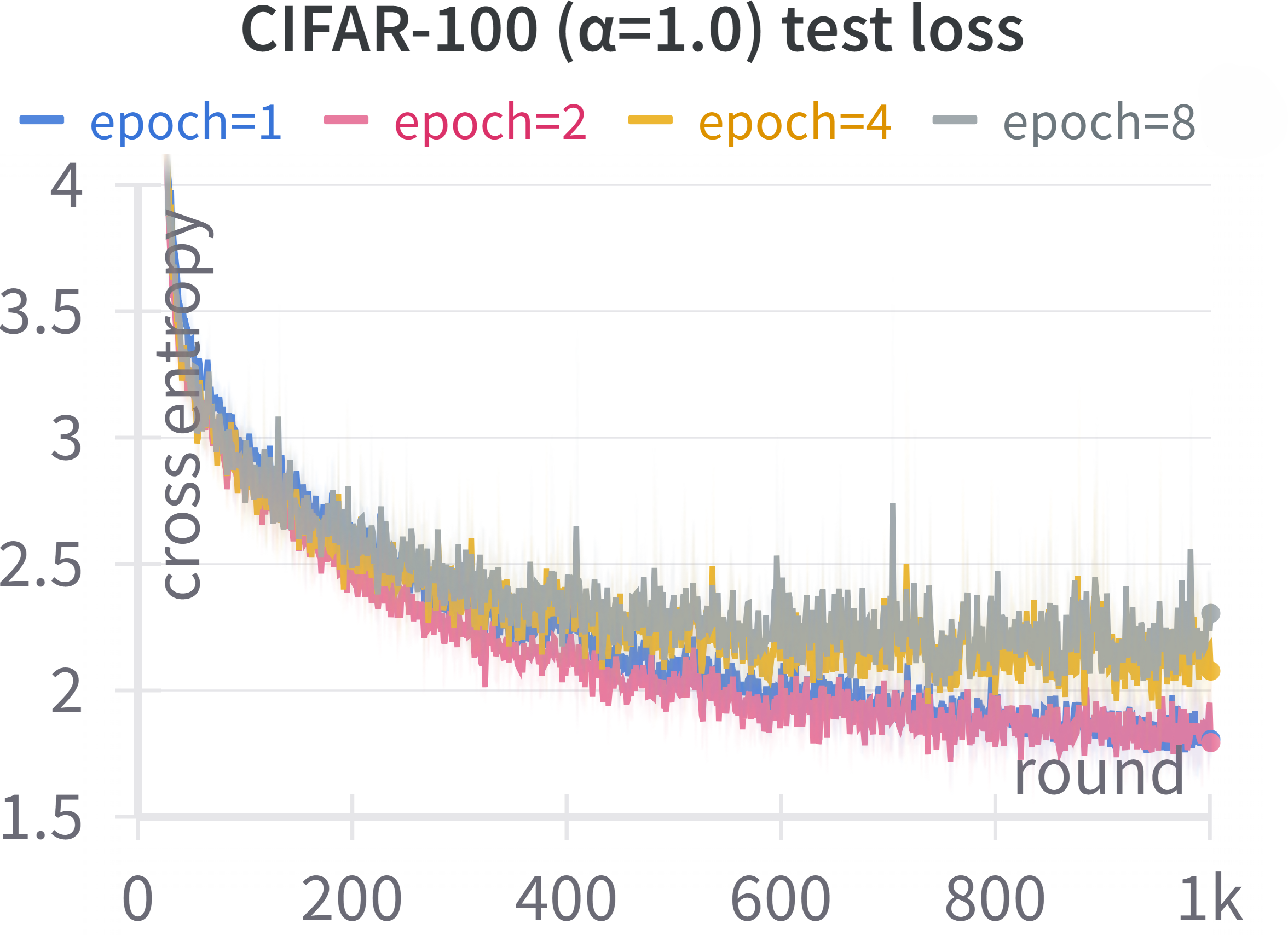}
    \label{fig:appendix_ablation_epoch_cifar100_1.0_loss}}
    \hfill
    \subfloat[$\alpha=0.5$]{\includegraphics[height=\myheightfourth, keepaspectratio]{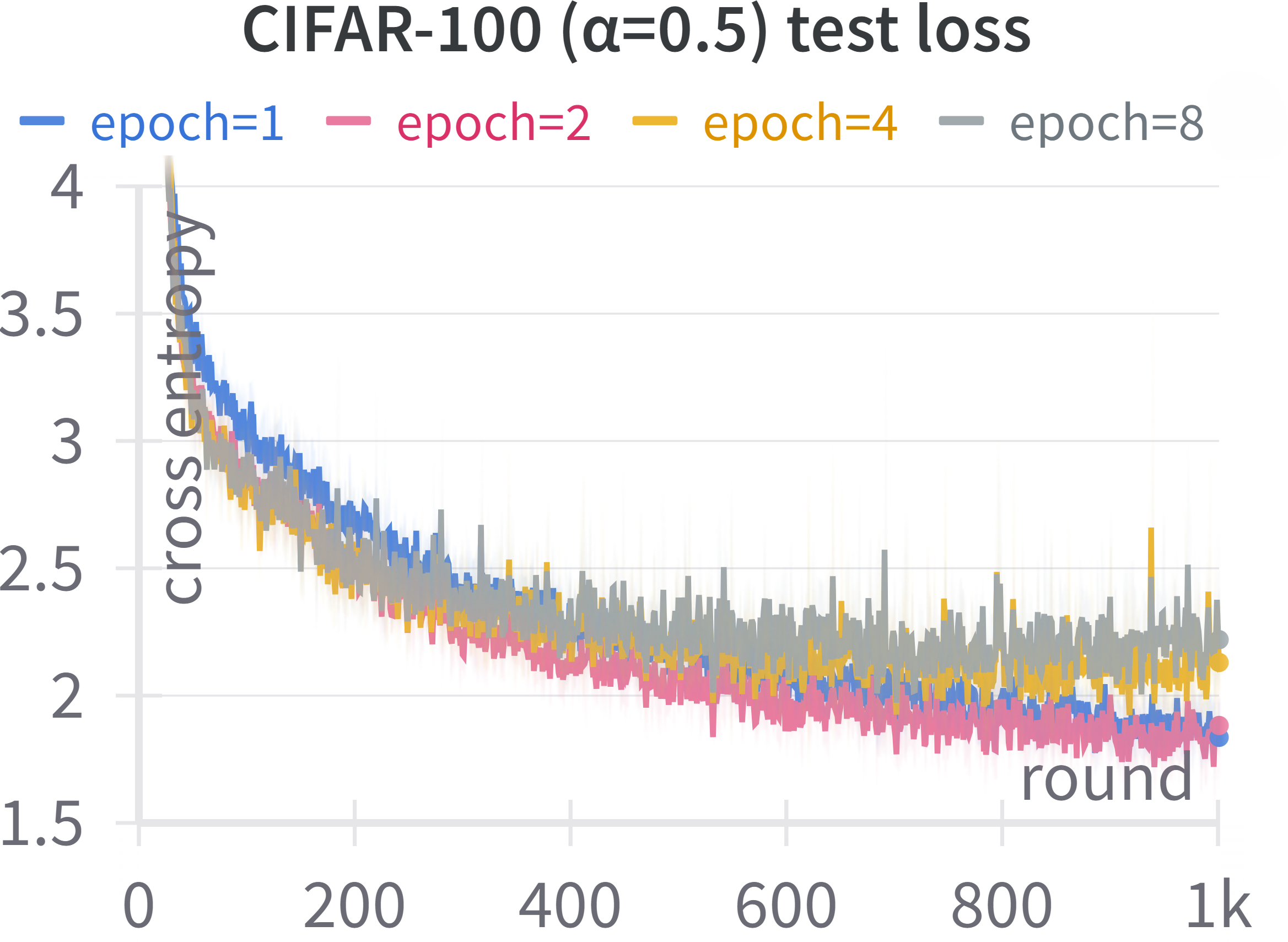}
    \label{fig:appendix_ablation_epoch_cifar100_0.5_loss}}
    \hfill
    \subfloat[$\alpha=0.1$]{\includegraphics[height=\myheightfourth, keepaspectratio]{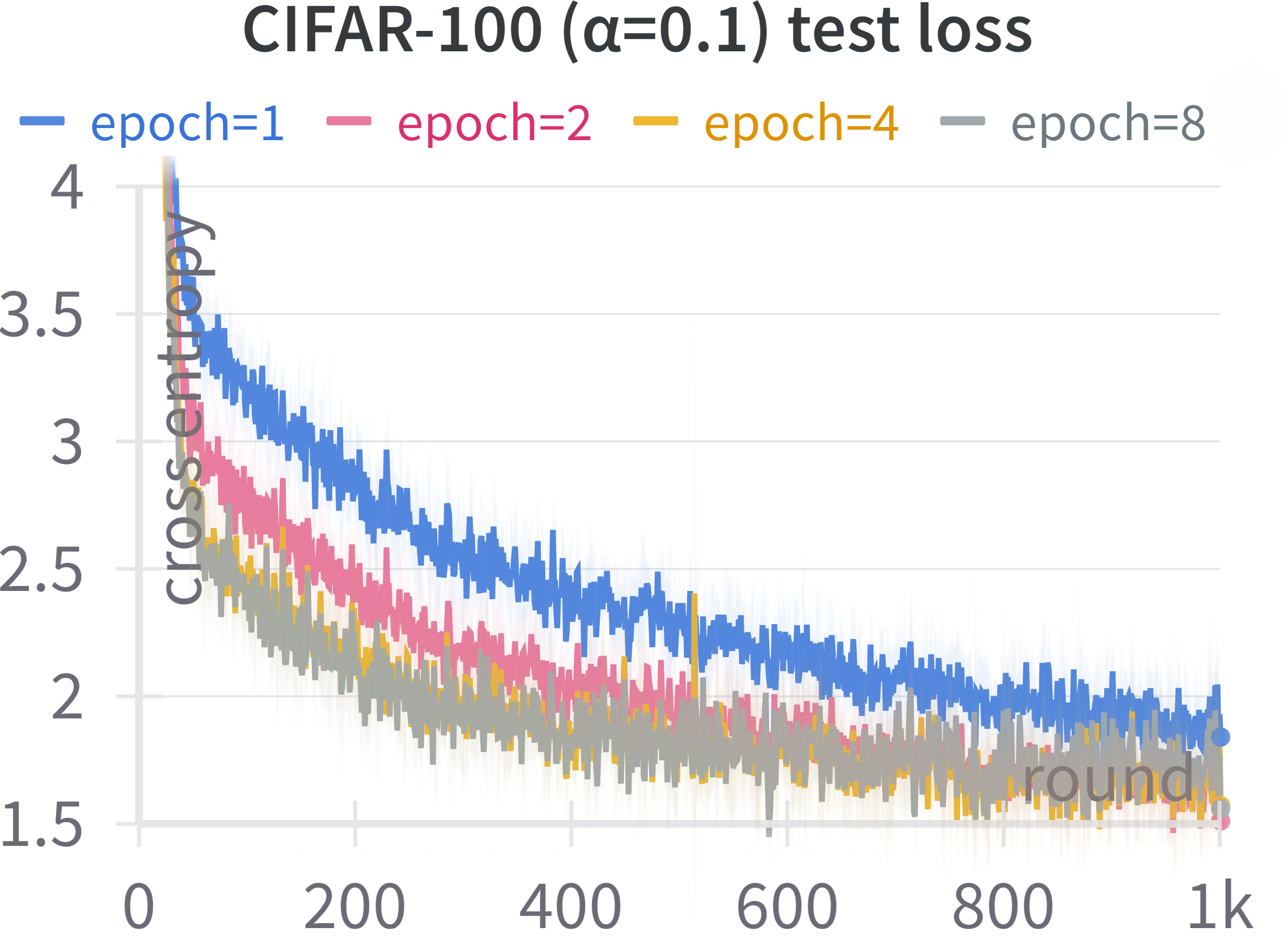}
    \label{fig:appendix_ablation_epoch_cifar100_0.1_loss}}
    \caption{Impact of server epoch on CycleSFL test loss on CIFAR-100.}
    \label{fig:appendix_ablation_epoch_cifar100_loss}
\end{figure}


\end{document}